\documentclass[sigconf]{acmart} 

\usepackage{comment}
\usepackage{color, colortbl}
\usepackage{caption}
\usepackage{subcaption}

\AtBeginDocument{%
  \providecommand\BibTeX{{%
    \normalfont B\kern-0.5em{\scshape i\kern-0.25em b}\kern-0.8em\TeX}}}

\definecolor{lightgreen}{rgb}{0.1, 0.7, 0.1}
\definecolor{darkyellow}{rgb}{0.9, 0.7, 0.0}
\definecolor{darkgreen}{rgb}{0.0, 0.4, 0.0}
\definecolor{darkred}{rgb}{0.6, 0.0, 0.0}

\def \debug{}
\ifx \debug \undefined
    \newcommand{\fx}[1]{{}}
\else
    \newcommand{\fx}[1]{{\textcolor{red}{#1}}}

\newcommand{\sysname}{IKIWISI}

\newcommand{\objs}{$\mathcal{O}$}
\newcommand{\objsub}{$\mathcal{O^*}$}
\newcommand{\foneglobal}{$F_{1}^{\mathcal{O}}$}
\newcommand{\fonelocal}{$F_{1}^{\mathcal{O^{*}}}$}

\newcommand{\mr}{Random}
\newcommand{\mgt}{GT}
\newcommand{\mgp}{GPV-1}
\newcommand{\mbp}{BLIP}
\newcommand{\mgpt}{GPT4V}

\def \billahdebug{}

\ifx \billahdebug \undefined
\newcommand{\fixmeth}[1]{{}}

\else

\newcommand{\fixmeth}[1]{{\bf\textcolor{red}{ [ Touhid FIXME: #1 ]}}}
\fi

\newcolumntype{L}[1]{>{\raggedright\let\newline\\\arraybackslash\hspace{0pt}}m{#1}}
\newcolumntype{C}[1]{>{\centering\let\newline\\\arraybackslash\hspace{0pt}}m{#1}}
\newcolumntype{R}[1]{>{\raggedleft\let\newline\\\arraybackslash\hspace{0pt}}m{#1}}

\newcolumntype{L}[1]{>{\raggedright\let\newline\\\arraybackslash\hspace{0pt}}m{#1}}
\newcolumntype{C}[1]{>{\centering\let\newline\\\arraybackslash\hspace{0pt}}m{#1}}
\newcolumntype{R}[1]{>{\raggedleft\let\newline\\\arraybackslash\hspace{0pt}}m{#1}}
\copyrightyear{2025}
\acmYear{2025}
\setcopyright{acmlicensed}\acmConference[DIS '25]{Designing Interactive Systems Conference}{July 5--9, 2025}{Funchal, Portugal}
\acmBooktitle{Designing Interactive Systems Conference (DIS '25), July 5--9, 2025, Funchal, Portugal}
\acmDOI{10.1145/3715336.3735754}
\acmISBN{979-8-4007-1485-6/2025/07}

\begin{document}

% \title{IKIWISI: A User-Centered Tool for Fostering Trust in Open-Vocabulary Multi-Modal Models in Real-World Tasks}

\title[IKIWISI: An Interactive Visual Pattern Generator]{IKIWISI: An Interactive Visual Pattern Generator for Evaluating the Reliability of Vision-Language Models Without Ground Truth}

\author[MT Islam]{Md Touhidul Islam}
\affiliation{%
  \institution{Pennsylvania State University}  
  \city{University Park}
  \state{PA}
  \country{USA}
}
\email{touhid@psu.edu}

\author[I Kabir]{Imran Kabir}
\affiliation{%
  \institution{Pennsylvania State University}  
  \city{University Park}
  \state{PA}
  \country{USA}
}
\email{ibk5106@psu.edu}

\author[MA Reza]{Md Alimoor Reza}
\affiliation{%
  \institution{Drake University}  
  \city{Des Moines}
  \state{IA}
  \country{USA}
}
\email{md.reza@drake.edu}

\author[SM Billah]{Syed Masum Billah}
\affiliation{%
  \institution{Pennsylvania State University}  
  \city{University Park}
  \state{PA}
  \country{United States}
}
\email{sbillah@psu.edu}

\begin{abstract}
We present IKIWISI ("\textbf{I} \textbf{K}now \textbf{I}t \textbf{W}hen \textbf{I} \textbf{S}ee \textbf{I}t"), an interactive visual pattern generator for assessing vision-language models in video object recognition when ground truth is unavailable. 
% IKIWISI transforms model outputs into a binary heatmap where green cells indicate human-model agreement on object presence and red cells show disagreement. 
IKIWISI transforms model outputs into a binary heatmap where green cells indicate object presence and red cells indicate object absence. 
This visualization leverages humans' innate pattern recognition abilities to evaluate model reliability. IKIWISI introduces "spy objects"—adversarial instances users know are absent—to discern models hallucinating on nonexistent items. 
The tool functions as a cognitive audit mechanism, surfacing mismatches between human and machine perception by visualizing where models diverge from human understanding.

Our study with 15 participants found that users considered IKIWISI easy to use, made assessments that correlated with objective metrics when available, and reached informed conclusions by examining only a small fraction of heatmap cells. This approach not only complements traditional evaluation methods through visual assessment of model behavior with custom object sets, but also reveals opportunities for improving alignment between human perception and machine understanding in vision-language systems.
\end{abstract}




% \input{sections/0.abstract}
%%%%%%%%%%%%%%%%%%%%%%%%%%%%%%%%%%%%%%%%%%%%%%%%%%%%%%%%%%%%%%%%
%%%%%%%%%%%%%%%%%%%%%% START OF THE PAPER %%%%%%%%%%%%%%%%%%%%%%
%%%%%%%%%%%%%%%%%%%%%%%%%%%%%%%%%%%%%%%%%%%%%%%%%%%%%%%%%%%%%%%%
\begin{teaserfigure}
    \centering
  \includegraphics[width=1.0\linewidth]{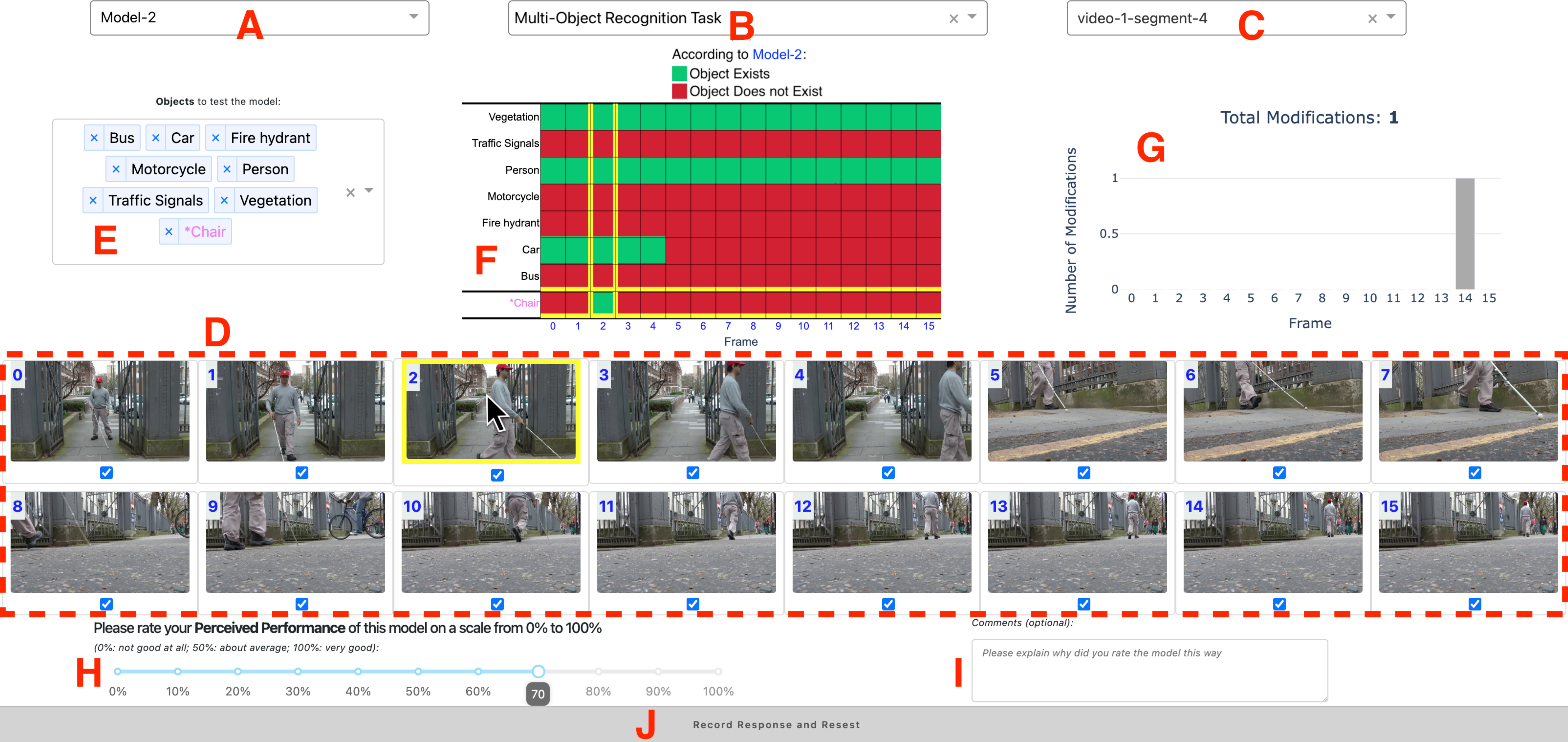}
    \caption{
    \sysname{}'s interactive interface for evaluating vision-language models (best viewed in color). The interface shows model and video selection options (A-C), video keyframes (D), object selection panel (E), and the core binary heat map (F), where green and red cells create visual patterns that help users assess model reliability. All the components and their roles are described in detail in Sec.~\ref{subsec:all-components}.
    }
    \label{fig:dashboard_final}
\end{teaserfigure}
\Description{The image is a screenshot of the proposed tool. It features a heatmap with red and green colored grid cells in the center, an image pane at the bottom, a few more user interface elements, such as two drop-down boxes, an input box for entering keywords, a bar chart, and a progress bar. }

\maketitle

\section{Introduction}
Human visual perception is a high-bandwidth input channel that allows sighted individuals to discern patterns, trends, and anomalies in the physical world~\cite{todorovic2008gestalt}. 
Furthermore, humans possess an intricate understanding of the world, i.e., commonsense, such as objects occupying physical space and obeying physical laws~\cite{jia-3d-stability-pami}. Combined, these abilities enable humans to easily see and judge things that are otherwise difficult to define or explain formally.

Recent open-vocabulary, large multi-modal (LMM) vision–language models such as GPT-4 have become increasingly integrated into everyday interactions, especially for individuals with sensory disabilities~\cite{xie2025beyond, xie2024emerging}. These LMMs interpret and generate information across various modalities, including text, images, and video. However, unlike humans, their responses often lack grounding in commonsense~\cite{brachman2023machines, forbes2019neural} and cannot guarantee accuracy.

Fact-checking the responses of an LMM is non-trivial for general users. For instance, in a scenario where an LMM is deployed to recognize a set of objects of interest (e.g., `an overhanging tree branch', `pet waste') for outdoor navigation~\cite{islam2024identifying}, checking the model's response -- whether an object exists in the current frame or not -- in real-time is difficult, if not impossible. 
Existing closed-vocabulary models like YOLOv7~\cite{wang2023yolov7} have limitations in recognizing many objects of interest. In the above scenario, YOLOv7 cannot recognize `pet waste' by default, whereas the response of open-vocabulary LMMs cannot be trusted by default.

This challenge reveals a fundamental alignment problem in AI literature: \textit{How can we bridge the gap between human commonsense understanding of the visual world and the capabilities of AI systems that lack this implicit knowledge?}
Building on this alignment challenge, our paper attempts to harmonize users' commonsense understanding of real-world object saliency with LMMs' ability to visually discriminate these objects in a dynamically perceptible way.
This leads to our key research question: \textit{``How do we design an interface for LMM output that enables users to evaluate the model's performance easily and subsequently fosters trust in the model in real-world applications?''}

To address this question, we designed {\sc \textbf{\sysname{}}}, pronounced ``icky-wissy,'' an acronym for ``\textbf{I} \textbf{K}now \textbf{I}t \textbf{W}hen \textbf{I} \textbf{S}ee \textbf{I}t.'' 
\sysname{} is an interactive tool that generates visual patterns to help users evaluate the reliability of LMMs for recognizing multiple objects in real-world videos, particularly when ground-truth data is unavailable.

Users can select their own set of objects on their chosen video and evaluate a model's output to determine its reliability (Fig.~\ref{fig:dashboard_final}). 
Given the highly visual nature of the tool, \sysname{} is designed for a range of users with varying levels of expertise. 
Technical users, such as AI researchers and engineers with specialized knowledge, can utilize the tool to evaluate various models and select the most reliable one for a specific environment. 
Meanwhile, domain experts with limited AI knowledge (such as urban planners or accessibility specialists) can use \sysname{} to assess whether AI models meet their specific needs, even without understanding the underlying technical details. 
Our user study, which included participants across this expertise spectrum, demonstrates that \sysname{} enables consistent reliability assessments regardless of technical background (Sec.~\ref{sec:expertise-role}).

We adopted a \textbf{Research-through-Design}~\cite{research_through_design} approach and iteratively refined the prototype based on user feedback and insights across multiple phases.
At the heart of \sysname{} is a binary heat map that abstracts video content into a collection of cells. 
Each cell represents a user-selected object and is assigned a color (green or red) based on the model's output to track its existence across time. 
These colors create high-level patterns that users can glance over to notice anomalies, guiding their attention to important cells for further investigation. 
Through this process, users can reach a conclusion about the model's reliability.
Importantly, users only need to inspect a small fraction of the heatmap cells to make informed decisions about a model. 
The binary heat map currently focuses on presence/absence detection, with potential for future integration of model confidence scores to provide additional transparency into the reliability of individual predictions.

A study conducted with 15 participants strongly suggests that \sysname{} is user-friendly and empowers participants to rate a model's reliability in a manner that correlates with its true performance (when available).
Our findings highlight the potential of \sysname{} as a valuable framework that complements existing automated evaluation techniques for AI models. 
By enabling laypeople to assess AI models according to their specific needs, \sysname{} democratizes the evaluation process. 
Furthermore, the tool promotes transparency by allowing users to interpret model performance visually.

While we demonstrate \sysname{} for multi-object recognition--a foundational task for applications like dynamic scene analysis~\cite{behley2019iccv}, video surveillance~\cite{blanch2024lidar}, robotics~\cite{karaoguz2019object, paul2021object}, and autonomous driving~\cite{li2019gs3d, feng2020deep, kabir2025logicrag}--our approach offers a framework that could extend to other visual AI tasks. 
The binary heat map approach could be adapted to evaluate image captioning, visual reasoning, or open-ended visual question answering, where ground truth may be subjective or unavailable.

Beyond its practical utility, \sysname{} functions as a ``cognitive audit tool'' that exposes discrepancies between human expectations and model behavior. 
When users notice inconsistent patterns in the heat map, they directly confront the boundaries of the model's understanding compared to their own commonsense reasoning. This audit process not only helps users make informed decisions about model reliability but also advances human-AI alignment by making these mismatches transparent rather than hidden within the model's black box.

The significance of our work lies in several key contributions to human-centered AI evaluation: \textbf{First}, we design \sysname{}, an interactive framework that bridges human commonsense reasoning and AI visual understanding, which enables users to evaluate the reliability of large multi-modal models without requiring ground truth data (Sec.~\ref{subsec:all-components}).
\textbf{Second}, \sysname{} introduces a simple yet effective heat map visualization that transforms complex video content into interpretable patterns, allowing users with varying levels of expertise to identify model limitations efficiently (Sec.~\ref{subsec:heatmap}). 
\textbf{Third}, our findings demonstrate that users can make accurate reliability assessments by inspecting only a small fraction of heat map cells, thus significantly reducing cognitive load while maintaining judgment quality (Sec.~\ref{subsec:role_of_patterns}). 
\textbf{Fourth}, a user study with 15 participants confirms that \sysname{} enables reliability assessments that align closely with objective performance metrics when available—validating its effectiveness for real-world deployment (Sec.~\ref{sec:results}).
\textbf{Finally}, \sysname{} represents a new paradigm for human-centered AI evaluation, functioning not just as a usability tool but as a ``cognitive audit mechanism'' that exposes misalignments between human commonsense expectations and model reasoning (Sec.~\ref{sec:discussion}).
\section{Background and Related Work}

\subsection{Open Vocabulary, Large, Multi-Modal Models}
Large Multi-Modal Models (LMMs) like GPT-4~\cite{openai2023gpt4,openai2023gpt4vsc,openai2023gpt4vtwa}, LLaVA~\cite{liu2023visual}, BLIP~\cite{li2022blip,li2022lavis}, and GPV-1~\cite{Gupta2021GPV} learn representations that integrate visual and textual data and demonstrate remarkable capabilities in out-of-distribution reasoning, common sense understanding, and knowledge retrieval~\cite{bubeck2023sparks}. These models use multi-modal learning strategies, primarily self-supervised learning on massive datasets, followed by image-text pair training and human feedback to align visual and linguistic features in a shared embedding space~\cite{gupta2022towards}.
This integration of language and vision offers significant promise for robotics, autonomous driving, and accessibility applications like wheelchair navigation~\cite{yang2023survey, zhou2023vision}. By combining textual or symbolic modalities with visual data, LMMs can potentially overcome interpretability challenges and decision-making opacity in current systems. In theory, their ability to incorporate language enables them to provide human-understandable explanations for their decisions and actions, making them more trustworthy~\cite{chen2023driving}.

Despite these advances, LMMs exhibit critical limitations. Studies reveal their struggles with fine-grained spatial relationships (e.g., \textit{in front of}, \textit{behind})~\cite{liu2023visualspatial, kamath2023s, kabir2025logicrag}, word order sensitivity (e.g., \textit{cat chased dog} vs. \textit{dog chased cat})~\cite{thrush2022winoground}, and visio-linguistic compositionality—the ability to combine visual and linguistic elements to understand novel concepts~\cite{yuksekgonul2022and}. Perhaps more concerning, these models can generate content they do not fully understand~\cite{west2023generative}. These capabilities are essential for building trust, a prerequisite for deploying LMMs in practical tasks within complex environments.

These limitations stem from two fundamental issues. \textit{First}, LMMs learn directly from data without explicitly encoding knowledge about physical world principles, such as objects occupying space and following physical laws~\cite{jia-3d-stability-pami}. This design choice leads to errors, and when prompted to explain their mistakes, these models often produce explanations that lack coherence while agreeing with any information the prompter provides—making their errors difficult to trace, reproduce, or diagnose.
\textit{Second}, commercial black-box LMMs lack transparent evaluation metrics that users can interpret and trust. Current models report ad hoc performance measures~\cite{gptv_system_card}, such as refusal rates for generating harmful, hateful, or biased content (known as ``jailbreaking''). These non-standard, heuristic-driven metrics, limited by vendors' internal testing protocols, resist reproduction and meaningful cross-model comparison.

This situation raises a practical question: what can users do to evaluate these models? Our tool addresses this need by enabling users to determine which models perform better for specific tasks (e.g., multi-object recognition in real-time) in particular contexts (e.g., urban environments). By testing models with representative videos of their intended settings, users can make informed decisions about model selection based on empirical evidence rather than vendor claims.

\subsection{Common Evaluation Metrics for LMMs}
The multi-modal capabilities that make LMMs powerful also create unique evaluation challenges, as these models process and integrate text, images, audio, and other modalities in ways that resist simple measurement. Current evaluation approaches fall into several categories, each with distinct strengths and limitations.

\textit{Cross-Modal Matching and Retrieval Accuracy.}
Many researchers assess LMMs by measuring their ability to match or retrieve information across different modalities~\cite{liu2024visual, chan2023clair}. These evaluations often focus on instruction-following capabilities—how well models understand and execute commands ranging from conversational requests to detailed instructions involving complex reasoning~\cite{liu2024visual}. 
A common methodology employs another Large Language Model, such as \textit{Llama 2}, as an evaluation judge~\cite{liu2024visual,chan2023clair,maaz2023video}. This judge analyzes the original question, the visual content, and the candidate model's responses, then rates each response on dimensions like helpfulness, relevance, accuracy, and detail, providing both a numerical score and explanatory reasoning~\cite{liu2024visual}.

\textit{Task-Specific Performance Metrics.}
For specialized applications like visual question answering (VQA)~\cite{antol2015vqa}, researchers apply domain-specific performance metrics~\cite{maaz2023video}. These include accuracy, F-scores, and Mean Reciprocal Rank (MRR), which evaluate how correctly models answer questions about visual content in standardized test datasets. Each metric captures a different aspect of model performance, with varying sensitivities to different types of errors.

For evaluating consistency in model-generated content across frames or prompts, traditional metrics include CLIP Cosine Similarity~\cite{qi2023fatezero} and Learned Perceptual Image Patch Similarity (LPIPS)~\cite{zhang2018unreasonable}. 
These similarity measures assess how consistently models maintain outputs when input frames contain minimal changes. However, these metrics prove less relevant for multi-object recognition, our primary focus in this work, which demands precise identification rather than general consistency.

Our approach diverges from these established metrics by prioritizing human perception as the evaluation baseline. To compare how well user judgments align with objective performance measures, we selected the classic $F_1$ score as our benchmark. This metric combines precision (how many identified objects are correct) and recall (how many actual objects were identified), providing a balanced assessment of whether an object $x$ appears in frame at time $t$. 

Since calculating $F_1$ scores requires ground truth data with predefined object categories, we created a specialized dataset and object taxonomy specifically for this purpose, described in detail in Sec.~\ref{sec:dataset}. This dataset enables us to measure the correlation between user perceptions of model reliability and the models' actual performance in controlled settings.

\subsection{Model Performance Visualization}
The majority of research on visually interpreting machine learning (ML) models focuses on the visualization of the internal workings of a model~\cite{kahng2017cti,liu2017analyzing,liu2017towards,hohman2018visual}. These visualization approaches primarily serve ML researchers and experts who need to understand the underlying mechanisms of model behavior, but they often remain inaccessible to everyday users seeking practical evaluations of model performance.
In contrast, tools designed for end-user model assessment take different approaches to visualization. Alsallakh et al.~\cite{alsallakh2014visual} present a \textit{Confusion Wheel}, which arranges different classes in a radial layout and displays the statistics of the confusion matrix associated with each class, along with the model's prediction confidence using a histogram. \textit{Squares}~\cite{ren2016squares} offers a visualization of prediction scores (confidence) of multi-class classifiers by combining a set of histograms and allows users to compare multiple histograms visually. To facilitate model comparison, \textit{Manifold}~\cite{zhang2018manifold} utilizes scatter plot-based visual summaries to provide an overview of the general outcomes of ML models, along with a customizable tabular view that reveals feature discrimination.

These existing visualization systems, while valuable for static image analysis, present three key limitations for our context. First, they often overlook the temporal aspect of the data, focusing solely on classifier performance for individual images rather than sequences. Second, interpreting confidence scores becomes challenging due to the use of different thresholds across various applications. Third, in closed-source LMMs, the management of confidence scores and thresholds typically happens internally, making these values unavailable for visualization.

To address these limitations, our work provides a graphical framework that enables humans to interactively evaluate model outputs in the absence of ground truth—a common situation in real-world tasks. The closest visualization to ours is ARGUS~\cite{castelo2023argus}, where 2D heat maps visualize models' output along the temporal axis. However, unlike ARGUS, where each cell represents a model's confidence of an object being present, cells in our heat map indicate either an agreement (green) or disagreement (red)—agreement if both the human and the model see the object at that time, disagreement otherwise. 

Our approach considers the user's perception as the baseline, treating any mismatch between model output and user perception as a disagreement. This design puts users at the forefront, prioritizes their objectives for the particular task for which they need assistance, and allows them to choose the best model from a set of candidates based on their specific needs rather than abstract performance metrics.

\subsection{Role of Human Perception in Decision Making}
Humans rely heavily on their perceptual abilities—visual, auditory, or tactile—when making judgments under adversarial or time-constrained conditions. Research demonstrates how these perceptual capabilities guide complex decision processes across various domains. Sighted users can detect visual anomalies—when an element differs from its peers—as quickly as 250 milliseconds through pre-attentive vision~\cite{treisman1985preattentive}. These anomalies can involve differences in color, shape, size, orientation, length, and even quantities~\cite{ward2010interactive}.

Visual pattern recognition proves especially valuable in domains requiring quick assessment. Search engines rely on network visualization algorithms like PageRank~\cite{page1999pagerank} to determine webpage relevance and authority through connection patterns, while social media platforms use similar principles to identify influential users and content. In academia specifically, researchers interpret network visualizations of co-authorship where patterns of connectivity help predict scholarly impact~\cite{newman2004coauthorship, billah2015social}. Financial traders likewise depend on candlestick chart patterns to make rapid stock market decisions, translating visual cues into actionable insights~\cite{nison2001japanese}. In all these cases, visual representations transform complex relationships into intuitive patterns that humans process more efficiently than raw data.

Beyond visual patterns, other sensory modalities demonstrate similar capabilities. When determining whether audio was generated by humans or AI, blind users depend on their auditory perception to detect distinctive human speech characteristics such as natural pauses, lip sounds, vocal fry, and regional accents~\cite{han2024uncovering}. In broadcast media, producers struggle to balance on-screen representation of phenotypic traits in real-time, yet perform this task efficiently when provided with visual aids such as bullet bar charts displaying demographic distributions~\cite{hoque2020toward}. These examples illustrate how humans naturally process perceptual patterns to form judgments, especially under constrained conditions. This understanding guided our interface design principles: create simple, real-time visual patterns that align with users' normative expectations, that enable them to leverage their inherent pattern recognition abilities when evaluating AI systems.

\begin{figure*}[!ht]
    \begin{minipage}{.48\textwidth}
        \centering
        \includegraphics[width=.9\linewidth]{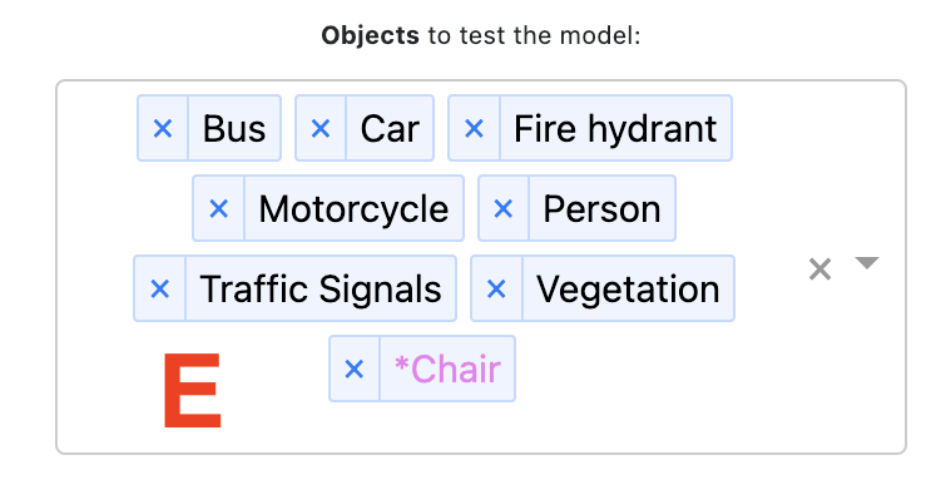}
        \caption{
          Object Selection Panel (E) enlarged from Fig.~\ref{fig:dashboard_final}. Objects prefixed with `*' and displayed in \textcolor{violet}{\textbf{violet}} function as adversarial `spy' instances (e.g., `Chair' in this case) that test the model's ability to recognize object absence.
        }
        \label{fig:panel_e}        
    \end{minipage}
    \hfill
    \begin{minipage}{.48\textwidth}
   \centering
   \includegraphics[width=\linewidth]{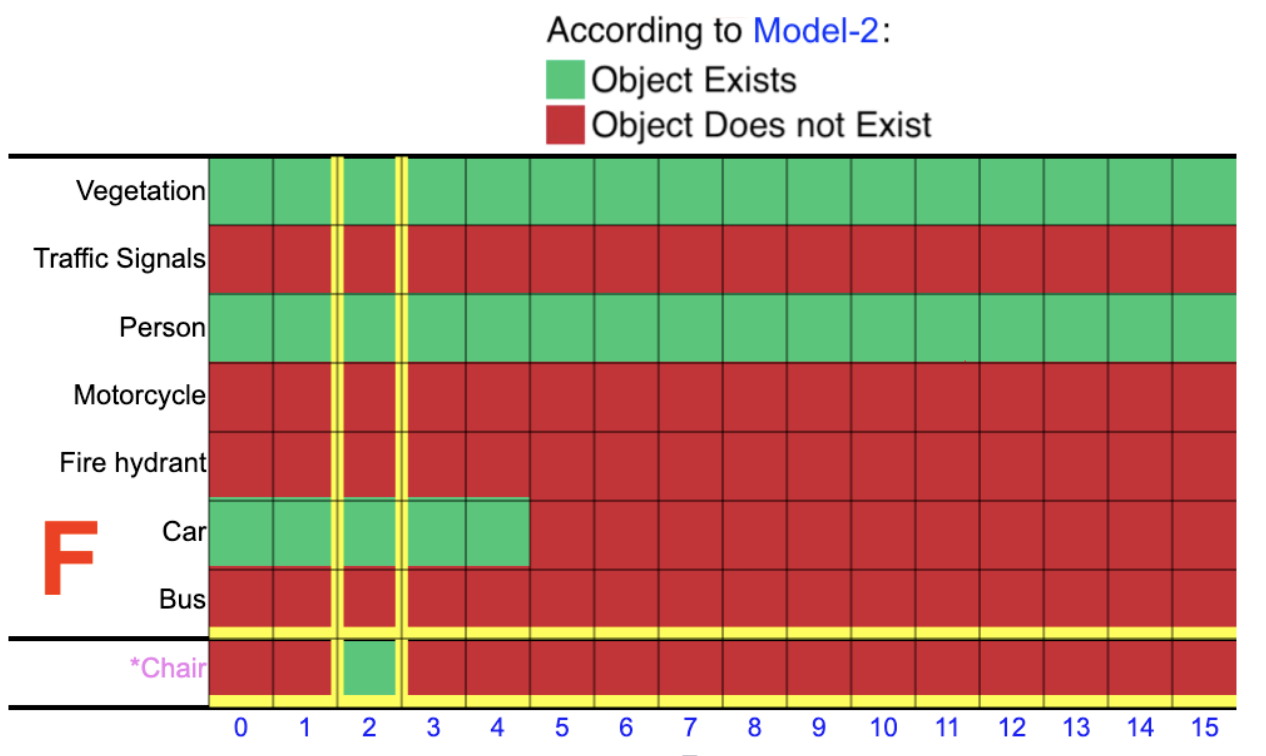}
   \caption{
       Binary Heat Map (F) enlarged from Fig.~\ref{fig:dashboard_final}, showing the core visualization where green cells indicate objects the model recognizes and red cells represent objects it does not recognize. 
   }
   \label{fig:panel_f}
\end{minipage}
\end{figure*}

\subsection{Human Trust in AI Models}
Historical studies from the mid-1980s outlined key principles of when and how humans trust intelligent systems~\cite{fernandez1998building, carroll1988mental, norman1986user, johnson1986cooperative}.
\textit{First}, when a system's rationale aligns with a user's understanding, this alignment bolsters trust and reduces skepticism toward the system's advice. Without such explanations, users often form incorrect interpretations or assumptions about how the system functions~\cite{datta2023s, wang2023human, carroll1988mental, qadir2022toward}.
\textit{Second}, users attribute intelligence to systems that demonstrate understanding of their needs, expectations, and objectives. Systems that fail to acknowledge these user priorities significantly erode trust over time.

For effective human-AI collaboration, systems must recognize and adapt to users' knowledge, intentions, and preferences.
Recent research~\cite{riedl2019human, wang2023human, qadir2022toward} has expanded these classical findings, broadening the focus to create AI systems that embody transparency, accountability, and alignment with human values. IKIWISI embodies these principles by creating a transparent interface where users can evaluate alignment between their visual understanding and the model's capabilities. This transparency allows users to build trust incrementally, adding objects of interest and observing whether models recognize these objects as humans do, thus making both alignment assessment and trust formation an interactive process.

\section{Overview of \sysname{}}

At \sysname{}'s core lies an interactive binary heat map where columns represent video keyframes and rows represent user-selected objects. Fig.~\ref{fig:dashboard_final} presents the key components and features of \sysname{}. This section describes these components, their contributions to the design, and the technical implementation of \sysname{}.

\subsection{Components of \sysname{}}
\label{subsec:all-components}

\subsubsection{\textbf{Model, Task and Video Dropdowns (\textbf{A}, \textbf{B}, \textbf{C})}}
\sysname{} features three selection dropdowns: a Model dropdown (A) for choosing from various Large Multi-Modal Vision Language Models, a Task dropdown (B) for selecting the specific task (currently limited to multi-object recognition in video), and a Video dropdown (C) for picking a specific video segment to analyze. 
Future versions could expand the available tasks. For testing, we provided five different models in the model dropdown, with details on these models and their output generation in Sec.~\ref{sec:models}.

\subsubsection{\textbf{Image Container (\textbf{D})}}
\begin{figure*}[!ht]
    \centering
    \includegraphics[width=\linewidth]{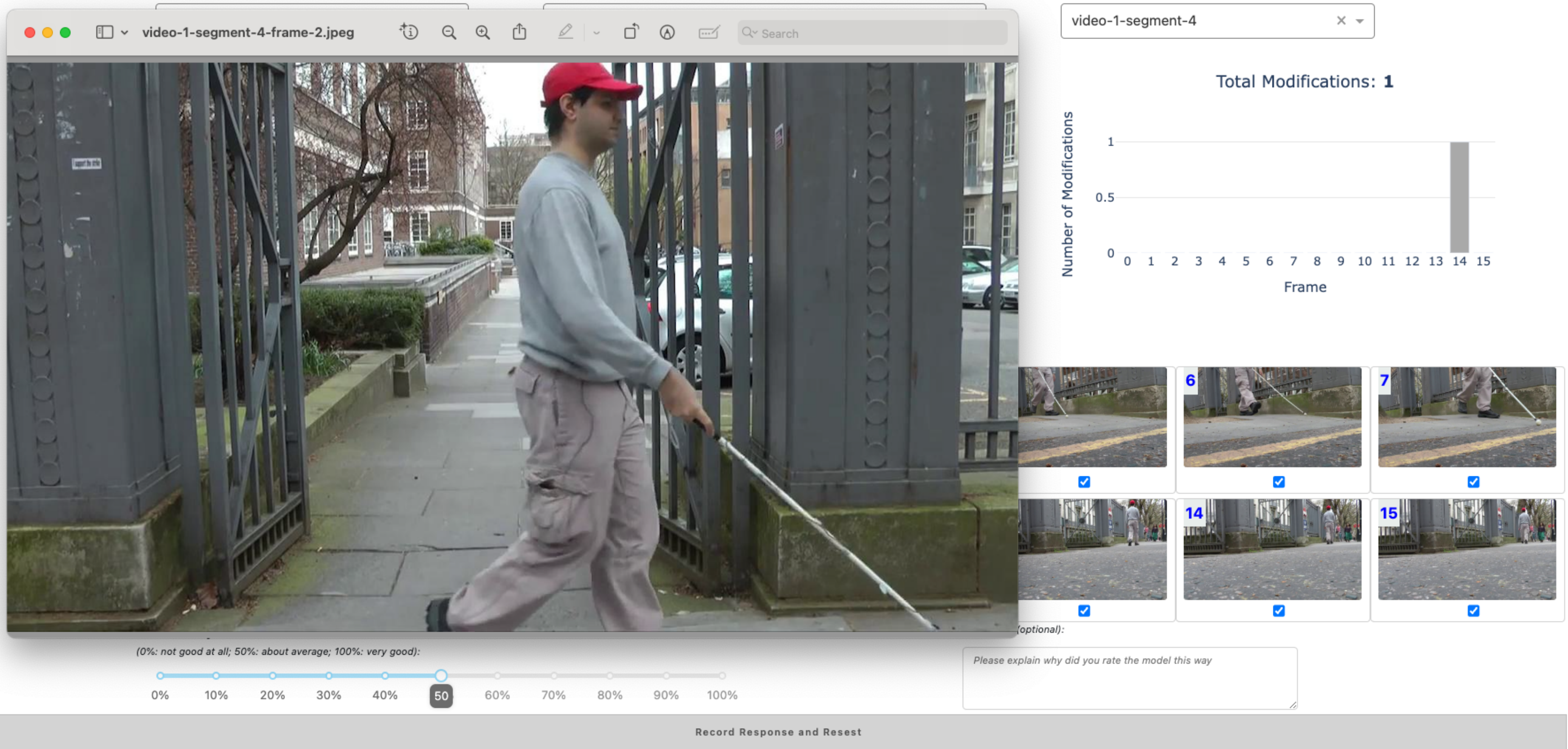}
    \caption{
   \sysname{}'s click-to-zoom feature in action. When a user clicks keyframe 2 (highlighted in Fig.~\ref{fig:dashboard_final}), the system opens an enlarged view in the operating system's default image viewer (shown here in MacOS \textit{Preview}). This external window allows users to inspect details, adjust magnification, and manipulate the view as needed for thorough analysis.
   }
    \label{fig:click-to-zoom}
\end{figure*}

When users select a video segment from dropdown C, the image container (D) displays up to 16 keyframes from that segment. 
Each keyframe shows a frame number (0 to N) in blue at the top-left corner. Users can click on any keyframe to view an enlarged version in their operating system's default image viewer, as shown in Fig.~\ref{fig:click-to-zoom}—a feature particularly helpful for users with low vision~\cite{islam2023spacex}. 
Checkboxes below each keyframe allow users to exclude or include frames based on quality criteria such as blurriness or poor camera angles. 
We limited the maximum number to 16 frames to prevent scrolling between components, which could create cognitive overload and impede understanding of the overall visualization~\cite{sanchez2009scroll}.

\subsubsection{\textbf{Object Selection Panel (\textbf{E})}}
\label{subsec:obj_selection}

It sits above the image container (see Fig.~\ref{fig:panel_e}). After examining the video keyframes, users decide which objects to test against the model. 
As users type object names, the dropdown suggests matches from our curated list of 90 objects (details in Sec.~\ref{sec:dataset}). Objects need not appear in every frame; users typically select objects visible across multiple frames.

\paragraph{Spy Objects} 
While detecting present objects matters, correctly identifying absent objects proves equally important. 
We introduce \textbf{\textit{`spy'}} objects as a form of adversarial probing, similar to how GAN architectures~\cite{goodfellow2014generative} challenge model discrimination through generative adversaries. 
These spy objects—such as \textit{Turnstile}, \textit{Snow}, \textit{Hose}, and \textit{Flush Door}—almost certainly do not appear in our evaluation dataset. 
Users add spy objects by prefixing names with '*', causing them to appear in \textcolor{violet}{\textbf{violet}} at the end of the selection list. In Fig.~\ref{fig:panel_e}, \emph{Chair} functions as a spy object.

Users can manage their object list by removing individual objects with the small cross icon beside each item or clearing the entire list with the large cross button next to the dropdown. 
To maintain visual clarity in the heat map, users can select up to 16 objects simultaneously, a limit established through pilot study feedback (Table~\ref{tab:round2_issues_solutions}, Row 4).

\subsubsection{\textbf{Model's Performance Summary: Binary Heat Map (\textbf{F})}}
\label{subsec:heatmap}

The binary heat map (F) presents a visual summary of model performance, with video frame numbers along the X-axis and selected objects along the Y-axis (see Fig.~\ref{fig:panel_f}). 
This visualization transforms complex model outputs into an easily interpretable pattern: \textcolor{lightgreen}{\textbf{green}} cells indicate objects the model recognizes in a frame, while \textcolor{red}{\textbf{red}} cells show objects it does not recognize.
In Fig.~\ref{fig:panel_f}, for example, \emph{Model-2} recognizes \emph{Vegetation} in \emph{Frame-2} but does not recognize \emph{Traffic Signals} in the same frame.

\begin{figure*}[!ht]
    \centering
    \begin{minipage}[t]{.48\linewidth}
        \centering
        \includegraphics[width=\linewidth]{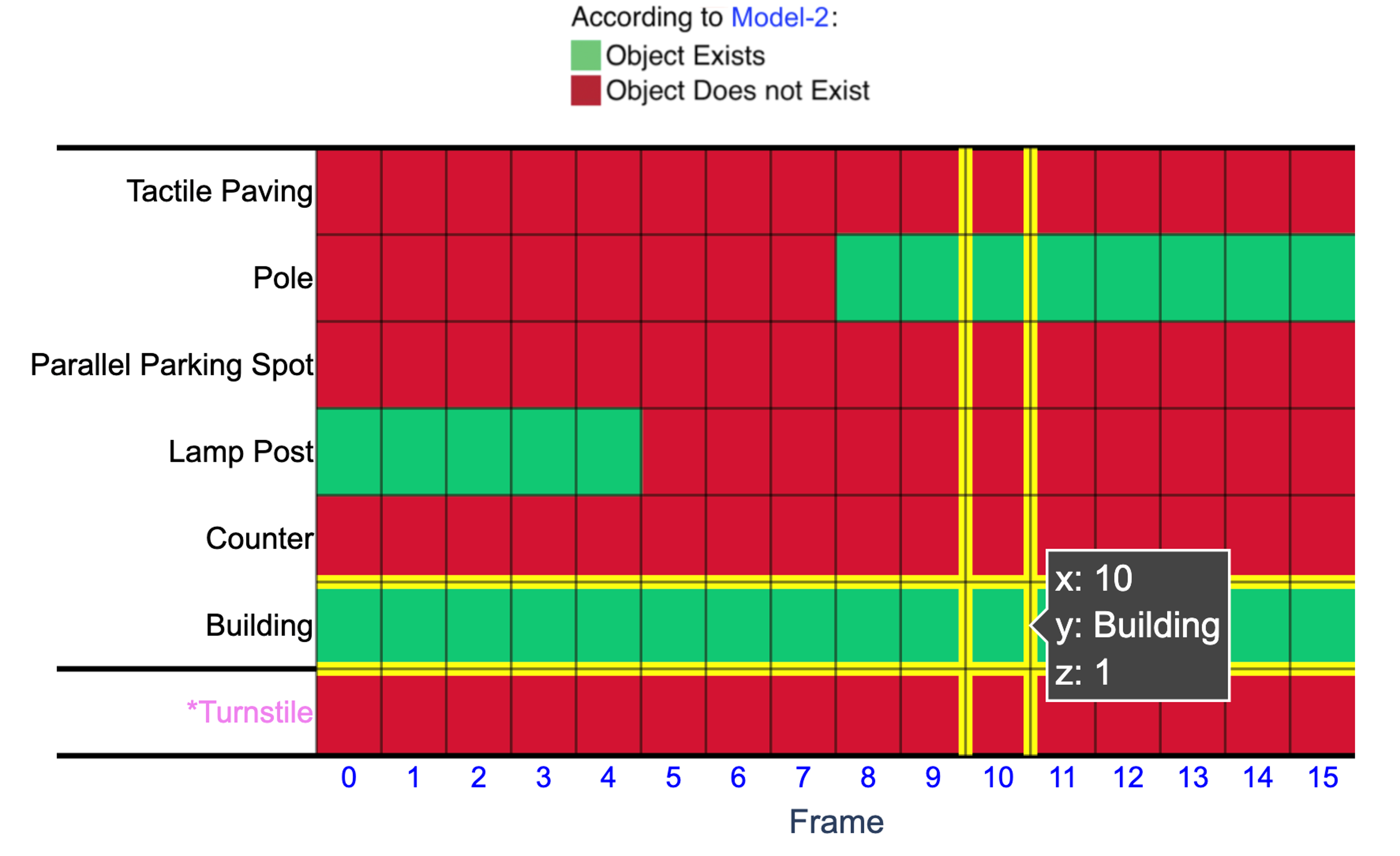}
        \subcaption{Default color codes in the heat map of \sysname{}. Green means the object exists, and red means the object does not exist.}
        \label{fig:non_colorblind}
    \end{minipage}
    \hfill
    \begin{minipage}[t]{.48\linewidth}
        \centering
        \includegraphics[width=\linewidth]{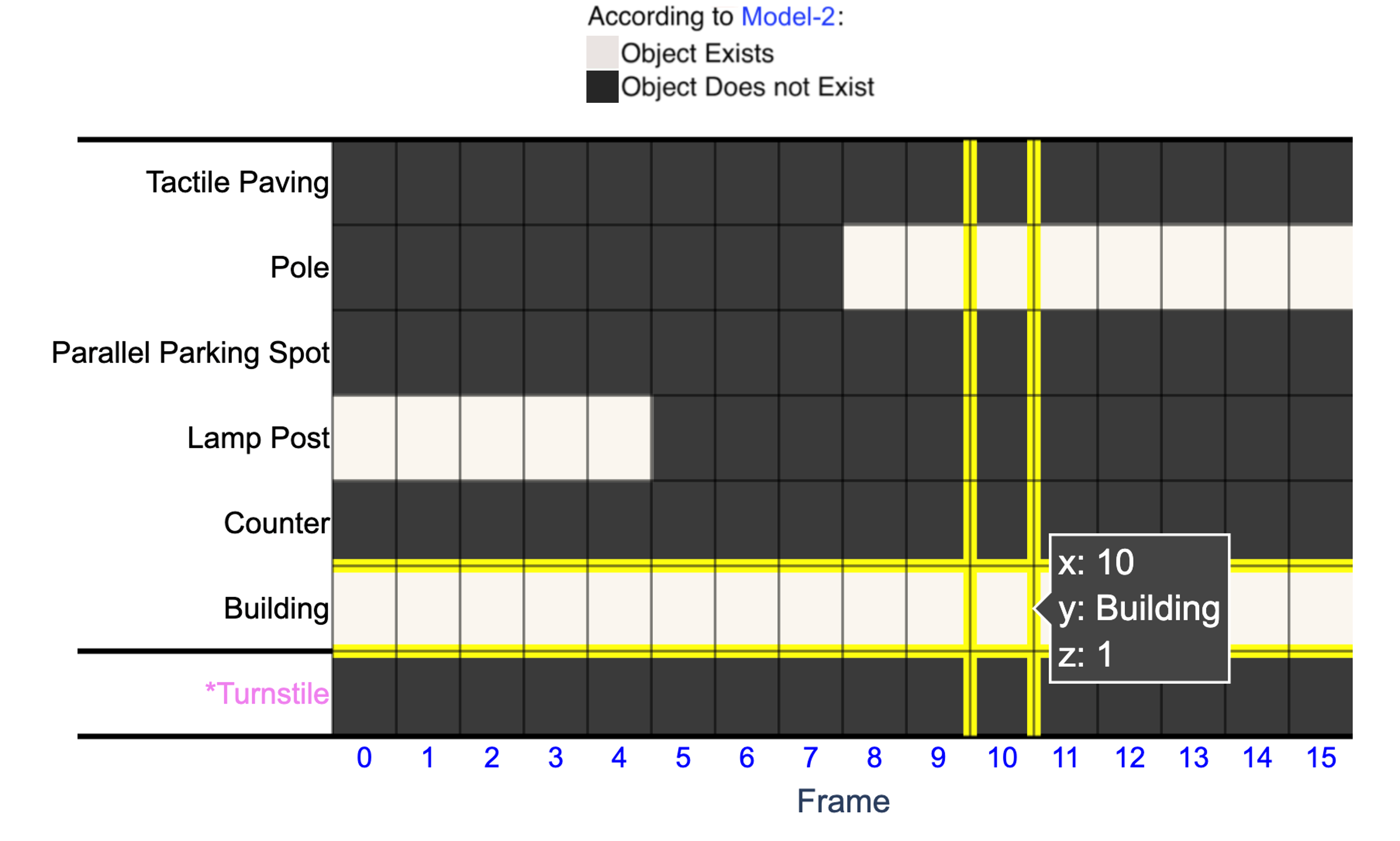}
        \subcaption{Accessible color codes, in the heat map of \sysname{}. White means the object exists, and black means the object does not exist.}
        \label{fig:colorblind}
    \end{minipage}
    \caption{Default and accessible color codes in the heat map of \sysname{}.}
    \label{fig:heatmap_color_codes}
\end{figure*}

For accessibility, we offer a colorblind mode that replaces green with white (light) and red with black (dark), as shown in Fig.~\ref{fig:colorblind}. 
Users can select their preferred color scheme within the interface.

Interactive features enhance the heat map's utility. Hovering over any cell highlights the corresponding frame in \textcolor{darkyellow}{\textbf{yellow}} throughout both the heat map and image container, allowing users to verify object presence against model predictions. Similarly, when users click a frame in the image container to examine it in detail, the system highlights the corresponding column in the heat map, maintaining a visual connection between the two components.

\subsubsection{\textbf{Modification Summary: Bar Graph (\textbf{G})}}
The heat map's colors indicate detection status rather than correctness—models can err by falsely recognizing absent objects or missing present ones. 
When users identify such errors, they can correct them by clicking cells to toggle between green and red. 
A supplementary bar graph (G) appears to the right of the heat map (see the top-right corner of Fig.~\ref{fig:click-to-zoom}), summarizing these user corrections by frame and helping users track their modifications to model outputs.

The modification feature serves two important purposes. First, it allows users to create cleaner visual patterns by eliminating distracting outliers, enabling more efficient scanning of the remaining heat map. 
Second, it provides explicit visual documentation of user interventions, helping users maintain awareness of their corrections when forming judgments about model performance. 
Making these corrections remains entirely optional—the feature exists to reduce cognitive load and support more effective pattern recognition during evaluation.

\subsubsection{\textbf{Rating Slider (\textbf{H}), Comments (\textbf{I}), and Reset Button (\textbf{J})}}
The final components include a Rating Slider (H), Comments Box (I), and Record and Reset Button (J). 
The slider lets users evaluate model performance from 0\% (completely random predictions) to 100\% (near-perfect accuracy) in 10\% increments. 
After rating, users may provide optional feedback in the comments box, highlighting significant trends or observations. 
Pressing the \emph{Record Response and Reset} button (J) saves the evaluation and prepares the system for the next assessment, whether with the same or a different model.

\subsection{\sysname{}'s Implementation Details}
\label{subsec:implementation_details}

\subsubsection{Notations}

Suppose there are $N_m$ available models,
\[
M = \{ M_1, M_2, \ldots, M_{N_m} \},
\]
for $N_t$ tasks,
\[
T = \{ T_1, T_2, \ldots, T_{N_t} \}.
\]
For each task $T_t$, there are $N_v$ representative videos available:
\[
V = \{ V_1, V_2, \ldots, V_{N_v} \},
\]
and each video $V_v$ contains a variable number of keyframes $N_f^{V_v}$.

$T_t$ is a task that involves recognizing multiple objects in the video. Let $\mathcal{O}$ be the domain of all objects $o$ present in any video in $V$, and let $N_o$ be the total number of objects in this domain. Note that $N_o$ can be countably infinite.

\begin{figure*}[!ht]
    \centering    
    \begin{minipage}[t]{.98\linewidth}
      \centering  \includegraphics[width=0.98\linewidth]{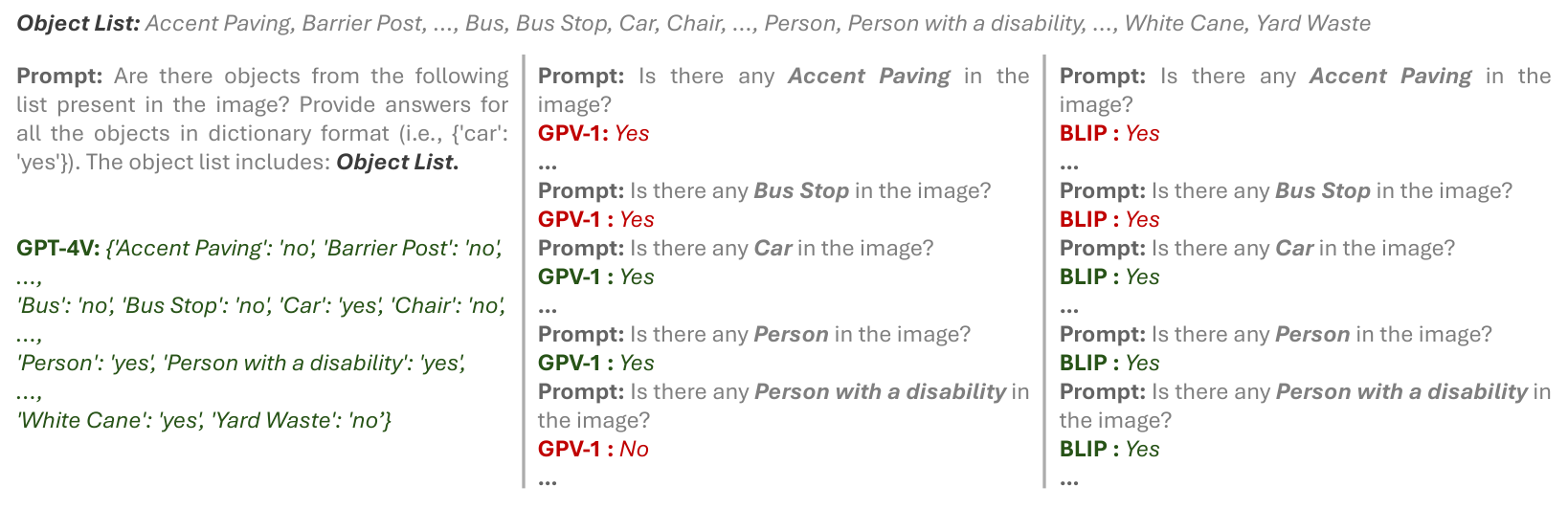} \\
    \end{minipage}
    \hspace{0pt}
    \vspace{0pt}
    \caption{
    Example prompts to \mgpt{} (left column), \mgp{} (center column), and \mbp{} (right column), and the model generated responses for the first frame in Fig.~\ref{fig:dashboard_final}.
    For one image, \mgpt{} was prompted once for a set of $N_o$ objects, and it responded with a dictionary, as shown in the left column.
    For one image, the other two models were prompted $N_o$ times, once for each object.
    Correct responses are in \textcolor{darkgreen}{\textbf{green}}, and incorrect ones are in \textcolor{darkred}{\textbf{red}}.
    }
    \label{fig:output_generation}
\end{figure*}

\subsubsection{Dataset Creation}
\label{sec:dataset}

Our dataset of video frames, key objects, and ground truth labels emerged from collaboration with blind individuals and careful analysis of navigation scenarios.

\paragraph{Background and Video Collection.}
Sighted companions of blind individuals—often without AI expertise—need effective ways to evaluate vision-language models intended for navigation assistance. 
While remote sighted assistance services like Aira~\cite{Aira2020} and Be My Eyes~\cite{BeMyEyes2020} connect users with human helpers, many blind individuals prefer smartphone applications that see the real world and provide real-time guidance, similar to NaviGPT~\cite{zhang2025enhancing}. 
For these AI-powered navigation tools to earn trust, companions must first assess whether the underlying models perform reliably enough for safe navigation.

We designed \sysname{} as a visual evaluation tool for sighted companions to assess model performance before blind users rely on these systems for navigation. Through discussions with blind collaborators, we considered collecting recordings of their daily navigation routes but identified unacceptable privacy risks in this approach. 
Following their advice, we instead examined content from \textit{YouTube} and \textit{Vimeo} where blind vloggers had publicly shared scripted navigation demonstrations, providing suitable evaluation materials without compromising privacy.

\paragraph{Key Object Identification.}
We identified 21 relevant videos from the two platforms (see Table~\ref{table:dataset} in Appendix~\ref{app:dataset}). 
Analyzing these videos, we compiled a list of objects crucial to blind and low-vision individuals' navigation. 
We then reviewed this list with members of the blind community, who helped narrow it down to 90 critical objects of interest (i.e., $|\mathcal{O}| = N_o = 90$).
These are the objects that appear in the object selection panel (\textbf{E}) of \sysname{} (Fig.~\ref{fig:dashboard_final} and Fig.~\ref{fig:panel_e}).

\paragraph{Ground Truth Labeling.}
We divided the 21 videos into smaller clips, called video segments, based on the appearance of navigation-relevant objects.
This resulted in 31 video segments (i.e., $N_v = |V| = 31$). 
These 31 video segments appear in the video dropdown (\textbf{C}) of 
\sysname{} (Fig.~\ref{fig:dashboard_final}).
Using the \textit{Katna} keyframe extraction tool\footnote{\url{https://katna.readthedocs.io/en/latest/}}, we further divided these video segments into keyframes.
We then manually labeled the presence of the 90 objects within each keyframe of these video segments, creating ground truth labeling.
Appendices~\ref{app:video-analysis} and ~\ref{app:gt-labeling} contain more details on the video segment creation, keyframe extraction, and ground truth labeling.

The object list, the video frames, and the ground truth labeling form our dataset. 
Our dataset is publicly available~\cite{islam2024identifying, islam2024dataset}\footnote{\href{https://github.com/Shohan29531/BLV-Road-Nav-Accessibility}{https://github.com/Shohan29531/BLV-Road-Nav-Accessibility}}.

\subsubsection{Supported Models and Their Output Generation.}
\label{sec:models}

The current \sysname{} server runs five models in the model dropdown (\textbf{A}) of \sysname{} (Fig.~\ref{fig:dashboard_final}), with a provision to add more as needed. 
The models are \textbf{\mgp{}}~\cite{gupta2022gpv}, \textbf{\mbp{}}~\cite{li2022lavis}, \textbf{\mgpt{}}~\cite{openai2023gpt4,openai2023gpt4vsc,openai2023gpt4vtwa}, \textbf{\mgt{}}, and \textbf{\mr{}}.
The \mr{} model makes predictions based on a coin toss, and the \mgt{} model contains our ground truth labeling (Sec.~\ref{sec:dataset}).

Among the other models, \textbf{\mgp{}}~\cite{gupta2022gpv} and \textbf{\mbp{}}~\cite{li2022lavis} run natively on our server machine.
For each video keyframe, an automated program prompted \mgp{} and \mbp{} 90 times, with one question per object (Fig.~\ref{fig:output_generation}).
On average, \mgp{} took $13.6$ seconds to answer the questions for a keyframe, while \mbp{} took $6.4$ seconds.
For \mgpt{}, we once prompted all 90 objects for a given keyframe, as shown in Fig.~\ref{fig:output_generation}.
\mgpt{} took an average of $27$ seconds per keyframe.
Note that these models' response times are still unsuitable for real-time interaction, as a response time of under 500 ms is required. 
As such, we pre-fetched the models' responses and served them from the cache in real time.

\subsubsection{Hardware.}
We employed \sysname{} using a client-server architecture. 
Our interface was implemented using Plotly Dash Python (v.2.14.2) and deployed on a server accessible via a private URL.
This server features a multi-threaded CPU (3.0 GHz, 16-core AMD EPYC), 128 GB of memory, and four NVIDIA RTX A6000 GPUs.

\begin{table*}[ht]
\centering
\caption{Visualization Frameworks Evaluated in Pilot Study 1 for \sysname{} and relevant participant feedback on each framework.}
\label{tab:visualization_feedback}
\begin{tabular}{p{3.5cm}|p{5cm}|p{7cm}}
\toprule
\rowcolor{gray!10} \begin{center}\textbf{Visualization Framework}\end{center} & \begin{center}\textbf{Description}\end{center} & \begin{center}\textbf{Challenges Identified by Participants} \end{center}\\ 
\toprule

Radial Layouts & Represent class relationships and confusion matrices, such as in Confusion Wheel~\cite{alsallakh2014visual}. & Feels visually cluttered with large datasets or many classes; participants struggled to interpret temporal relationships; circular design added unnecessary complexity. \\ 
\hline

\rowcolor{gray!10} Histogram Combinations & Display prediction scores across multiple classes, as used in Squares~\cite{ren2016squares}. & Ineffective for tracking temporal patterns; interpreting multiple histograms simultaneously added cognitive load. \\ 
\hline

Scatter Plot Summaries & Offer overviews of model outcomes and feature discriminations, as seen in Manifold~\cite{zhang2018manifold}. & Lacked temporal alignment; required expertise to interpret, limiting accessibility for non-technical users. \\ 
\hline

\rowcolor{gray!10} Temporal Confusion \newline Matrices & Extend traditional confusion matrices to track temporal dynamics, exemplified by ConfusionFlow~\cite{hinterreiter2020confusionflow}. & Useful for tracking aggregated class-level errors but unsuitable for task-specific evaluations; visual complexity was a drawback. \\ 
\hline

Multimodal Data Streams & Visualize real-time sensor data and AI outputs for AR applications, such as ARGUS~\cite{castelo2023argus}. & Overly complex; sole focus on AR applications; participants believed it would not align with our requirement of \sysname{}. \\ 
\hline

\rowcolor{gray!10} Binary Heat Maps & Use color-coded cells to represent object presence or absence in sequential video data, inspired by classical theories such as the Feature Integration Theory~\cite{treisman1980feature}. & While intuitive, simplicity might overlook subtleties like confidence score variations; still considered an advantage for non-technical users. \\ 
\bottomrule

\end{tabular}
\end{table*}

\section{Ideation and Design Evolution}

\sysname{} was developed using an iterative \textbf{Research-through-Design} (RtD)~\cite{research_through_design} methodology, combining technology-driven development with human-centered exploration. 
The development process involved three pilot studies that shaped the tool through brainstorming, prototyping, and iterative refinement. 

\subsection{Pilot Study Setup and Procedure}
\subsubsection{Participants}
All three pilot studies involved the same six participants. 
Four participants were experts in machine learning and computer vision with extensive research or industry experience, and two were non-experts with no prior experience in these fields.
Among the participants, five were male, and one was female; the average age was around 32.
All studies were IRB-approved.
Participants were recruited through mailing lists and by word of mouth. 
Participation in the studies was voluntary.
Two researchers conducted each study session---while one presented the design sketches and prototypes to the participants, the other facilitated discussions by asking questions and taking detailed notes. 
All participants attended the sessions simultaneously. 
Each study session lasted approximately two hours.

\subsubsection{Procedure}
The first pilot study focused on brainstorming and conceptualizing the most appropriate visualization for \sysname{}. 
Participants were introduced to various existing visualization frameworks, such as radial layouts~\cite{alsallakh2014visual}, histogram combinations~\cite{ren2016squares}, scatter plot summaries~\cite{zhang2018manifold}, temporal confusion matrices~\cite{hinterreiter2020confusionflow}, and binary heat maps.
Participants evaluated each framework for its intuitiveness, clarity, and suitability for temporal object recognition tasks without ground truth.
The second pilot study aimed to refine the initial \sysname{} prototype by identifying the usability issues and rooms for improvement.
The third pilot study focused on polishing the design and addressing the participants' desired changes after the second pilot study.

\begin{figure*}[!ht]
    \begin{minipage}[t]{.48\linewidth}
        \centering
        \includegraphics[width=\linewidth]{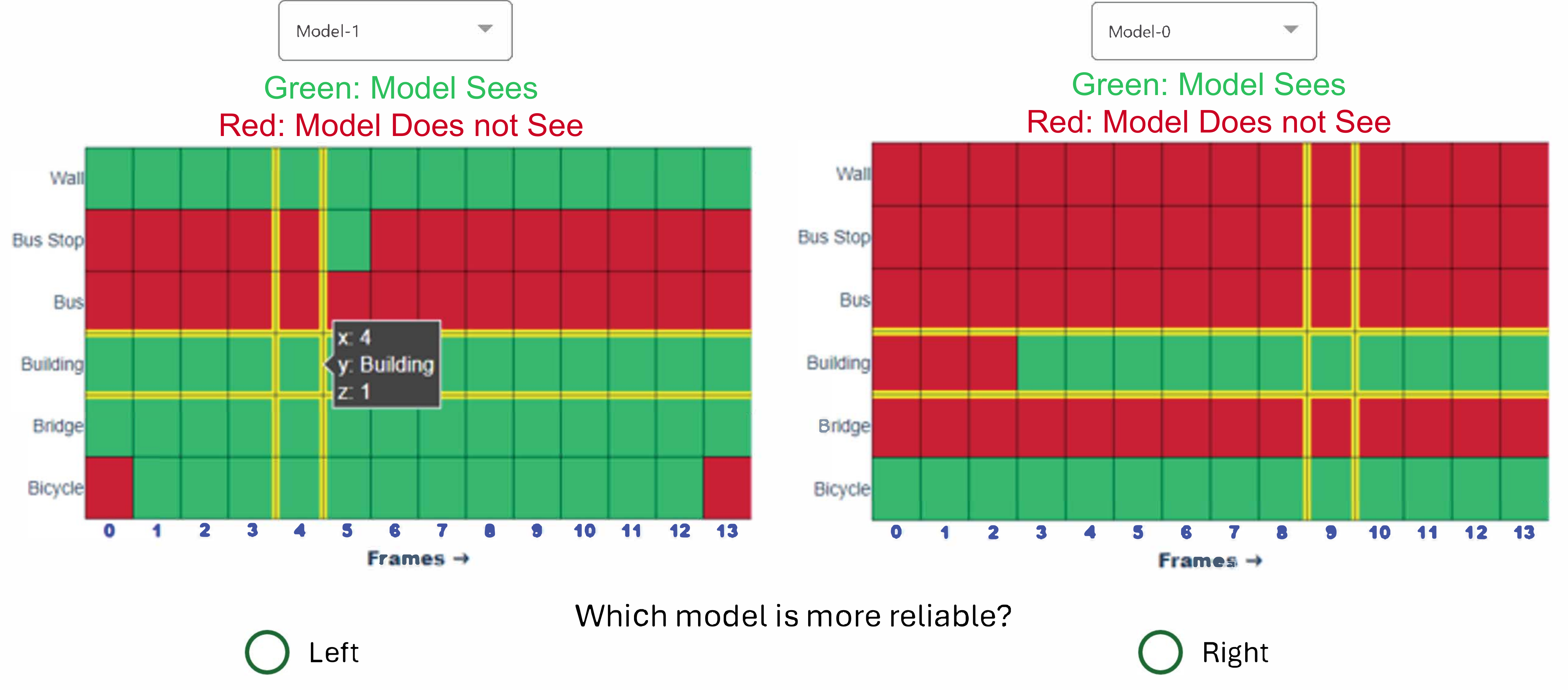}
        \caption{\textbf{Early design 1}: Two heat maps, two models, and the same objects. Users can pick two models and compare their heat maps for the same selected objects.}
        \label{fig:two-models}
    \end{minipage}
    \hfill
    \begin{minipage}[t]{.48\linewidth}
        \centering
        \includegraphics[width=\linewidth]{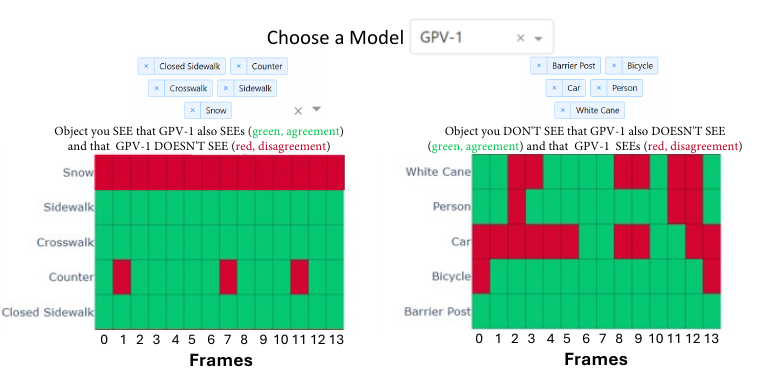}
        \caption{\textbf{Early Design 2}: Two heat maps for a single model—on the left map, users select objects they can see, while on the right map, they select objects they cannot see. The color meanings for red and green are reversed between the two heat maps.
        }
        \label{fig:see-dont-see}
    \end{minipage}
\end{figure*}

\subsection{Pilot Study 1: Choosing the Right Visualization Framework}
\label{subsec:first_pilot_study}
In the first pilot study, our primary objective was to determine the most suitable visualization framework for \sysname{}. 
At first, we provided participants with a clear explanation of the task---\textit{evaluating multi-object detection performance of vision language models in video data without ground truth}. 
We used numerous examples to make sure each participant understood the task.

Following this introduction, we presented the participants with \textbf{six} design sketches representing different visualization frameworks, as listed in Table~\ref{tab:visualization_feedback}.
Some sketches were hand-drawn; some were drawn using tools such as Microsoft PowerPoint and Zoom Whiteboard.
This study phase did not involve any functional prototypes; instead, the focus was on fostering open-ended discussions. 
Participants were encouraged to critically assess each candidate and highlight their feasibility, advantages, and drawbacks in our specific scenario. 
Table~\ref{tab:visualization_feedback} summarizes all the proposed frameworks, their descriptions, and key drawbacks, as discussed by the participants.

While each framework had its strengths and limitations, the choice ultimately narrowed down to two contenders: the binary heat map and temporal confusion matrices. 
Participants appreciated temporal confusion matrices (e.g., ConfusionFlow~\cite{hinterreiter2020confusionflow}) for their ability to provide a detailed, aggregated view of model performance over time. 
They noted that the structured representation of confusion metrics across temporal dimensions could provide insights into how models handle changes in object detection accuracy over time and across different objects in the task.
This design particularly appealed to participants with machine learning expertise, who valued its analytical depth. 
However, they also highlighted its drawbacks, particularly for non-expert users. 
The visual complexity of temporal confusion matrices and their reliance on aggregated metrics made it harder for users to focus on specific objects or interpret the results intuitively without additional training.

In contrast, the binary heatmap was unanimously praised for its simplicity and accessibility. 
Participants highlighted its ability to directly represent "Object Exists" (green) and "Object Does Not Exist" (red) states in the model's output, eliminating the need to interpret confidence scores or aggregated metrics.
While some participants acknowledged that the heatmap might lack the analytical depth of temporal confusion matrices--such as confidence metrics or aggregated class-level errors---they emphasized that its intuitive design was better suited for the specific context of \sysname{}. 
The task required non-expert users to quickly assess temporal patterns and anomalies, making simplicity a critical factor.

Ultimately, all participants agreed that the binary heatmap was the most feasible and user-friendly choice for this scenario.
This strong endorsement from participants motivated us to adopt the binary heat map design as the foundation of \sysname{}.

\subsection{Pilot Study 2: Refining the Heat Map Design}

Before the second pilot study, we designed two separate versions of \sysname{} with different roles for the heat map.

\textbf{First}, we experimented with comparing two models side by side on separate heat maps (Fig.~\ref{fig:two-models}). 
This design aimed to allow users to visually compare the models’ outputs for the same objects at corresponding keyframes. 
However, our participants in the second pilot study found this approach cognitively taxing, as they struggled to track and correlate cells across the two heat maps.

\textbf{Second}, we explored a design where users compared a single model’s performance on two sets of objects (Fig.~\ref{fig:see-dont-see}): 
\textbf{i)} those the user could see and \textbf{ii)} those they could not. 
We also introduced the concepts of “agreement” (green) and “disagreement” (red) to represent the model’s correctness with respect to the user's view. 
However, this design required participants to mentally reverse their interpretation of colors depending on the object set, confusing some participants.
Moreover, the terms “agreement” and “disagreement” did not sit well with some participants, who reported cognitive overload when trying to remember the meanings of the two terms, in addition to the colors.

To address these issues from the second pilot study, we simplified the final design to one heat map with one set of objects for a trial, with easy-to-interpret color coding: 
\textcolor{lightgreen}{\textbf{green}} when "Object Exists" and \textcolor{red}{\textbf{red}} when "Object Does Not Exist," according to the model (see Fig.~\ref{fig:dashboard_final}).

Participants also reported numerous other usability issues and provided recommendations for improvement. 
Table~\ref{tab:round2_issues_solutions} lists these issues and our implemented solution to address them before the next round.

\begin{table*}[ht]
\centering
\caption{Issues Identified in Pilot Study 2 and Their Corresponding Solutions Before Pilot Study 3.}
\label{tab:round2_issues_solutions}
\begin{tabular}{p{3.5cm}|p{6.5cm}|p{6cm}}
\toprule
\rowcolor{gray!10} \begin{center}\textbf{Issue} \end{center}& \begin{center}\textbf{Description}\end{center} & \begin{center}\textbf{Solution} \end{center}\\ 
\toprule

Hard to See Keyframes In the Image Container & Some participants mentioned that they could not see the video keyframes clearly within the image container (\textbf{D} in Fig.~\ref{fig:dashboard_final}) because they were too small. & \textbf{Details-on-Demand}: Clicking on a keyframe opened an enlarged view, allowing users to inspect specific frames without distraction (Fig.~\ref{fig:click-to-zoom}). \\ \hline

\rowcolor{gray!10} No Tracking of Changes in the Heat Map & The heat maps were interactable, but the changes were not tracked. Two participants mentioned that tracking the number of corrections in the heat map and showing where they made them would be helpful. & \textbf{Change Summary Bar Graph}: A bar graph (\textbf{G} in Fig.~\ref{fig:dashboard_final}) was added to track the number of toggled cells in each frame, helping users identify error trends. \\ \hline

Inaccessible Color Codes & One participant brought up the accessibility issue with the red-green color codes within the heat map, as users with color blindness would struggle to differentiate these colors. & \textbf{Use Accessible Color Codes}: We introduced a colorblind mode, replacing green and red colors with white and black, respectively. The stark contrast between the two new colors made it easier for users with color blindness to differentiate them (Fig.~\ref{fig:heatmap_color_codes}). \\ \hline

\rowcolor{gray!10} Excessive Objects in the Heat Map & The heat map accommodated as many objects as participants wanted, often resulting in too many cells to track or a distorted view. Three participants suggested limiting the maximum number of objects to 12–15. & \textbf{Maximum Objects in the Heat Map}: The number of objects in the heat map was capped at \textbf{16} to maintain usability and avoid distortion. \\ \hline

Provision for Spy Objects & One expert participant suggested that having some ``spy'' objects that are never present in any video frame may help users to filter out bad performing models quickly. & \textbf{Spy Objects:} We introduced the notion of ``spy'' objects (details available in Sec.~\ref{subsec:obj_selection}). \\
\bottomrule
\end{tabular}
\end{table*}

\subsection{Pilot Study 3: Final Testing}
After all the enhancements discussed in Table~\ref{tab:round2_issues_solutions}, we conducted the third round of the pilot study.
For this round,  participants provided overwhelmingly positive feedback. 
The enhancements were seen as intuitive and effective in enabling users to evaluate model performance quickly and accurately. 
No significant usability issues were reported at this stage, indicating that the design was ready for broader evaluation.

\section{Evaluation of \sysname{}}
To evaluate \sysname{}, we conducted a within-subject IRB-approved study with 15 sighted participants.
We now describe the study's hypotheses, conditions, trials, and results.

\subsection{Hypotheses}
We aimed to validate the following hypotheses:
\begin{itemize}
    \item[$H_1$:] \sysname{} will enable users to rate a model's reliability in a manner that correlates with the model's true performance (if available).
    % 
    % \item[$H_2$:] Participants' ratings for a model will be robust against statistical anomalies.
    % 
    \item[$H_2$:] Visual patterns generated by \sysname{} will assist users in making decisions more easily.
\end{itemize}

\subsection{Participants}

\begin{table}[!ht]
\caption{Participants' demographics, including their age group, gender, profession, and expertise level in Machine Learning/Computer Vision.} 
\begin{center}
\small{
\begin{tabular}{l C{2cm} C{2.25cm} C{2cm} l}
\toprule
 \rowcolor{gray!10} \textbf{ID} & 
 \textbf{Age Group/Gender} & \textbf{\textsc{Profession}} & \textbf{\textsc{Experience in ML/CV}} \\ 
 \toprule
 P1 & 25-29/M & Graduate Student   & Expert  \\ \hline
 \rowcolor{gray!10} 
 P2 & 30-34/M & Graduate Student & Non-Expert  \\ \hline
 P3 & 20-24/M & Graduate Student  & Non-Expert  \\ \hline
 \rowcolor{gray!10} 
 P4 & 25-29/M & Graduate Student  & Expert  \\ \hline 
 P5 & 20-24/M & Graduate Student & Expert  \\ \hline 
 \rowcolor{gray!10} 
 P6 & 25-29/M & Graduate Student & Non-Expert  \\ \hline
 P7 & 25-29/M & Graduate Student & Non-Expert  \\ \hline
  \rowcolor{gray!10} 
 P8 & 25-29/M & Graduate Student & Non-Expert  \\ \hline 
 P9 & 20-24/F & Graduate Student & Non-Expert  \\ \hline
  \rowcolor{gray!10} 
 P10 & 20-24/F & Graduate Student & Non-Expert\\ \hline
 P11 & 25-29/M & Graduate Student & Expert  \\ \hline
   \rowcolor{gray!10} 
 P12 & 40-44/M & Professor (Ph.D.) & Expert  \\ \hline
 P13 & 25-29/F & Graduate Student & Non-Expert  \\ \hline
  \rowcolor{gray!10} 
 P14 & 35-39/M & Professor (Ph.D.) & Expert  \\  \hline
 P15 & 35-39/M & R\&D Engineer (CV) & Expert  \\
 \bottomrule
\end{tabular}
}

\label{table:participants}
\end{center}
\end{table}

We recruited 15 sighted participants (12 males and 3 females) for the study (Table~\ref{table:participants}). 
The majority were graduate students (12), with a nearly even split between experts (7) and non-experts (8) in Machine Learning or Computer Vision. 
Participants were recruited through a combination of convenience sampling and word-of-mouth, primarily within the university community, departmental mailing lists, and personal networks, including individuals from other academic institutions and the industry.
Since our goal was to design a tool that both lay people and experts could use, we emphasized recruiting an equal number of participants from each group.
Non-expert participants might have taken machine learning courses but did not actively work in AI.
Experts were active in AI research or worked in the AI industry. 
For example, out of the seven experts, two were professors with Ph.D. degrees, one was an R\&D engineer in Computer Vision working in the industry,
and others had at least a publication in mainstream AI conferences (e.g., \texttt{AAAI} and \texttt{CVPR}).
In summary, all expert participants were actively involved in AI research and had relevant publications to support their expertise.

\subsection{The Task}
The task was to rate a model $M_m \in M$ for a particular video $V_v \in V$ by selecting a subset of objects \objsub{} $\subseteq$ \objs{} using \sysname{}. 

\subsection{Model's Underlying Performance Metrics}
\label{sec:f1}

%\subsubsection{\textbf{Performance Metric: $F_1$}}
We utilized the $F_1$-score as the metric for evaluating a model's underlying performance. 
It is essential to note that this metric was concealed from users, who only saw the model's predictions in the heat map.
Despite some criticism, we chose the $F_1$-score because it is the de facto metric for reporting the performance of object recognition models in machine learning. 
One such criticism is that it gives equal importance to precision and recall~\cite{hand2018note}, which may not always be desirable. 
Another is that it is sensitive to changes in class distribution in multi-class problems~\cite{powers2020evaluation}.

However, in our study, we intentionally wanted precision and recall to be equally important, as both false positives and false negatives are equally undesirable in our scenario. Therefore, the first criticism was not a concern for us. 
Additionally, since the objects in our dataset are all relevant to a specific task (blind navigation assistance, see Sec.~\ref{sec:dataset}), and we used micro averaging for aggregating scores of different classes, we do not face the issue of sensitivity to class distribution.

\begin{figure*}[!ht]
\centering
    \begin{minipage}[b]{0.43\linewidth}
        \centering
        \includegraphics[height=0.5\linewidth]{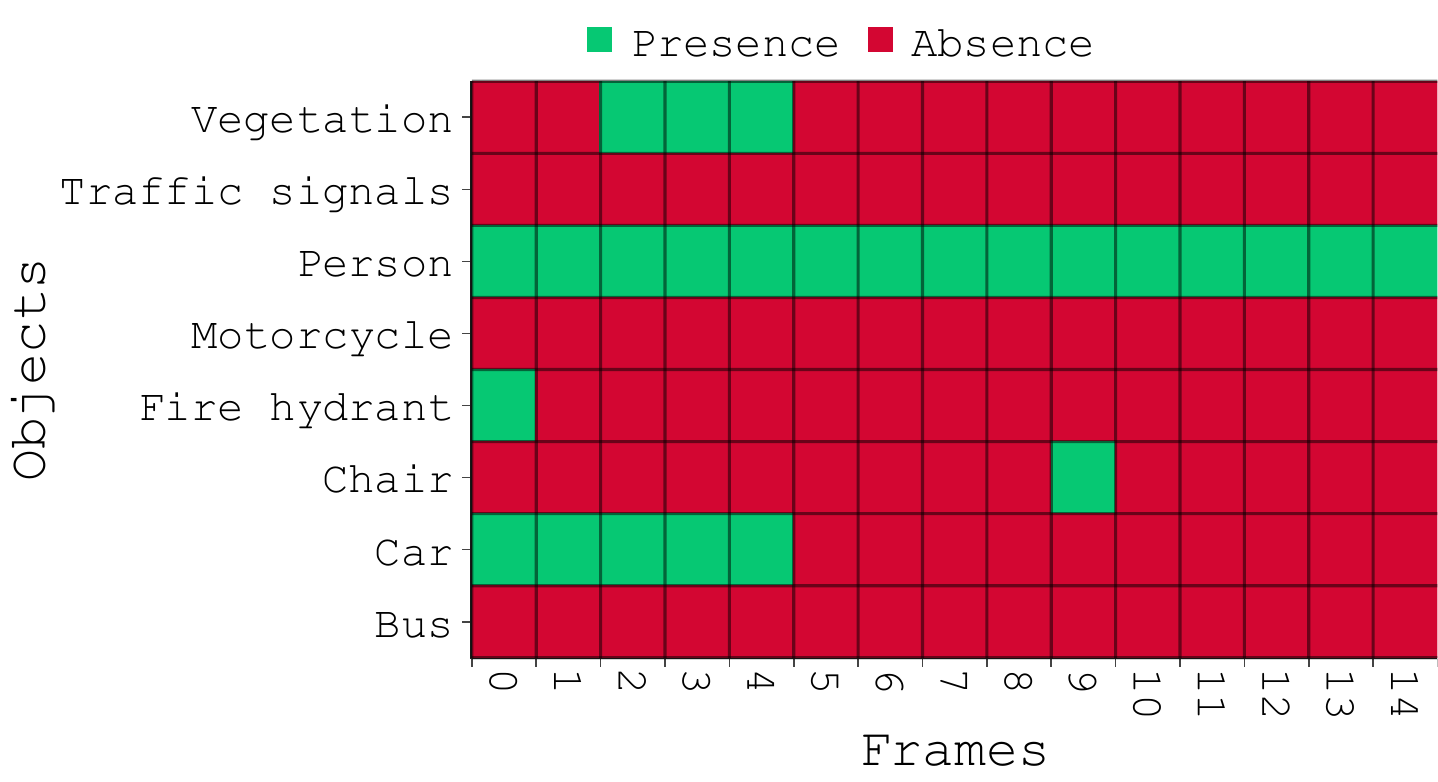}\\
        \small(a)~Raw output from the \mbp{} model.
    \end{minipage}
    \hspace{10pt}
    \begin{minipage}[b]{0.43\linewidth}
        \centering
        \includegraphics[height=0.5\linewidth]{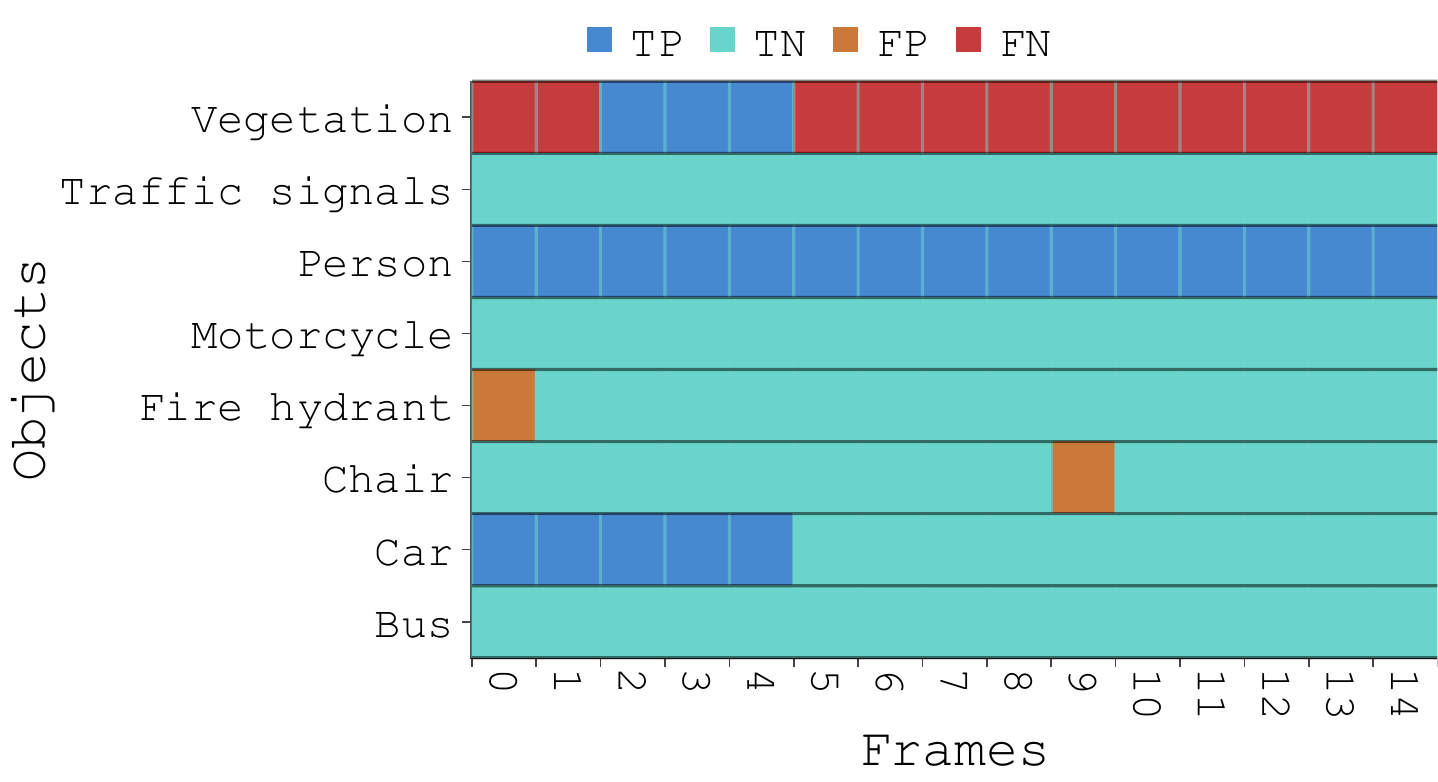}\\
        \small(b)~\mbp{} output evaluated with ground truth.
    \end{minipage}
   
    \caption{\mbp{} models' outputs for the first 15 frames from Fig.~\ref{fig:dashboard_final}.
    The left heat map (in red and green) displays the models' raw outputs.
    The right heat map (b) shows the performance of \mbp{} for the given scenario when evaluated against ground truth data.
    The right heat map's colors (best viewed in color) indicate true positives (TP: dark blue), true negatives (TN: teal), false positives (FP: orange), and false negatives (FN: brick red). 
    With TP, FP, TN, and FN known, we can calculate metrics such as $F_1$, Precision, and Recall. 
    The \fonelocal{} score here is 0.77.
    }
\label{fig:org_v_shadow}
\end{figure*}

\subsubsection{\textbf{Performance Metric: \foneglobal{}}}
\label{sec:f1global}
It is worth noting that the models in our study are open vocabulary but made predictions on our dataset containing \objs{} objects.
These predictions are compared against the ground truth to compute the $F_1$-score.
We use a special notation, \foneglobal{}, to report a model's $F_1$-score on our entire dataset (\objs{}).

\subsubsection{\textbf{Performance Metric: \fonelocal{}}}
\label{sec:f1local}
To gain a more fine-grained measure of a model's $F_1$-score on the specific objects an individual used during the study, we employed another notation, \fonelocal{}, which reports the model's $F_1$-score only on the subset of selected objects (\objsub{}).
Calculation of \fonelocal{} is demonstrated in Fig.~\ref{fig:org_v_shadow}.

\subsection{Study Conditions and Trials}
\label{subsec:study-conditions}
\subsubsection{\textbf{Conditions for hypothesis $H_1$}}
\label{sec:main-models}

We had five study conditions for testing $H_1$, each representing a model. 
These models included two baselines ($M_1$ and $M_2$) and three LMMs ($M_3$, $M_4$, and $M_5$) as follows:

\begin{itemize}
    \item[$M_1$] \textbf{\mr{} model}: This serves as our \textit{baseline} for the \textit{worst-performing} model. This model flips a fair coin to predict whether an object exists in a keyframe. If the coin lands on heads, it outputs yes; otherwise, it outputs no.
    \item[$M_{2}$] \textbf{\mgt{} model}: This serves as our \textit{baseline} for the \textit{highest-performing} model. This model uses the ground truth annotations (Sec.~\ref{sec:dataset}) to predict whether an object exists in a keyframe. Note that this model serves as the oracle, which is not applicable to real-world tasks. We only used it to rigorously test how user ratings are affected if they notice the output of an oracle.
\item[$M_{3}$] \textbf{\mgp{} model}: A general-purpose, open vocabulary, vision-language model~\cite{gupta2022gpv}.
\item[$M_{4}$] \textbf{\mbp{} model}: Another open vocabulary model for unified vision-language understanding and caption/description generation~\cite{li2022blip}.
\item[$M_{5}$] \textbf{\mgpt{} model}: This is one of the most popular, open vocabulary, commercial LMMs~\cite{openai2023gpt4,openai2023gpt4vsc,openai2023gpt4vtwa}.
\end{itemize}

\subsubsection{\textbf{Trials}}
Each participant rated 5 video segments (i.e., trials) for each condition and recorded a total of 25 ratings ($=5 \times 5$).
We counterbalanced the conditions and the videos using a Latin Square.
Combining all participants, we collected a total of 375 user ratings ($=15 \times 25$).

\subsection{Study Procedure}
\subsubsection{Setup.}
Except for two participants (P12 and P14), all sessions were conducted in person in a quiet room.
The interface of \sysname{} ran on a study computer, an M1 Pro 16-inch MacBook with 16 GB of RAM and a screen resolution of 3456 × 2234.
P12 and P14 interacted with the study computer via Zoom teleconferencing software's remote control and screen-sharing features.
Two researchers conducted each study session -- one facilitated the study, guiding participants through trials, while the other observed closely, took notes, and monitored participants' interaction with \sysname{}.

\subsubsection{Procedure}
We began each session by obtaining consent and collecting participants' demographics and experience in AI research.
We then discussed the potential of LMMs in critical everyday tasks such as blind navigation assistance, medical diagnosis, and autonomous driving. 
However, we emphasized that these models must perform at a very high level of accuracy and reliability to be used effectively in such scenarios.
Next, we provided an in-depth demonstration of \sysname{}, explaining its various components, functionalities, and how to assess the model's performance.
Participants then interacted with the system using a dummy model and a non-study video until they felt confident in its use. This process took less than 5 minutes on average.

Next, we provided the participants with a model ID (e.g., model 3 from the model dropdown, A in Fig.~\ref{fig:dashboard_final}) and a video segment ID (e.g., video-1 segment-4 from the video dropdown, C in Fig.~\ref{fig:dashboard_final}). 
We asked them to rate the model's performance for that specific video segment. 
\textit{Note that model IDs were randomly initialized for each participant, and they did not know the name of the underlying model.}
Halfway through the study, we inquired about any challenges they experienced related to the system or in decision-making, as well as their usage patterns. 
For instance, if a participant was correcting all model mistakes by clicking on the corresponding cell on the heat map, we reminded them that this action did not actually improve the model's performance; it merely overrode the model's mistakes for that specific instance.
After each trial, participants rated the model’s reliability in that video using a slider ranging from 0\% (not reliable) to 100\% (highly reliable), selectable in 10-point intervals (e.g., 20\%, 30\%, 70\%), functioning as a discrete Likert-like scale.

Following the final trial, we requested detailed feedback about their experience, including the system's usability. 
We also asked them to elaborate on their decision-making process and what positively or negatively influenced their ratings.
Finally, we asked them to complete the NASA-TLX questionnaire to assess their perceived workload during the study.
Each session lasted approximately 90-100 minutes, and participants were compensated with a \$25-Amazon gift card for their time and effort.

\subsection{Data Logging and Analysis}
\label{subsec:data_collection_analysis}

With the participants' consent, we recorded the screen and all conversations for post-processing and analysis.
Our system automatically logged user ratings, \fonelocal{} scores, and participant comments.
Additionally, it included an internal logger that recorded participants' cursor movements, clicks, and the objects they clicked.
Two researchers manually reviewed the screen recordings, transcribed the conversations, and cross-checked the task completion times using video data, their notes, and internal click logs.
They also analyzed which visual patterns required more (or less) time for participants to rate, the comments made after seeing a pattern, and how participants hovered their cursor over the heat map.

\subsubsection{Normalization of Ratings}
\label{subsec:normalization}

Subjective ratings are usually prone to individuals' biases~\cite{jin2004study}. For example, some participants rated generously, while others confined their ratings within a narrow range, and some others rated on a broader spectrum.
To remove these individual biases, we applied mean-centering~\cite{howe2008re} for each participant, followed by Min-Max normalization among all participants to keep the ratings between 0 and 1 for ease of interpretability.

\subsubsection{Statistical Tests}
\label{subsec:stat_tests}
We first used the Shapiro–Wilk test to determine whether the study data (e.g., ratings and completion times) were normally distributed.
The test confirmed that user ratings were not normally distributed.
Therefore, we used non-parametric tests. Specifically, we employed the Kruskal–Wallis test to assess whether the ratings were statistically different for $H_1$ conditions.
% Additionally, we used the Mann–Whitney test to investigate whether the ratings of a model and its shadow were significantly different.

\section{Results}
\label{sec:results}
In this section, we discuss the major results of our study. 
\textbf{R}esults which are crucial are marked using the notation $\mathbf{R_{x}}$, with $\mathbf{x}={\{1, 2, 3, ...\}}$

\subsection{Users' Ratings Correlate with Models' Underlying Performance}
\label{subsec:all_org_model_ratings}

We consider two metrics -- \foneglobal{} (Sec.~\ref{sec:f1global}) and \fonelocal{} (Sec.~\ref{sec:f1local}) -- to measure the model's underlying performance.

\begin{figure*}[!ht]
    \centering
    \begin{minipage}{0.48\linewidth}
        \centering
        \includegraphics[width=\linewidth]{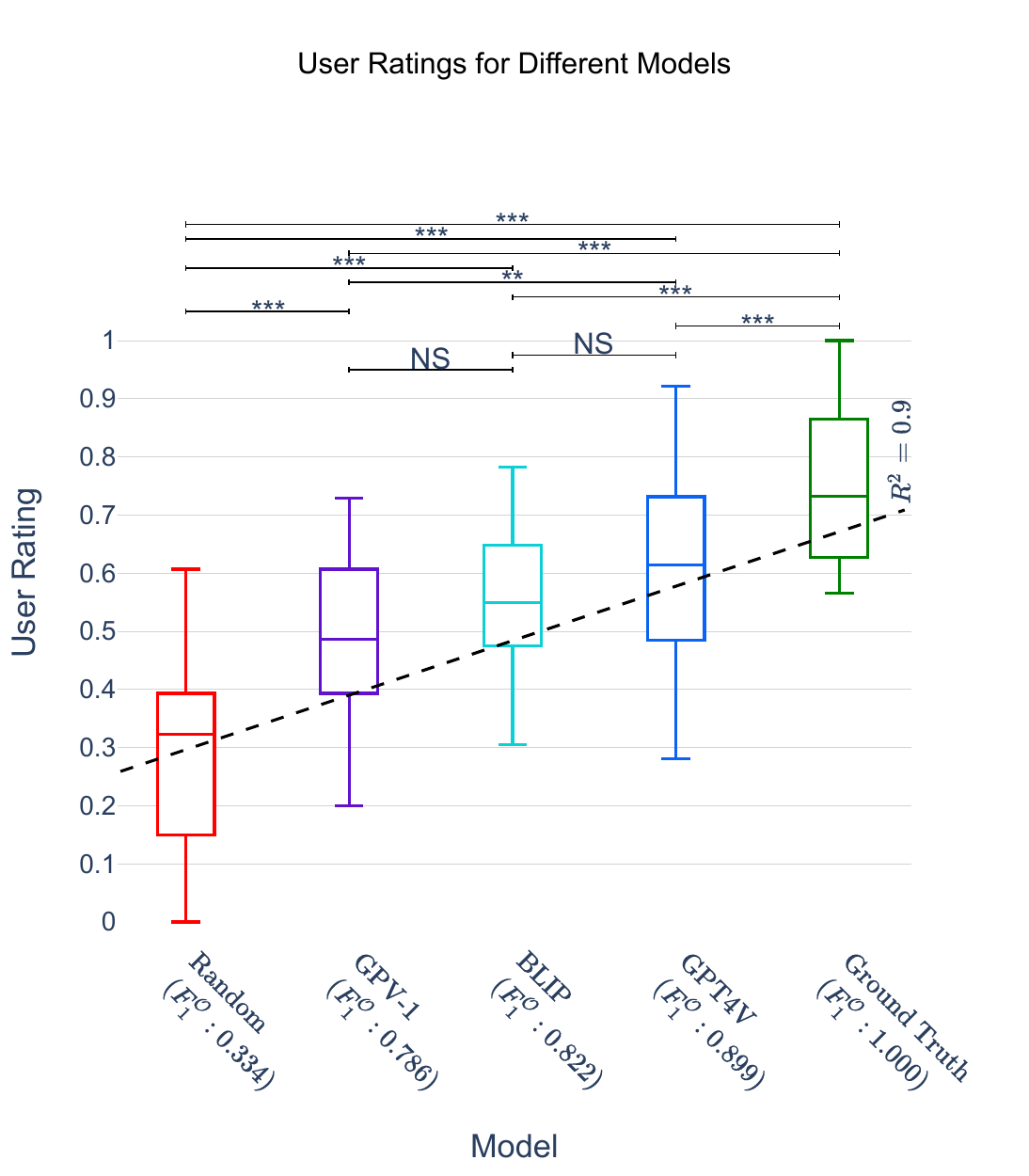} 
        \caption{Boxplots of normalized user ratings (higher is better) for five models, including the random model (leftmost) and the ground truth model (rightmost).
        The models are sorted on the x-axis based on their \foneglobal{}-scores (higher is better).}
        \label{fig:all_org_models}
    \end{minipage}
    \hspace{10pt}
    \begin{minipage}{0.48\linewidth}
        % \centering
        % \includegraphics[width=\linewidth]{figures/all_shadow_model_scores.pdf}
        % \caption{Boxplots of normalized user ratings for three pairs, where each pair is made up of a model and its shadow. Note that the \foneglobal{} score of a model and its shadow model is the same.
        % % , but their \fonelocal{} scores are different due to randomness in sampling.
        % }
        % \label{fig:all_shadow_models}

        \centering
        \includegraphics[width=\linewidth]{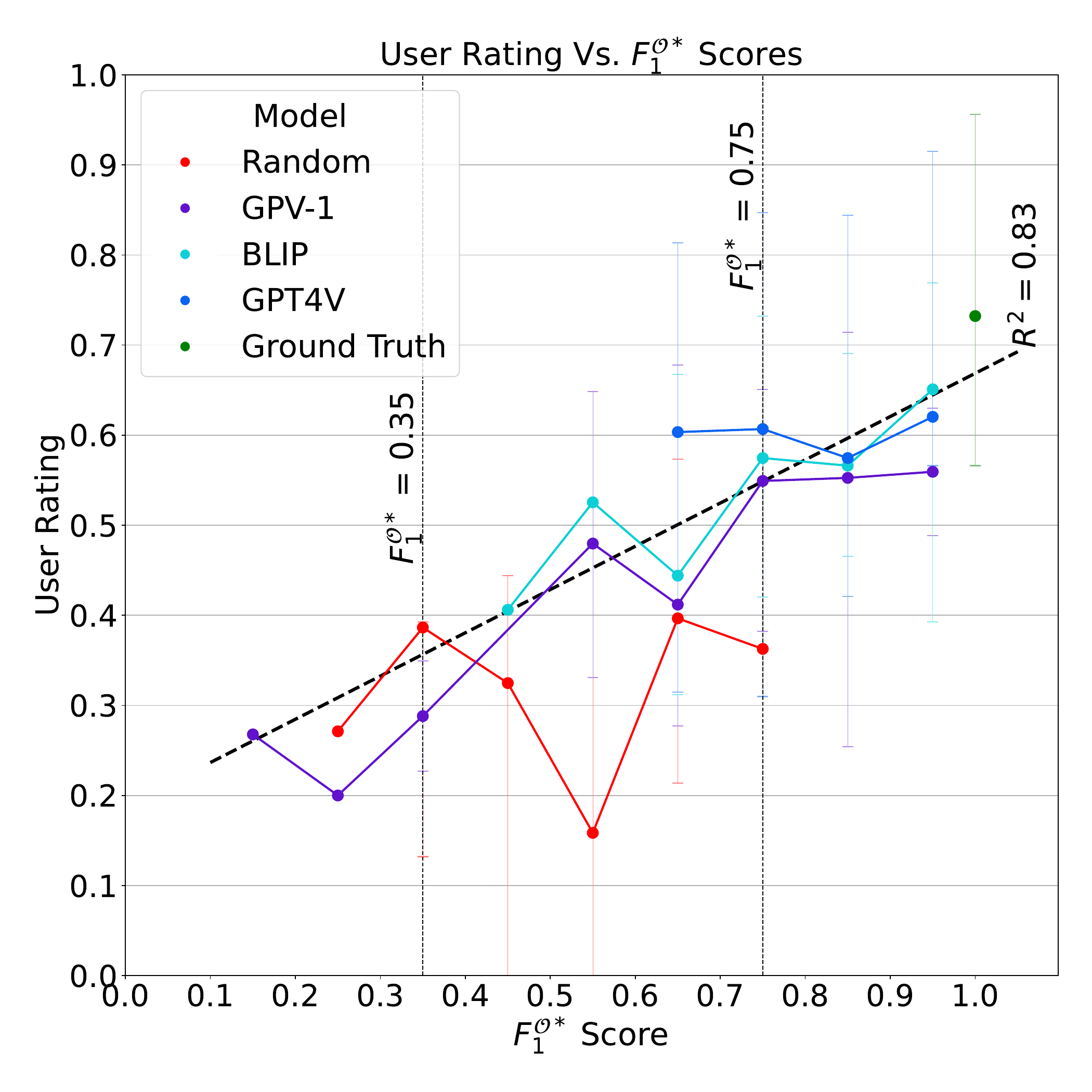}
        \caption{Normalized user ratings plotted against different values of \fonelocal{}.
        %between $0$ and $1$ with an interval of $0.1$.
        Each dot represents the median user rating within the range of \fonelocal{}. 
        The range of \fonelocal{} for each model was different,
        with the \mr{} models never crossing 0.7.
        This explains the different numbers of median dots for each model.
        % The ratings for the random models are shown in red. 
        Note that the ground truth model appears as a point at the top-right corner since \fonelocal{}-scores for ground truth models are always 1.0. 
        The regression line fits all models except the Random.
        }
        \label{fig:dev_vs_F1}
    \end{minipage}

\end{figure*}

\subsubsection{Users' Ratings Correlate with Models' \foneglobal{} }
\label{subsec:rating_vs_global_f1}
Recall that \foneglobal{} is the model's performance on the entire dataset containing all objects (\objs{}).
The most striking results (R1 to R3) about the participants' ratings and the models' \foneglobal{} (which was hidden from the users) are as follows:
\begin{itemize}
    \item[$\mathbf{R_1}$] By merely observing the patterns generated on the heat map, participants were able to recognize the \mr{} models and the Ground Truth (\mgt{}) models. We elaborate further on these patterns in the following section.
    \item[$\mathbf{R_2}$] Participants consistently rated the \mr{} models as the lowest (median rating: $0.32$) and \mgt{}s as the highest (median rating: $0.73$), as shown in the leftmost and the rightmost box plots in Fig.~\ref{fig:all_org_models}.
    \item[$\mathbf{R_3}$] For non-random and non-GT models, such as $M_3$:\mgp{}, $M_4$:\mbp{}, and $M_5$:\mgpt{}, participants' ratings strongly and positively correlated with these models' \foneglobal{} ($R^2=0.90$), as shown by the diagonal line in Fig.~\ref{fig:all_org_models}.
\end{itemize}

Fig.~\ref{fig:all_org_models} shows the box plots of users' ratings for all 5 models (along the y-axis), sorted by their \foneglobal{} (along the x-axis).
A Kruskal-Wallis test across the five groups confirmed that their median ratings are statistically significantly different (H: $100.7$, $p \approx 0 $).
Tukey's post-hoc HSD test with Bonferroni Correction reveals that, except for two pairs, \mbp{} vs.\  \mgp{} and \mbp{} vs.\  \mgpt{}, the median ratings of all pairs are statistically different.

These suggest that the best overall model for this case is \mgpt{}, 
% according to Eq.~\ref{eq:1}, 
because the \mgt{} model, even though yielding the highest overall user rating, is not available in real-world tasks.
A byproduct of our results is that one can expect a similar performance by trading \mgpt{} with \mbp{}, a $4.5 \times$ smaller model (362M  vs.\ 1.7T params), as the median user ratings for these two models are not statistically different.

\subsubsection{Users' Ratings Also Correlate with Models' \fonelocal{}}
\label{subsec:rating_vs_local_f1}

Recall that \fonelocal{} reports the model's $F_1$-score calculated only on the objects selected by the participants during a trial.
In Fig.~\ref{fig:dev_vs_F1}, we plotted different \fonelocal{} scores from various trials on the x-axis and users' ratings for those trials on the y-axis.
We summarize the most interesting results from this graph as follows:

\begin{itemize}
    \item[$\mathbf{R_4}$] Another striking result is that user ratings \textbf{do not} correlate with the increasing \fonelocal{} scores of the random models ($R^2=0.04$, red line in Fig.~\ref{fig:dev_vs_F1}). 
    This is because the higher \fonelocal{} scores of the random models were due to sampling issues that occurred by chance. It strongly suggests that users' ratings are resilient to a model's randomness.
    \item[$\mathbf{R_5}$] Like \foneglobal{}, user ratings are correlated with all non-random models' \fonelocal{} scores ($R^2=0.83$).
\end{itemize}

Thus, our results (R1 to R5) validate hypothesis $H_1$. \hfill\qedsymbol{}

\subsection{Visual Patterns Affect Participants' Decisions}
\label{subsec:role_of_patterns}

% \begin{figure}[!ht]
%         \centering
%         \includegraphics[width=0.4\linewidth]{figures/8_patterns.pdf}
%         \caption{Different predominant visual patterns that affect participants' decisions.}
%         \label{fig:8_patterns}
% \end{figure}

% \begin{figure}[!ht]
%         \centering
%         \includegraphics[width=0.5\linewidth]{figures/examples_of_patterns.pdf}
%         \caption{Illustrations of some predominant visual patterns marked in the heat map.}
%         \label{fig:all_patterns}
% \end{figure}

\begin{figure}[!ht]
    \centering
    \includegraphics[width=0.8\linewidth]{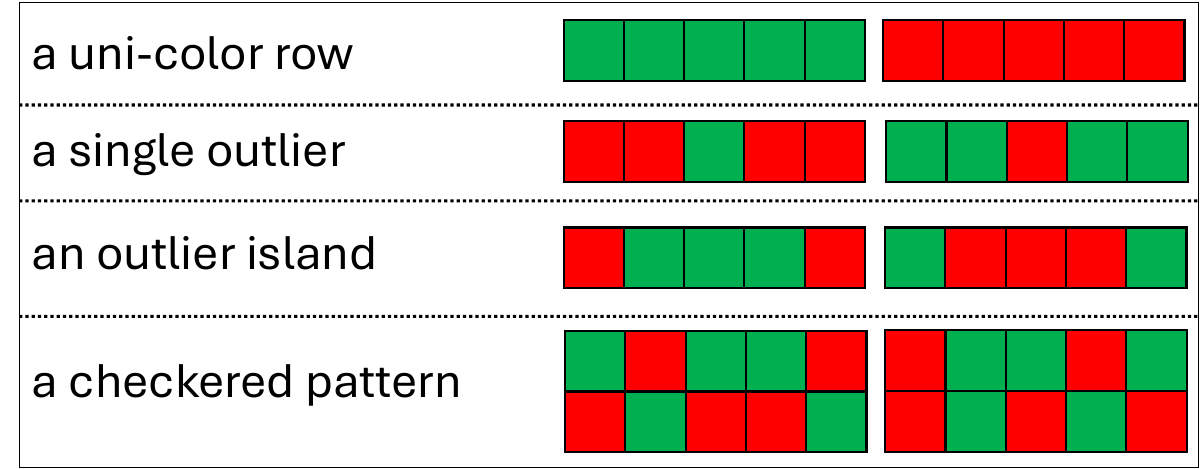}
    \caption{Different predominant visual patterns that affect participants' decisions.}
    \label{fig:8_patterns}
\end{figure}

\begin{figure}[!ht]
    \centering
    \includegraphics[width=\linewidth]{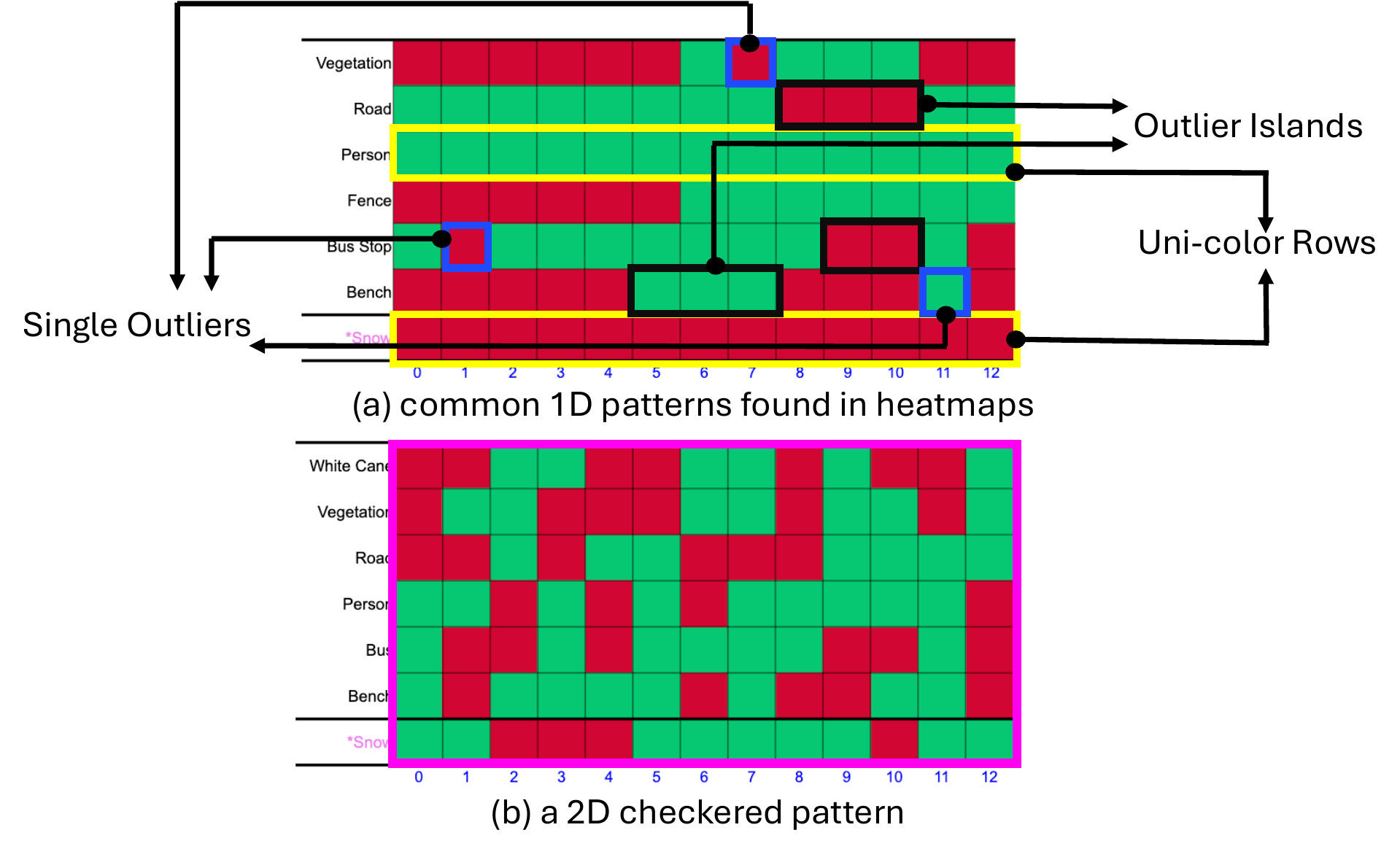}
    \caption{Illustrations of some predominant visual patterns marked in the heat map.}
    \label{fig:all_patterns}
\end{figure}

We observed several recurrent patterns in the heat map and analyzed how participants reacted to these patterns. The most notable patterns include uni-color rows, single outlier cells, outlier islands, and checkered-like patterns (as seen in Fig.\ref{fig:8_patterns} and illustrated in Fig.\ref{fig:all_patterns}).

\subsubsection{Uni-color Rows}
This pattern is characterized by rows that consist entirely of green or red cells, as shown in yellow in Fig.~\ref{fig:all_patterns}.a. 
Such homogeneity in prediction outcomes fosters a positive perception of the model's performance among participants.
Our observations revealed that participants often do not meticulously examine each cell within these homogeneously colored rows. Instead, they validate the model's performance by reviewing only the first one or two cells of each row.
Moreover, when these rows contained minor errors—specifically, one or two cells colored differently from the rest—some participants were inclined to disregard these anomalies. This reveals a threshold of error tolerance influenced by the prevailing pattern of correctness or incorrectness within a given row.

\subsubsection{Single Outliers}
An outlier is a single cell surrounded by cells of different colors on both the left and right, as shown in blue in Fig.~\ref{fig:all_patterns}.a. These cells drew attention from all participants throughout the study.
The typical response from the participants was to check the correctness of that specific outlier cell. If an outlier cell was determined to be correct, it significantly boosted the participants' confidence in the model. According to P4:

\begin{quotation}
    \emph{``I am inclined to check these outliers constantly. It boosts my confidence more than seeing correct predictions in other places.''}
\end{quotation}

\subsubsection{Outlier Islands}
An outlier island is a group of outlier cells, marked in black in Fig.~\ref{fig:all_patterns}.a.
When stumbling upon an island, participants usually look at the frames immediately before and after the island to see what specifically changed. Some participants also examined the frames within the island itself.
However, when an island spanned more than five cells, participants usually reviewed only the first one or two frames of the island, as they did in uni-color rows. 
This behavior suggests that the frames immediately preceding and following the island are more influential than those within the island in shaping participants' decisions about the model.

\subsubsection{Checkered-like Patterns}
\label{subsec:checkered_pattern}

These are 2D patterns with frequent color changes between green and red across frames, creating a checkerboard-like effect in the heat map (Fig.~\ref{fig:all_patterns}.b). 
Note that, unlike other patterns, checkered patterns don't follow a strict format but are defined by these rapid color shifts.
% While mainly found in the Random model's outputs, some shadow models occasionally produced these patterns.
Such patterns are commonly found in the output of the \mr{} model.
We observed that the appearance of a checkered pattern almost always decreased participants' trust in the model. When faced with a checkered pattern, participants quickly concluded that the model was performing poorly.

\paragraph{{\textbf{Summary:}}} We can summarize the participants' behavior in response to different patterns as follows:

\begin{itemize}
    \item[$\mathbf{R_6}$] With the sole exception of single outliers, we found that participants do not inspect all the heat map cells; instead, they inspect only a fraction of the cells, depending on the pattern in which the cell resides.

    \item[$\mathbf{R_7}$] An entirely green or entirely red row (i.e., a uni-color row) in the heat map fosters a positive perception regarding the model's performance among participants, even before a thorough inspection.
    \item[$\mathbf{R_8}$] If there is an outlier cell in a row, and upon inspection, it turns out to be correct, it heavily tips the participants' judgment in the model's favor.
    \item[$\mathbf{R_{9}}$] The existence of 2D checkered-like patterns in a heat map leads participants to form a negative opinion regarding the model's performance quickly.
\end{itemize}

In conclusion, the visual patterns allowed participants to narrow the decision space from all cells in the heat map to a few critical cells, simplifying the decision-making process.
This provides strong evidence in support of our hypothesis $H_2$, confirming its validity. \hfill\qedsymbol{}

\subsection{Perceived Difficulty in Rating a Model}

% \begin{figure*}[!ht]
%     \centering
%     % \hfill
%     \begin{minipage}[t]{0.48\linewidth}
%         \centering
%         \includegraphics[width=\linewidth]{figures/comp_time_vs_f1_all_models.pdf}  
%         \caption{Individual task completion times of participants for study trials, plotted against the \fonelocal{} scores of the trial heat maps.
%         Each dot represents the median value within that range. 
%         }
%         \label{fig:Comp_time_vs_F1}
%     \end{minipage}
%     \hfill
%     \begin{minipage}[t]{0.48\linewidth}
%         \centering
%         \includegraphics[width=\linewidth]{figures/objects_vs_f1_all_models.pdf}  
%         \caption{Number of objects used by our study participants during study trials, plotted against the \fonelocal{} scores of the trial heat maps.
%         Each dot represents the median value within that range. 
%         }
%         \label{fig:Objects_vs_F1}
%     \end{minipage}
% \end{figure*}

% \begin{figure}
%         \centering
%         \includegraphics[width=0.5\linewidth]{figures/objects_vs_f1_all_models.pdf}  
%         \caption{Number of objects used by our study participants during study trials, plotted against the \fonelocal{} scores of the trial heat maps.
%         Each dot represents the median value within that range. 
%         }
%         \label{fig:Objects_vs_F1}
% \end{figure}

We consider the \textit{perceived difficulty} of rating a model to be higher if:
i) the user takes a longer time to make a decision and/or
ii) the user uses more objects to build confidence in their decision.
As such, \textit{completion times} and \textit{the number of objects used} in a trial are reasonable proxies for a trial's perceived difficulty.

\begin{figure}[!ht]
    \centering
    \includegraphics[width=\linewidth]{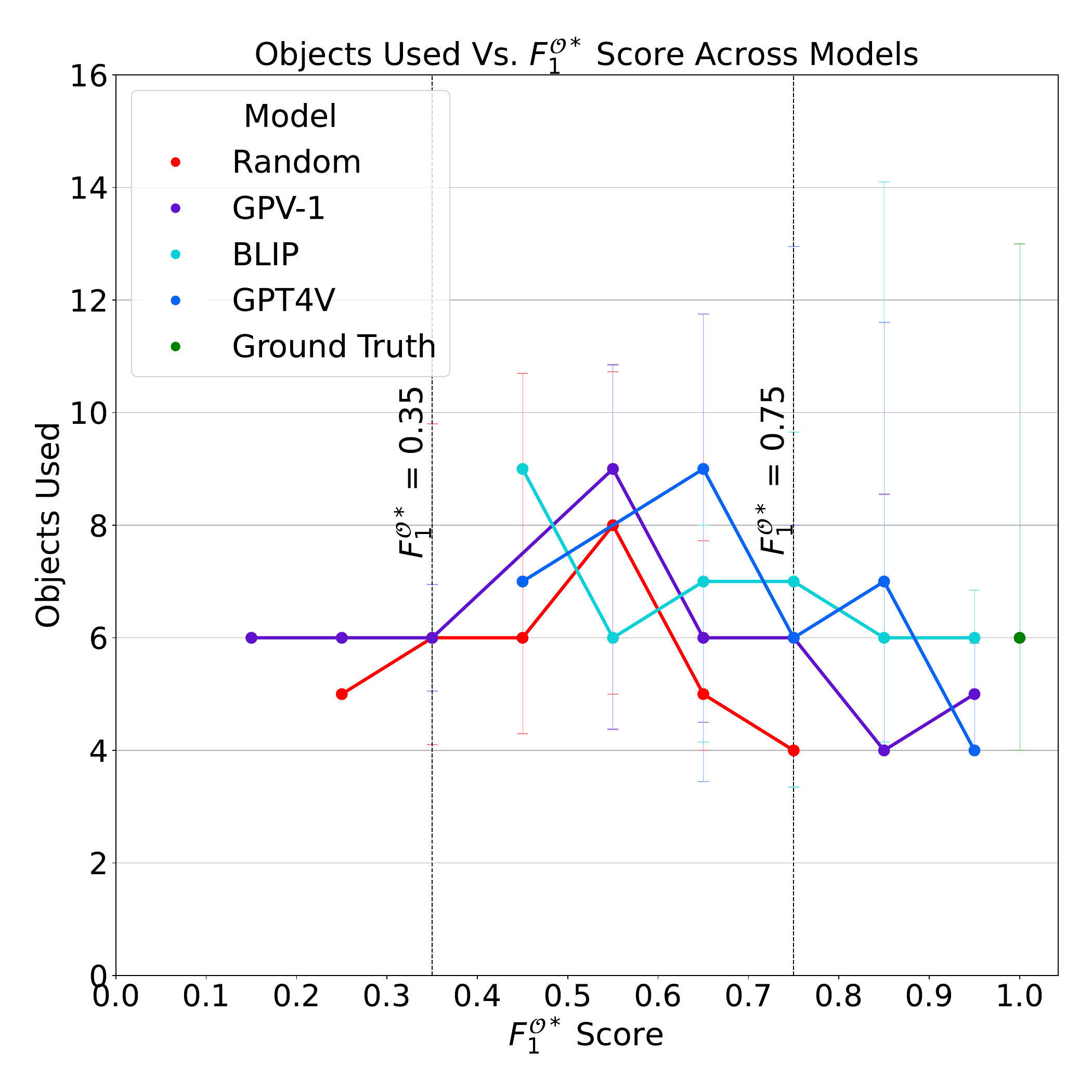}  
    \caption{Number of objects used by our study participants during study trials, plotted against the \fonelocal{} scores of the trial heat maps.
    Each dot represents the median value within that range.}
    \label{fig:Objects_vs_F1}
\end{figure}

\subsubsection{Task Completion Times}
% We found that task completion time is a function of the model's \fonelocal{} and participants' expertise in AI.
% Fig.~\ref{fig:Comp_time_vs_F1} illustrates the relationship between the time participants took to complete a task during the study and the \fonelocal{} scores of the different models used in the task.
We summarize the dominant trends and general observations in task completion times as follows:
\begin{itemize}
    % \item[$\mathbf{R_{11}}$]
    % Participants could decide on a model in under 2 minutes (120 seconds) if the model's underlying performance was poor (\fonelocal{} < $0.35$) or very good (\fonelocal{} > $0.75$). The perceived complexity of such models is low.
    % % 
    % \item[$\mathbf{R_{12}}$]
    % Participants faced difficulty evaluating models whose underlying performance fell between 0.35 and 0.75 ($0.35 \le$ \fonelocal{} $\le 0.75$); their completion time increased to over 2 minutes, with some participants taking more than 5 minutes.

    \item[$\mathbf{R_{10}}$]
    For the \mr{} model, participants were generally able to make a decision within 2 minutes (120 seconds), owing to the model's consistently poor performance. This pattern also held true for models that performed well, such as \mgpt{} and \mgt{}.
    In both cases, the low variation in performance made it easier for participants to assess the model’s reliability, resulting in lower task completion times. Consequently, the perceived complexity of the task was relatively low.
    
    \item[$\mathbf{R_{11}}$]
    In contrast, when a model's performance was more ambiguous---neither clearly good nor definitively poor---participants faced greater difficulty in evaluating the model. 
    This increased uncertainty resulted in longer task completion times, as participants had to spend more time interpreting the results. In these cases, the perceived complexity of the task was higher.

    % \item[$\mathbf{R_{13}}$]
    % Expert participants became more cautious when they perceived a model as good. They increased the number of objects and used `spy' objects to stress-test the model before rating it as highly reliable. This is evident in the long whisker of the rightmost data point, representing the GT model. Some of these experts recognized random models in as little as 50 seconds.
\end{itemize}

\subsubsection{The Number of Objects Used}
We found that the number of objects used in a study task---another indicator of task difficulty---showed patterns similar to those we observed for task completion times.
% displays a similar trend to task completion time when plotted against \fonelocal{}, as shown in Fig.~\ref{fig:Objects_vs_F1}.
Fig.~\ref{fig:Objects_vs_F1} shows the number of objects used in different study trials against \fonelocal{}.

We advised the participants to start the trial by selecting 4-6 objects, which most participants followed. 
However, as the model's perceived difficulty increased or the model showed oracle-like performance, this number increased to over 10.
Key findings from this section are:

\begin{itemize}
    \item[$\mathbf{R_{12}}$]
    \textit{Adding More Objects to Reduce Uncertainty}: This was a predominant trend among all our participants  --  they added more objects to the heat map when unsure about the model's performance with the current set of objects. Some participants used as many as 16 objects (P13). This behavior aligns with the concept of gradual trust-building in technology; when the level of trust is insufficient, users tend to augment the number of objects to enhance their confidence in their evaluation.
    \item[$\mathbf{R_{13}}$]
    \textit{Utilizing Inter-Object Relationships}: Expert participants utilized the inter-object relationship as a criterion for evaluating a model's performance. For example, some participants used \emph{Person with a Disability} and \emph{White Cane} in the heat map as these two objects are interlinked. Participants tended to rate the model highly if the model recognized both objects in a frame.
\end{itemize}

%done

\subsection{Role of Participants' Machine Learning Expertise in using \sysname{}}
\label{sec:expertise-role}

\subsubsection{Quality of Ratings for Users with Varied Expertise}

\begin{figure}[!ht]
    \centering
    \includegraphics[width=\linewidth]{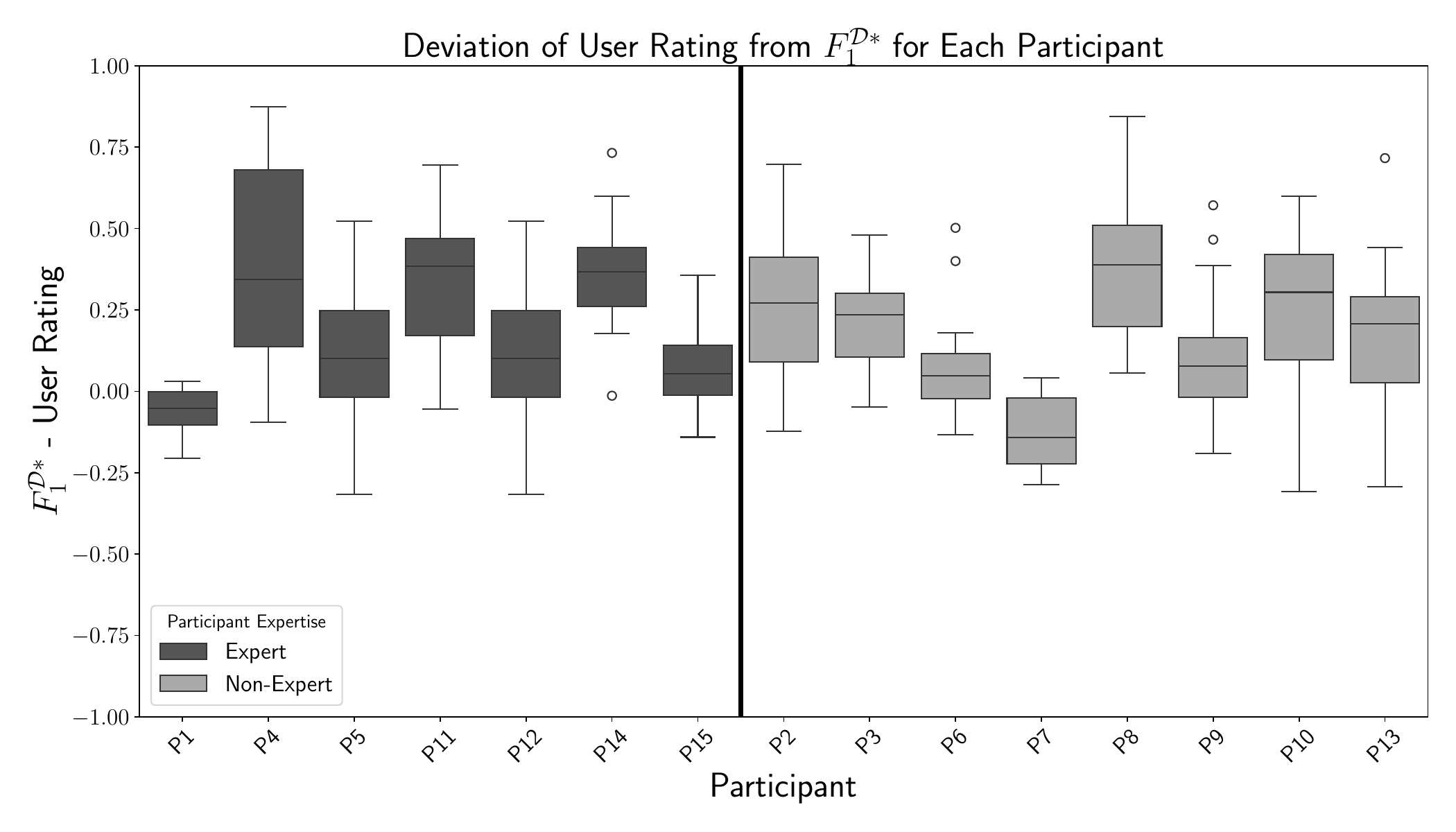}
    \caption{Boxplots showing the deviation of user ratings (\fonelocal{} - Rating) for all participants, who are divided into two groups: Experts (on the left) and Non-Experts (on the right). }
    \label{fig:QR_vs_participant}
\end{figure}

We define rating quality as the degree of agreement between a participant’s reliability rating and the corresponding \fonelocal{} score; the greater the deviation, the lower the rating quality. Figure~\ref{fig:QR_vs_participant} visualizes these deviations across all trials. Participants are grouped into two categories—Experts and Non-Experts—as defined in Table~\ref{table:participants}. To assess potential differences in performance between the groups, we conducted a two-tailed Mann-Whitney U test on trial completion times. The results indicate no statistically significant difference between Experts and Non-Experts ($U$ = 15118.5, $p$ = 0.495), suggesting comparable efficiency in using \sysname{} across experience levels.

\subsubsection{Trial Completion Times for Users with Varied Expertise}
\begin{figure}[!ht]
    \centering
    \includegraphics[width=\linewidth]{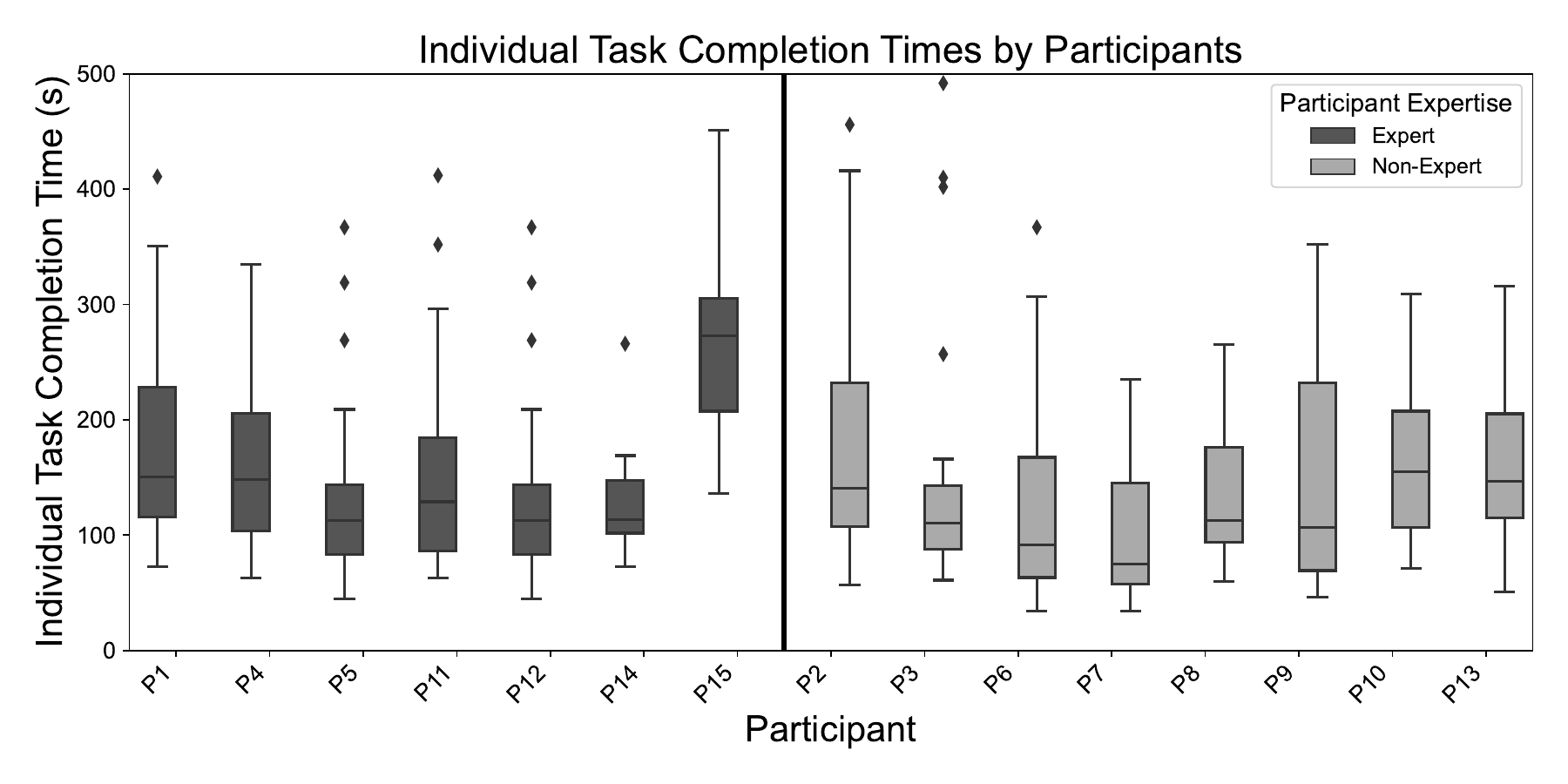}
    \caption{Boxplots showing the trial completion times for all participants, who are divided into two groups: Experts (on the left) and Non-Experts (on the right). No statistically significant difference was found between the two groups.}
    \label{fig:average_timing_per_participant}
\end{figure}

Figure~\ref{fig:average_timing_per_participant} presents the average trial completion time for each participant. 
Participants are categorized into two groups—Experts and Non-Experts—as defined in Table~\ref{table:participants}. 
A two-tailed Mann-Whitney U test was conducted to compare trial completion times between the groups. 
The test revealed no statistically significant difference ($U$ = 16142.5, $p$ = 0.071), suggesting that experience level did not substantially impact task efficiency when using \sysname{}.

\paragraph{Summary.} Based on the above findings, we conclude that prior experience in Machine Learning (ML) and Computer Vision (CV) does not significantly affect a user's ability to use \sysname{} effectively. 
Both expert and non-expert participants performed similarly in terms of rating accuracy and task completion time.

\begin{figure*}[!ht]
    \centering
    \includegraphics[width=0.95\linewidth]{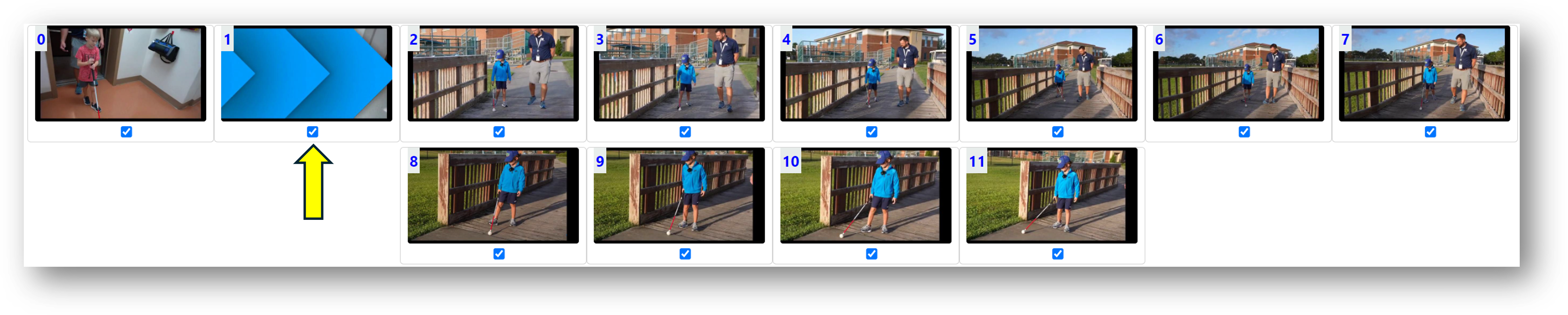}
    \caption{Spy column (i.e., frame), shown using a yellow arrow. }
    \label{fig:limus_column}
\end{figure*}

\subsection{Usability, User Experience, and Observations}
\subsubsection{Users Pay Attention to Inter-frame Similarity.}
\label{subsec:inter-frame}
Participants mentioned that a reliable model is expected to produce identical outputs when two subsequent frames are nearly or entirely similar. 
However, models inadequately trained on particular objects or those making random predictions might not exhibit this consistency. Several participants (P2, P4, P5, P12, P13) identified and heavily penalized this inconsistency issue. 
As articulated by P4:

\begin{quotation}
\emph{``... look at this one (frame), almost identical to its previous one, yet the model is saying so many different things (in the heat map). Either this model has not learned anything or makes random predictions.''}
\end{quotation}

%done
\subsubsection{Users Allow Leniency for Uncertain Objects.}
We identified two distinct scenarios where some participants (P1, P2, P4, P7, P8, P12) exhibited a more forgiving attitude towards the model's mistakes: i) encounters with unknown objects and ii) dealing with confusing objects. 
Common unknown objects include \emph{Turnstile}, \emph{Sloped Curb}, \emph{Sloped Driveway}.
Additionally, participants demonstrated a higher tolerance for mistakes for objects considered confusing, such as \emph{Vegetation}, \emph{Flush Doors}, \emph{Gates}, and \emph{Fences}.

% In this regard, P15 remarked:

% \begin{quotation}
%     \emph{``... I don't mind a mistake or two if I know my use case is video or image description---as I can always build my context from nearby frames. However, if the use case is blind navigation or automated driving, I cannot help penalizing a model for missing a `Car' or a `Crosswalk.'''}
% \end{quotation}

\subsubsection{Use of ``Spy'' Rows and Columns.}
As mentioned in Sec.~\ref{subsec:obj_selection}, "Spy" objects are highly improbable in any video within our dataset. 
In an ideal scenario, these objects should produce entirely red rows in the heat map, indicating the model's reliability in correctly identifying their absence.
Participants leveraged this criterion, adding objects like \emph{Snow} and \emph{Turnstile} to assess model performance. 
They promptly scanned heat map rows for these objects, noting any green cells, which usually resulted in lower model ratings.

One participant (P11) also used a video frame as a ``Spy''. 
This specific video frame was between the transition of two keyframes and contained no information, as shown in Fig.~\ref{fig:limus_column}.
P11 used this frame as a ``Spy'' column instead.
In other words, any green cell in this column in the heat map would reveal the weakness of the corresponding model. 

\subsubsection{Context is Crucial in Decision Making.}
\label{subsec:role_of_contect}
Recall that we briefed the participants on the potential use cases of LMMs, such as automated driving and blind navigation assistance, emphasizing their potential when they perform reliably.
We observed that this information influenced some participants' evaluation process, making them particularly cautious when assigning high ratings (90-100\%) to a model.
Participant P15 highlighted how his ratings were shaped by considering the practical applications of the LMM, stating:

\begin{quotation}
\emph{"... I don't mind a mistake or two if I know my use case is video or image description---as I can always build my context from nearby frames. However, if the use case is blind navigation or automated driving, I cannot help penalizing a model for missing a 'Car' or a 'Crosswalk.'"}
\end{quotation}

\subsubsection{Objects Not Present in Every Frame Yields More Confident Judgement.}
Although objects that are consistently present (resulting in an all-green row) or consistently absent (resulting in an all-red row) across all frames make scanning more manageable for users, some participants (P2, P6, P11) suggested that using objects that appear intermittently could provide a more robust test for the model.
When the model generates an intermittent yet correct pattern, P11 said he feels more confident about its robustness.

\begin{figure}[!ht]
    \centering
    \includegraphics[width=\linewidth]{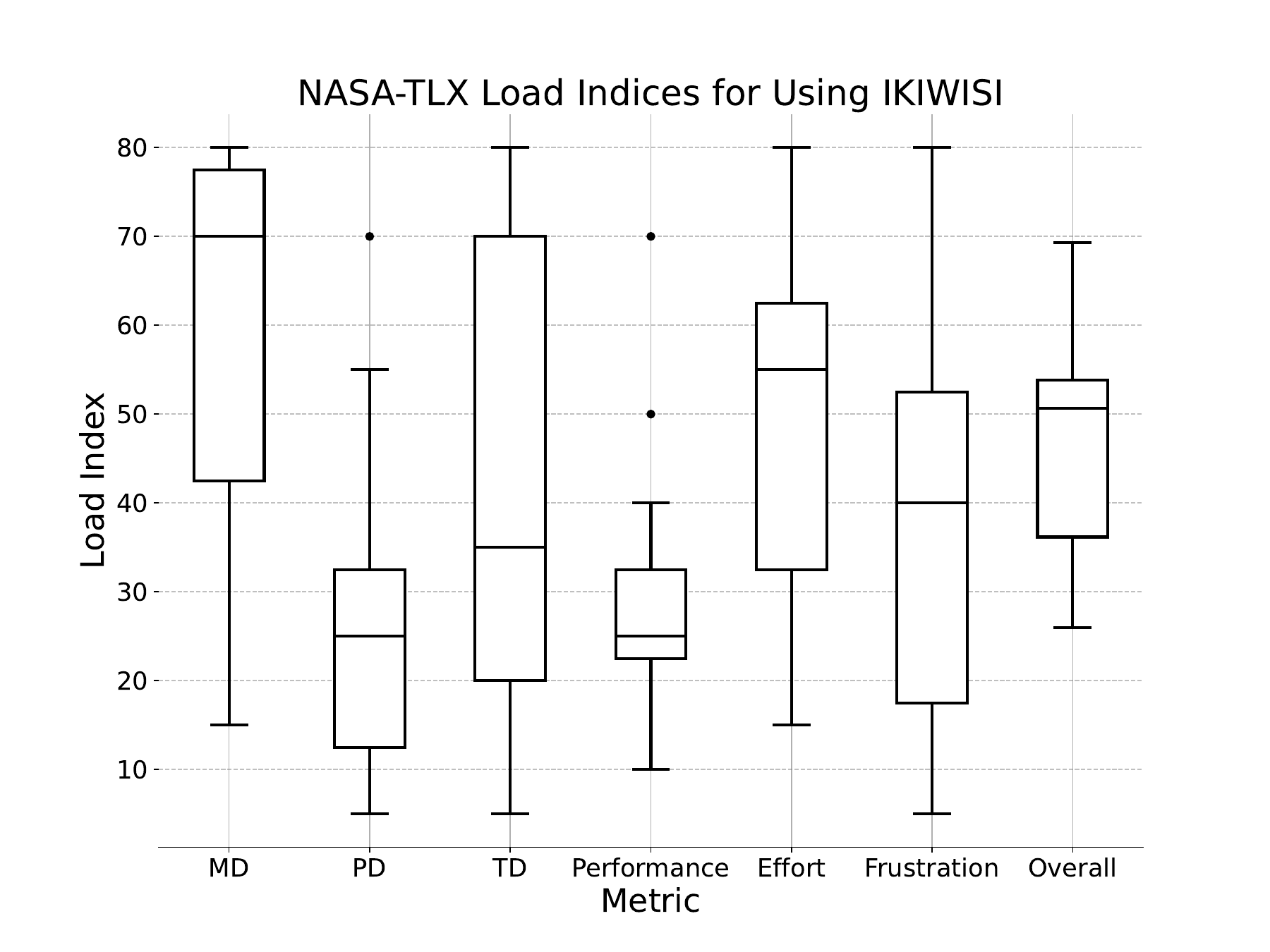}
    \caption{NASA-TLX load indices for using \sysname{}. MD = Mental Demand, PD = Physical Demand, TD = Temporal Demand. Lower scores indicate better performance.}
    \label{fig:nasa_tlx}
\end{figure}

\subsubsection{Usage of the Click-to-Zoom Feature Reduces Over Time.}
All participants utilized the click-to-zoom feature during the initial 4-6 trials. 
However, as the study progressed, users demonstrated an increased ability to navigate the heat map quickly, analyze relevant information, and conclude about the model's performance without relying on the click-to-zoom feature. 
Participants quickly internalized the objects they needed to identify, allowing them to efficiently scan subsequent frames without needing to use the zoom feature.

\subsubsection{NASA-TLX Evaluation}

The NASA-TLX score for \sysname{} was 47.55 (SD: 13.17), as shown in the rightmost box in Fig.\ref{fig:nasa_tlx}. This is lower than the threshold of 68~\cite{grier2015high}, suggesting that participants' mental workload was low during the study. 
Out of the six components in NASA-TLX, the scores for mental demand were $60.67$ (SD: $20.78$), and for effort were $49.67$ (SD: $21.75$). These two contributed the most to the overall score.
The physical demand (mean: $26$, SD: $18.73$) was relatively low, which can be attributed to the interactive nature of \sysname{}, where most physical activity is limited to moving the mouse or trackpad.
Overall, the feedback was overwhelmingly positive regarding the user experience and usability of \sysname{}.

\section{Discussion and Future Work}
\label{sec:discussion}
We now discuss the implications of our findings, potential extensions, and limitations of \sysname{}.

\subsection{Balancing Familiar Visualization with Novel Application}
While heatmaps represent a well-established visualization method, \sysname{} innovates through its application—evaluating vision-language models without ground truth. Unlike ConfusionFlow~\cite{hinterreiter2020confusionflow} or ARGUS~\cite{castelo2023argus}, which emphasize technical metrics and confidence scores, \sysname{} prioritizes human perception as the baseline for model assessment. Its interactive design enables users to toggle cells and correct errors, making alignment gaps immediately visible.

This design prioritizes usability over novelty—a conscious trade-off validated by our pilot study participants (Sec.~\ref{subsec:first_pilot_study}). When comparing various visualization approaches, they consistently preferred the heatmap for its interpretability and efficiency despite alternative interfaces appearing more innovative. This feedback reinforced our approach: even sophisticated visualization techniques lose value when users struggle to interpret them. By adapting a familiar format to a new purpose, \sysname{} creates an accessible evaluation framework for both AI specialists and non-experts supporting visually impaired users.

\subsection{As a Cognitive Audit Tool for AI Systems}
Beyond its practical utility as an evaluation interface, \sysname{} functions as a cognitive audit mechanism that exposes misalignments between human commonsense reasoning and machine perception. Our study reveals that users leverage distinctive visual patterns to detect these misalignments without examining the entire heatmap in detail.

Users intuitively found meaning in emergent patterns—participants quickly penalized checkered patterns as indicating random predictions (Sec.~\ref{subsec:role_of_patterns}, \textbf{R9}) while viewing consistent rows as evidence of reliability (\textbf{R7}). Single outlier cells drew particular scrutiny, with users verifying these specific instances to build confidence in or doubt about the model (\textbf{R8}). Users also identified temporal inconsistencies across visually similar frames as violations of physical plausibility (Sec.~\ref{subsec:inter-frame}), applying their implicit understanding of world continuity to judge model performance.

The ``spy object'' mechanism further enhanced this cognitive auditing capability. By deliberately including objects users knew were absent, \sysname{} enabled hypothesis-driven exploration of model limitations—a hallmark of effective cognitive auditing. This approach transforms model evaluation from passive inspection to active inquiry, where users probe edge cases and construct experiments that reveal the boundaries of model understanding.

\subsection{Advancing Human-in-the-Loop Evaluation}
\sysname{} addresses three critical limitations in current AI evaluation approaches. First, it overcomes the absence of ground truth in many real-world applications by positioning human perception as the reference standard. Second, it bridges the expertise gap by creating an evaluation interface accessible to both experts and non-experts. Third, it enables context-specific assessment tailored to users' unique needs rather than generic benchmarks.

The framework extends beyond technical metrics like $F_1$-scores, which depend on object taxonomies and labeled datasets~\cite{gptv_system_card}. Instead, \sysname{} enables lightweight, human-centered assessments through selective inspection of objects and frames. Our user study demonstrated strong alignment between participant ratings and objective performance metrics, confirming that users can reach accurate conclusions without comprehensive examination of all data points.

This approach adapts to diverse tasks beyond object recognition. For image captioning, rows could represent key phrases with cells indicating whether these elements appear consistently across frames. For multimodal reasoning, rows could track inference steps across temporal sequences. For anomaly detection, rows could represent expected states with cells showing conformance or violation. In each case, \sysname{} transforms assessment from abstract metrics to visible patterns that leverage human perceptual strengths.

Most importantly, \sysname{} democratizes model evaluation by creating a low-barrier interface that allows users to contribute expertise based on real-world context and expectations. Unlike many specialized evaluation frameworks that require technical knowledge, \sysname{} enables domain experts (such as accessibility specialists) to evaluate models based on their practical needs. This inclusivity supports iterative improvement: users identify specific limitations relevant to their use cases, helping developers target enhancements that improve reliability in deployed settings.

\subsection{Future Work}
Several promising directions could extend \sysname{}'s capabilities and impact:
\subsubsection{Enhancing Visual and Interactive Design}
Current design choices create opportunities for further refinement. The red-green color scheme, while effective for communicating presence/absence distinctions, carries cultural associations with correctness/error that may subtly influence user judgment. Future versions could incorporate neutral color palettes, customizable schemes, or alternative visual encodings to minimize unintended interpretation biases.

We also plan to explore alternative input modalities that enhance navigation efficiency. Rotational input devices like Surface Dial~\cite{surface_dial}, Speed-Dial~\cite{Billah_speeddial}, and Wheeler~\cite{islam2024wheeler, islam2024wheelerdemo} could improve interaction with dense heatmaps compared to traditional clicking. Wheeler's three-wheel configuration presents particular promise, allowing users to navigate horizontally across frames, vertically across objects, and control focus granularity with separate wheels.

\subsubsection{Developing a Standardized Evaluation API}
Current vision-language model evaluation systems use proprietary interfaces that hinder consistent assessment across platforms. Drawing inspiration from accessibility APIs like UI Automation in Windows, we envision a standardized, low-bandwidth evaluation framework for LMMs similar to our prior work on remote access systems for visually impaired users~\cite{Billah_sinter}. Such an API would establish a uniform protocol for input parameters (video source, frame range, object list, model identifier) and standardized output formats (JSON-structured dictionaries directly compatible with visualization tools like \sysname{}).

This standardization would yield three key benefits. First, it would enable consistent cross-model comparisons without requiring developers to implement custom integrations for each proprietary system. Second, it would significantly reduce bandwidth requirements compared to current approaches that transmit full images and videos to cloud-based models. Finally, it would create a foundation for automated testing pipelines that could evaluate models across diverse scenarios without human intervention, while preserving human-in-the-loop capabilities when needed.

Commercial LMM vendors could implement this API using their existing service architectures, much as operating system developers have incorporated standardized accessibility APIs into their platforms. By separating the evaluation interface from proprietary implementation details, this approach would advance model transparency while preserving vendors' intellectual property—creating a more accessible ecosystem for both model developers and end users.

\subsubsection{Expanding Evaluation Scope}
Our current evaluation primarily involved participants with academic backgrounds. To strengthen ecological validity, future studies should include diverse participant populations including accessibility researchers, robotics engineers, and laypeople without technical expertise. This broader evaluation would reveal how different user groups interpret visual patterns and whether the system maintains its effectiveness across varying expertise levels.

\sysname{} also needs testing in time-sensitive, high-stakes contexts such as assistive navigation or robotic vision applications. These scenarios would test not only usability but also effectiveness in supporting critical judgment under pressure—an essential requirement for real-world deployment.

\subsection{Limitations}
This study has several constraints. First, we used pre-generated model outputs rather than real-time predictions due to performance limitations of current models. Second, we focused exclusively on multi-object recognition in video data, though the approach could extend to other tasks. Third, \sysname{} cannot currently track multiple instances of the same object type within a scene (e.g., distinguishing between several cars)—resolving this ambiguity would require analyzing spatial coordinates in model outputs beyond the current design scope. Finally, our evaluation used a limited set of curated videos that may not fully represent the diversity and complexity of real-world environments.

\section{Conclusion}
\sysname{} transforms how humans evaluate AI vision systems by generating distinctive visual patterns that reveal model reliability in real-world contexts where ground truth rarely exists. At its core, a binary heat map creates an interactive interface where users identify patterns that expose both a model's capabilities and its limitations in multi-object recognition tasks.

Our research-through-design process, grounded in human visual perception principles, yielded an evaluation tool that balances technical rigor with intuitive design. Through iterative refinement and user feedback, we created an interface where even non-experts can make sophisticated judgments about complex AI systems. Our study with 15 participants confirmed \sysname{}'s effectiveness: users made reliability assessments that correlated strongly with objective performance metrics, yet needed to examine only a small fraction of heat map cells to reach these conclusions.

Beyond its practical utility, \sysname{} represents a shift in AI evaluation philosophy—from reliance on automated metrics toward human-centered assessment frameworks that democratize the evaluation process. By enabling people with diverse expertise levels to assess AI systems through visual patterns rather than technical specifications, \sysname{} bridges the gap between AI development and real-world deployment. This approach not only complements existing evaluation techniques but also creates opportunities for more inclusive, context-sensitive, and human-aligned AI systems that better serve the needs of their intended users.

\begin{acks}
We thank the anonymous reviewers whose insightful feedback substantially improved this paper. The National Federation of the Blind, Happy Valley Chapter, provided invaluable guidance that shaped our dataset creation. 
We also express our gratitude to the blind content creators whose videos form the foundation of our dataset. This research received partial support from the National Science Foundation (grant \#2326406) through a subaward.

\end{acks}

\balance
\bibliographystyle{ACM-Reference-Format}
\bibliography{Bibliography, Bibliography2, Bibliography3}


\begin{thebibliography}{85}

%%% the Web address.

\ifx \showCODEN    \undefined \def \showCODEN     #1{\unskip}     \fi
\ifx \showDOI      \undefined \def \showDOI       #1{#1}\fi
\ifx \showISBNx    \undefined \def \showISBNx     #1{\unskip}     \fi
\ifx \showISBNxiii \undefined \def \showISBNxiii  #1{\unskip}     \fi
\ifx \showISSN     \undefined \def \showISSN      #1{\unskip}     \fi
\ifx \showLCCN     \undefined \def \showLCCN      #1{\unskip}     \fi
\ifx \shownote     \undefined \def \shownote      #1{#1}          \fi
\ifx \showarticletitle \undefined \def \showarticletitle #1{#1}   \fi
\ifx \showURL      \undefined \def \showURL       {\relax}        \fi
% invisible to TeX
\providecommand\bibfield[2]{#2}
\providecommand\bibinfo[2]{#2}
\providecommand\natexlab[1]{#1}
\providecommand\showeprint[2][]{arXiv:#2}

\bibitem[\protect\citeauthoryear{??}{Air}{[n.d.]}]%
        {Aira2020}
 \bibinfo{year}{[n.d.]}\natexlab{}.
\newblock \bibinfo{title}{Aira}.
\newblock \bibinfo{howpublished}{\url{https://aira.io/}}.
\newblock


\bibitem[\protect\citeauthoryear{Alsallakh, Hanbury, Hauser, Miksch, and Rauber}{Alsallakh et~al\mbox{.}}{2014}]%
        {alsallakh2014visual}
\bibfield{author}{\bibinfo{person}{Bilal Alsallakh}, \bibinfo{person}{Allan Hanbury}, \bibinfo{person}{Helwig Hauser}, \bibinfo{person}{Silvia Miksch}, {and} \bibinfo{person}{Andreas Rauber}.} \bibinfo{year}{2014}\natexlab{}.
\newblock \showarticletitle{Visual methods for analyzing probabilistic classification data}.
\newblock \bibinfo{journal}{\emph{IEEE transactions on visualization and computer graphics}} \bibinfo{volume}{20}, \bibinfo{number}{12} (\bibinfo{year}{2014}), \bibinfo{pages}{1703--1712}.
\newblock


\bibitem[\protect\citeauthoryear{Antol, Agrawal, Lu, Mitchell, Batra, Zitnick, and Parikh}{Antol et~al\mbox{.}}{2015}]%
        {antol2015vqa}
\bibfield{author}{\bibinfo{person}{Stanislaw Antol}, \bibinfo{person}{Aishwarya Agrawal}, \bibinfo{person}{Jiasen Lu}, \bibinfo{person}{Margaret Mitchell}, \bibinfo{person}{Dhruv Batra}, \bibinfo{person}{C~Lawrence Zitnick}, {and} \bibinfo{person}{Devi Parikh}.} \bibinfo{year}{2015}\natexlab{}.
\newblock \showarticletitle{{VQA}: {Visual} {Question} {Answering}}. In \bibinfo{booktitle}{\emph{Proceedings of the IEEE International Conference on computer vision}}.
\newblock


\bibitem[\protect\citeauthoryear{Behley, Garbade, Milioto, Quenzel, Behnke, Stachniss, and Gall}{Behley et~al\mbox{.}}{2019}]%
        {behley2019iccv}
\bibfield{author}{\bibinfo{person}{J. Behley}, \bibinfo{person}{M. Garbade}, \bibinfo{person}{A. Milioto}, \bibinfo{person}{J. Quenzel}, \bibinfo{person}{S. Behnke}, \bibinfo{person}{C. Stachniss}, {and} \bibinfo{person}{J. Gall}.} \bibinfo{year}{2019}\natexlab{}.
\newblock \showarticletitle{{SemanticKITTI: A Dataset for Semantic Scene Understanding of LiDAR Sequences}}. In \bibinfo{booktitle}{\emph{Proc. of the IEEE/CVF International Conf.~on Computer Vision (ICCV)}}.
\newblock


\bibitem[\protect\citeauthoryear{BeMyEyes}{BeMyEyes}{2021}]%
        {BeMyEyes2020}
\bibfield{author}{\bibinfo{person}{BeMyEyes}.} \bibinfo{year}{2021}\natexlab{}.
\newblock \bibinfo{title}{Be My Eyes}.
\newblock \bibinfo{howpublished}{\url{https://www.bemyeyes.com/}}.
\newblock


\bibitem[\protect\citeauthoryear{Billah, Ashok, Porter, and Ramakrishnan}{Billah et~al\mbox{.}}{2017}]%
        {Billah_speeddial}
\bibfield{author}{\bibinfo{person}{Syed~Masum Billah}, \bibinfo{person}{Vikas Ashok}, \bibinfo{person}{Donald~E. Porter}, {and} \bibinfo{person}{I.V. Ramakrishnan}.} \bibinfo{year}{2017}\natexlab{}.
\newblock \showarticletitle{Speed-Dial: A Surrogate Mouse for Non-Visual Web Browsing}. In \bibinfo{booktitle}{\emph{Proceedings of the 19th International ACM SIGACCESS Conference on Computers and Accessibility}}. \bibinfo{publisher}{ACM}, \bibinfo{address}{3132531}, \bibinfo{pages}{110--119}.
\newblock
\showISBNx{ISBN}
\urldef\tempurl%
\url{https://doi.org/10.1145/3132525.3132531}
\showDOI{\tempurl}


\bibitem[\protect\citeauthoryear{Billah and Gauch}{Billah and Gauch}{2015}]%
        {billah2015social}
\bibfield{author}{\bibinfo{person}{Syed~Masum Billah} {and} \bibinfo{person}{Susan Gauch}.} \bibinfo{year}{2015}\natexlab{}.
\newblock \showarticletitle{Social network analysis for predicting emerging researchers}. In \bibinfo{booktitle}{\emph{2015 7th International Joint Conference on Knowledge Discovery, Knowledge Engineering and Knowledge Management (IC3K)}}, Vol.~\bibinfo{volume}{1}. IEEE, \bibinfo{pages}{27--35}.
\newblock


\bibitem[\protect\citeauthoryear{Billah, Porter, and Ramakrishnan}{Billah et~al\mbox{.}}{2016}]%
        {Billah_sinter}
\bibfield{author}{\bibinfo{person}{Syed~Masum Billah}, \bibinfo{person}{Donald~E. Porter}, {and} \bibinfo{person}{I.~V. Ramakrishnan}.} \bibinfo{year}{2016}\natexlab{}.
\newblock \showarticletitle{Sinter: low-bandwidth remote access for the visually-impaired}. In \bibinfo{booktitle}{\emph{Proceedings of the Eleventh European Conference on Computer Systems}}. \bibinfo{publisher}{ACM}, \bibinfo{address}{2901335}, \bibinfo{pages}{1--16}.
\newblock
\showISBNx{ISBN}
\urldef\tempurl%
\url{https://doi.org/10.1145/2901318.2901335}
\showDOI{\tempurl}


\bibitem[\protect\citeauthoryear{Blanch, Li, Escalera, and Nasrollahi}{Blanch et~al\mbox{.}}{2024}]%
        {blanch2024lidar}
\bibfield{author}{\bibinfo{person}{Miquel~Romero Blanch}, \bibinfo{person}{Zenjie Li}, \bibinfo{person}{Sergio Escalera}, {and} \bibinfo{person}{Kamal Nasrollahi}.} \bibinfo{year}{2024}\natexlab{}.
\newblock \showarticletitle{LiDAR-Assisted 3D Human Detection for Video Surveillance}. In \bibinfo{booktitle}{\emph{Proceedings of the IEEE/CVF Winter Conference on Applications of Computer Vision}}. \bibinfo{pages}{123--131}.
\newblock


\bibitem[\protect\citeauthoryear{Brachman and Levesque}{Brachman and Levesque}{2023}]%
        {brachman2023machines}
\bibfield{author}{\bibinfo{person}{Ronald~J Brachman} {and} \bibinfo{person}{Hector~J Levesque}.} \bibinfo{year}{2023}\natexlab{}.
\newblock \bibinfo{booktitle}{\emph{Machines like us: toward AI with common sense}}.
\newblock \bibinfo{publisher}{MIT Press}.
\newblock


\bibitem[\protect\citeauthoryear{Bubeck, Chandrasekaran, Eldan, Gehrke, Horvitz, Kamar, Lee, Lee, Li, Lundberg, et~al\mbox{.}}{Bubeck et~al\mbox{.}}{2023}]%
        {bubeck2023sparks}
\bibfield{author}{\bibinfo{person}{S{\'e}bastien Bubeck}, \bibinfo{person}{Varun Chandrasekaran}, \bibinfo{person}{Ronen Eldan}, \bibinfo{person}{Johannes Gehrke}, \bibinfo{person}{Eric Horvitz}, \bibinfo{person}{Ece Kamar}, \bibinfo{person}{Peter Lee}, \bibinfo{person}{Yin~Tat Lee}, \bibinfo{person}{Yuanzhi Li}, \bibinfo{person}{Scott Lundberg}, {et~al\mbox{.}}} \bibinfo{year}{2023}\natexlab{}.
\newblock \showarticletitle{Sparks of artificial general intelligence: Early experiments with gpt-4}.
\newblock \bibinfo{journal}{\emph{arXiv preprint arXiv:2303.12712}} (\bibinfo{year}{2023}).
\newblock


\bibitem[\protect\citeauthoryear{Carroll and Olson}{Carroll and Olson}{1988}]%
        {carroll1988mental}
\bibfield{author}{\bibinfo{person}{John~M Carroll} {and} \bibinfo{person}{Judith~Reitman Olson}.} \bibinfo{year}{1988}\natexlab{}.
\newblock \showarticletitle{Mental models in human-computer interaction}.
\newblock \bibinfo{journal}{\emph{Handbook of human-computer interaction}} (\bibinfo{year}{1988}), \bibinfo{pages}{45--65}.
\newblock


\bibitem[\protect\citeauthoryear{Castelo, Rulff, McGowan, Steers, Wu, Chen, Roman, Lopez, Brewer, Zhao, et~al\mbox{.}}{Castelo et~al\mbox{.}}{2023}]%
        {castelo2023argus}
\bibfield{author}{\bibinfo{person}{Sonia Castelo}, \bibinfo{person}{Joao Rulff}, \bibinfo{person}{Erin McGowan}, \bibinfo{person}{Bea Steers}, \bibinfo{person}{Guande Wu}, \bibinfo{person}{Shaoyu Chen}, \bibinfo{person}{Iran Roman}, \bibinfo{person}{Roque Lopez}, \bibinfo{person}{Ethan Brewer}, \bibinfo{person}{Chen Zhao}, {et~al\mbox{.}}} \bibinfo{year}{2023}\natexlab{}.
\newblock \showarticletitle{Argus: Visualization of ai-assisted task guidance in ar}.
\newblock \bibinfo{journal}{\emph{IEEE Transactions on Visualization and Computer Graphics}} (\bibinfo{year}{2023}).
\newblock


\bibitem[\protect\citeauthoryear{Chan, Petryk, Gonzalez, Darrell, and Canny}{Chan et~al\mbox{.}}{2023}]%
        {chan2023clair}
\bibfield{author}{\bibinfo{person}{David Chan}, \bibinfo{person}{Suzanne Petryk}, \bibinfo{person}{Joseph~E Gonzalez}, \bibinfo{person}{Trevor Darrell}, {and} \bibinfo{person}{John Canny}.} \bibinfo{year}{2023}\natexlab{}.
\newblock \showarticletitle{Clair: Evaluating image captions with large language models}.
\newblock \bibinfo{journal}{\emph{arXiv preprint arXiv:2310.12971}} (\bibinfo{year}{2023}).
\newblock


\bibitem[\protect\citeauthoryear{Chen, Sinavski, H{\"u}nermann, Karnsund, Willmott, Birch, Maund, and Shotton}{Chen et~al\mbox{.}}{2023}]%
        {chen2023driving}
\bibfield{author}{\bibinfo{person}{Long Chen}, \bibinfo{person}{Oleg Sinavski}, \bibinfo{person}{Jan H{\"u}nermann}, \bibinfo{person}{Alice Karnsund}, \bibinfo{person}{Andrew~James Willmott}, \bibinfo{person}{Danny Birch}, \bibinfo{person}{Daniel Maund}, {and} \bibinfo{person}{Jamie Shotton}.} \bibinfo{year}{2023}\natexlab{}.
\newblock \showarticletitle{Driving with llms: Fusing object-level vector modality for explainable autonomous driving}.
\newblock \bibinfo{journal}{\emph{arXiv preprint arXiv:2310.01957}} (\bibinfo{year}{2023}).
\newblock


\bibitem[\protect\citeauthoryear{Datta and Dickerson}{Datta and Dickerson}{2023}]%
        {datta2023s}
\bibfield{author}{\bibinfo{person}{Teresa Datta} {and} \bibinfo{person}{John~P Dickerson}.} \bibinfo{year}{2023}\natexlab{}.
\newblock \showarticletitle{Who's Thinking? A Push for Human-Centered Evaluation of LLMs using the XAI Playbook}.
\newblock \bibinfo{journal}{\emph{arXiv preprint arXiv:2303.06223}} (\bibinfo{year}{2023}).
\newblock


\bibitem[\protect\citeauthoryear{Feng, Haase-Sch{\"u}tz, Rosenbaum, Hertlein, Glaeser, Timm, Wiesbeck, and Dietmayer}{Feng et~al\mbox{.}}{2020}]%
        {feng2020deep}
\bibfield{author}{\bibinfo{person}{Di Feng}, \bibinfo{person}{Christian Haase-Sch{\"u}tz}, \bibinfo{person}{Lars Rosenbaum}, \bibinfo{person}{Heinz Hertlein}, \bibinfo{person}{Claudius Glaeser}, \bibinfo{person}{Fabian Timm}, \bibinfo{person}{Werner Wiesbeck}, {and} \bibinfo{person}{Klaus Dietmayer}.} \bibinfo{year}{2020}\natexlab{}.
\newblock \showarticletitle{Deep multi-modal object detection and semantic segmentation for autonomous driving: Datasets, methods, and challenges}.
\newblock \bibinfo{journal}{\emph{IEEE Transactions on Intelligent Transportation Systems}} \bibinfo{volume}{22}, \bibinfo{number}{3} (\bibinfo{year}{2020}), \bibinfo{pages}{1341--1360}.
\newblock


\bibitem[\protect\citeauthoryear{Fernandez-Manjon and Fernandez-Valmayor}{Fernandez-Manjon and Fernandez-Valmayor}{1998}]%
        {fernandez1998building}
\bibfield{author}{\bibinfo{person}{Baltasar Fernandez-Manjon} {and} \bibinfo{person}{Alfredo Fernandez-Valmayor}.} \bibinfo{year}{1998}\natexlab{}.
\newblock \showarticletitle{Building educational tools based on formal concept analysis}.
\newblock \bibinfo{journal}{\emph{Education and Information Technologies}} \bibinfo{volume}{3}, \bibinfo{number}{3} (\bibinfo{year}{1998}), \bibinfo{pages}{187--201}.
\newblock


\bibitem[\protect\citeauthoryear{Forbes, Holtzman, and Choi}{Forbes et~al\mbox{.}}{2019}]%
        {forbes2019neural}
\bibfield{author}{\bibinfo{person}{Maxwell Forbes}, \bibinfo{person}{Ari Holtzman}, {and} \bibinfo{person}{Yejin Choi}.} \bibinfo{year}{2019}\natexlab{}.
\newblock \showarticletitle{Do neural language representations learn physical commonsense?}
\newblock \bibinfo{journal}{\emph{arXiv preprint arXiv:1908.02899}} (\bibinfo{year}{2019}).
\newblock


\bibitem[\protect\citeauthoryear{Goodfellow, Pouget-Abadie, Mirza, Xu, Warde-Farley, Ozair, Courville, and Bengio}{Goodfellow et~al\mbox{.}}{2014}]%
        {goodfellow2014generative}
\bibfield{author}{\bibinfo{person}{Ian~J. Goodfellow}, \bibinfo{person}{Jean Pouget-Abadie}, \bibinfo{person}{Mehdi Mirza}, \bibinfo{person}{Bing Xu}, \bibinfo{person}{David Warde-Farley}, \bibinfo{person}{Sherjil Ozair}, \bibinfo{person}{Aaron Courville}, {and} \bibinfo{person}{Yoshua Bengio}.} \bibinfo{year}{2014}\natexlab{}.
\newblock \showarticletitle{Generative Adversarial Nets}. In \bibinfo{booktitle}{\emph{Advances in Neural Information Processing Systems}}, Vol.~\bibinfo{volume}{27}. \bibinfo{publisher}{Curran Associates, Inc.}, \bibinfo{pages}{2672--2680}.
\newblock


\bibitem[\protect\citeauthoryear{Grier}{Grier}{2015}]%
        {grier2015high}
\bibfield{author}{\bibinfo{person}{Rebecca~A Grier}.} \bibinfo{year}{2015}\natexlab{}.
\newblock \showarticletitle{How high is high? A meta-analysis of NASA-TLX global workload scores}. In \bibinfo{booktitle}{\emph{Proceedings of the human factors and ergonomics society annual meeting}}, Vol.~\bibinfo{volume}{59}. Sage Publications Sage CA: Los Angeles, CA, \bibinfo{pages}{1727--1731}.
\newblock


\bibitem[\protect\citeauthoryear{Gupta, Kamath, Kembhavi, and Hoiem}{Gupta et~al\mbox{.}}{2021}]%
        {Gupta2021GPV}
\bibfield{author}{\bibinfo{person}{Tanmay Gupta}, \bibinfo{person}{A. Kamath}, \bibinfo{person}{Aniruddha Kembhavi}, {and} \bibinfo{person}{Derek Hoiem}.} \bibinfo{year}{2021}\natexlab{}.
\newblock \showarticletitle{Towards General Purpose Vision Systems}.
\newblock \bibinfo{journal}{\emph{ArXiv}}  \bibinfo{volume}{abs/2104.00743} (\bibinfo{year}{2021}).
\newblock


\bibitem[\protect\citeauthoryear{Gupta, Kamath, Kembhavi, and Hoiem}{Gupta et~al\mbox{.}}{2022a}]%
        {gupta2022gpv}
\bibfield{author}{\bibinfo{person}{Tanmay Gupta}, \bibinfo{person}{A. Kamath}, \bibinfo{person}{Aniruddha Kembhavi}, {and} \bibinfo{person}{Derek Hoiem}.} \bibinfo{year}{2022}\natexlab{a}.
\newblock \showarticletitle{Towards General Purpose Vision Systems}.
\newblock \bibinfo{journal}{\emph{Conference of Computer Vision and Pattern Recognition (CVPR)}} (\bibinfo{year}{2022}).
\newblock


\bibitem[\protect\citeauthoryear{Gupta, Kamath, Kembhavi, and Hoiem}{Gupta et~al\mbox{.}}{2022b}]%
        {gupta2022towards}
\bibfield{author}{\bibinfo{person}{Tanmay Gupta}, \bibinfo{person}{Amita Kamath}, \bibinfo{person}{Aniruddha Kembhavi}, {and} \bibinfo{person}{Derek Hoiem}.} \bibinfo{year}{2022}\natexlab{b}.
\newblock \showarticletitle{Towards general purpose vision systems: An end-to-end task-agnostic vision-language architecture}. In \bibinfo{booktitle}{\emph{Proceedings of the IEEE/CVF Conference on Computer Vision and Pattern Recognition}}. \bibinfo{pages}{16399--16409}.
\newblock


\bibitem[\protect\citeauthoryear{Han, Mitra, and Billah}{Han et~al\mbox{.}}{2024}]%
        {han2024uncovering}
\bibfield{author}{\bibinfo{person}{Chaeeun Han}, \bibinfo{person}{Prasenjit Mitra}, {and} \bibinfo{person}{Syed~Masum Billah}.} \bibinfo{year}{2024}\natexlab{}.
\newblock \showarticletitle{Uncovering Human Traits in Determining Real and Spoofed Audio: Insights from Blind and Sighted Individuals}. In \bibinfo{booktitle}{\emph{Proceedings of the CHI Conference on Human Factors in Computing Systems}}. \bibinfo{pages}{1--14}.
\newblock


\bibitem[\protect\citeauthoryear{Hand and Christen}{Hand and Christen}{2018}]%
        {hand2018note}
\bibfield{author}{\bibinfo{person}{David Hand} {and} \bibinfo{person}{Peter Christen}.} \bibinfo{year}{2018}\natexlab{}.
\newblock \showarticletitle{A note on using the F-measure for evaluating record linkage algorithms}.
\newblock \bibinfo{journal}{\emph{Statistics and Computing}}  \bibinfo{volume}{28} (\bibinfo{year}{2018}), \bibinfo{pages}{539--547}.
\newblock


\bibitem[\protect\citeauthoryear{Hinterreiter, Ruch, Stitz, Ennemoser, Bernard, Strobelt, and Streit}{Hinterreiter et~al\mbox{.}}{2020}]%
        {hinterreiter2020confusionflow}
\bibfield{author}{\bibinfo{person}{Andreas Hinterreiter}, \bibinfo{person}{Peter Ruch}, \bibinfo{person}{Holger Stitz}, \bibinfo{person}{Martin Ennemoser}, \bibinfo{person}{J{\"u}rgen Bernard}, \bibinfo{person}{Hendrik Strobelt}, {and} \bibinfo{person}{Marc Streit}.} \bibinfo{year}{2020}\natexlab{}.
\newblock \showarticletitle{ConfusionFlow: A model-agnostic visualization for temporal analysis of classifier confusion}.
\newblock \bibinfo{journal}{\emph{IEEE Transactions on Visualization and Computer Graphics}} \bibinfo{volume}{28}, \bibinfo{number}{2} (\bibinfo{year}{2020}), \bibinfo{pages}{1222--1236}.
\newblock


\bibitem[\protect\citeauthoryear{Hohman, Kahng, Pienta, and Chau}{Hohman et~al\mbox{.}}{2018}]%
        {hohman2018visual}
\bibfield{author}{\bibinfo{person}{Fred Hohman}, \bibinfo{person}{Minsuk Kahng}, \bibinfo{person}{Robert Pienta}, {and} \bibinfo{person}{Duen~Horng Chau}.} \bibinfo{year}{2018}\natexlab{}.
\newblock \showarticletitle{Visual analytics in deep learning: An interrogative survey for the next frontiers}.
\newblock \bibinfo{journal}{\emph{IEEE transactions on visualization and computer graphics}} \bibinfo{volume}{25}, \bibinfo{number}{8} (\bibinfo{year}{2018}), \bibinfo{pages}{2674--2693}.
\newblock


\bibitem[\protect\citeauthoryear{Hoque, Saquib, Billah, and Mueller}{Hoque et~al\mbox{.}}{2020}]%
        {hoque2020toward}
\bibfield{author}{\bibinfo{person}{Md~Naimul Hoque}, \bibinfo{person}{Nazmus Saquib}, \bibinfo{person}{Syed~Masum Billah}, {and} \bibinfo{person}{Klaus Mueller}.} \bibinfo{year}{2020}\natexlab{}.
\newblock \showarticletitle{Toward Interactively Balancing the Screen Time of Actors Based on Observable Phenotypic Traits in Live Telecast}.
\newblock  \bibinfo{volume}{4}, \bibinfo{number}{CSCW2}, Article \bibinfo{articleno}{154} (\bibinfo{date}{oct} \bibinfo{year}{2020}), \bibinfo{numpages}{18}~pages.
\newblock
\urldef\tempurl%
\url{https://doi.org/10.1145/3415225}
\showDOI{\tempurl}


\bibitem[\protect\citeauthoryear{Howe and Forbes}{Howe and Forbes}{2008}]%
        {howe2008re}
\bibfield{author}{\bibinfo{person}{Adele~E Howe} {and} \bibinfo{person}{Ryan~D Forbes}.} \bibinfo{year}{2008}\natexlab{}.
\newblock \showarticletitle{Re-considering neighborhood-based collaborative filtering parameters in the context of new data}. In \bibinfo{booktitle}{\emph{Proceedings of the 17th ACM conference on Information and knowledge management}}. \bibinfo{pages}{1481--1482}.
\newblock


\bibitem[\protect\citeauthoryear{Islam and Billah}{Islam and Billah}{2023}]%
        {islam2023spacex}
\bibfield{author}{\bibinfo{person}{Md~Touhidul Islam} {and} \bibinfo{person}{Syed~Masum Billah}.} \bibinfo{year}{2023}\natexlab{}.
\newblock \showarticletitle{SpaceX Mag: An Automatic, Scalable, and Rapid Space Compactor for Optimizing Smartphone App Interfaces for Low-Vision Users}.
\newblock \bibinfo{journal}{\emph{Proceedings of the ACM on Interactive, Mobile, Wearable and Ubiquitous Technologies}} \bibinfo{volume}{7}, \bibinfo{number}{2} (\bibinfo{year}{2023}), \bibinfo{pages}{1--36}.
\newblock


\bibitem[\protect\citeauthoryear{Islam, Kabir, Pearce, Reza, and Billah}{Islam et~al\mbox{.}}{2024a}]%
        {islam2024dataset}
\bibfield{author}{\bibinfo{person}{Md~Touhidul Islam}, \bibinfo{person}{Imran Kabir}, \bibinfo{person}{Elena~Ariel Pearce}, \bibinfo{person}{Md~Alimoor Reza}, {and} \bibinfo{person}{Syed~Masum Billah}.} \bibinfo{year}{2024}\natexlab{a}.
\newblock \bibinfo{title}{A Dataset for Crucial Object Recognition in Blind and Low-Vision Individuals' Navigation}.
\newblock
\newblock
\showeprint[arxiv]{2407.16777}~[cs.CV]
\urldef\tempurl%
\url{https://arxiv.org/abs/2407.16777}
\showURL{%
\tempurl}


\bibitem[\protect\citeauthoryear{Islam, Kabir, Pearce, Reza, and Billah}{Islam et~al\mbox{.}}{2024b}]%
        {islam2024identifying}
\bibfield{author}{\bibinfo{person}{Md~Touhidul Islam}, \bibinfo{person}{Imran Kabir}, \bibinfo{person}{Elene~Ariel Pearce}, \bibinfo{person}{Md~Alimoor Reza}, {and} \bibinfo{person}{Syed~Masum Billah}.} \bibinfo{year}{2024}\natexlab{b}.
\newblock \showarticletitle{Identifying Crucial Objects in Blind and Low-Vision Individuals' Navigation}. In \bibinfo{booktitle}{\emph{The 26th International ACM SIGACCESS Conference on Computers and Accessibility (ASSETS'24)}}. \bibinfo{publisher}{ACM}.
\newblock
\urldef\tempurl%
\url{https://doi.org/10.1145/3663548.3688538}
\showDOI{\tempurl}


\bibitem[\protect\citeauthoryear{Islam, Sojib, Kabir, Amit, Ruhul~Amin, and Billah}{Islam et~al\mbox{.}}{2024c}]%
        {islam2024wheelerdemo}
\bibfield{author}{\bibinfo{person}{Md~Touhidul Islam}, \bibinfo{person}{Noushad Sojib}, \bibinfo{person}{Imran Kabir}, \bibinfo{person}{Ashiqur~Rahman Amit}, \bibinfo{person}{Mohammad Ruhul~Amin}, {and} \bibinfo{person}{Syed~Masum Billah}.} \bibinfo{year}{2024}\natexlab{c}.
\newblock \showarticletitle{Demonstration of Wheeler: A Three-Wheeled Input Device for Usable, Efficient, and Versatile Non-Visual Interaction}. In \bibinfo{booktitle}{\emph{The 37th Annual ACM Symposium on User Interface Software and Technology (UIST Adjunct '24)}}. \bibinfo{publisher}{Association for Computing Machinery}, \bibinfo{address}{Pittsburgh, PA, USA}.
\newblock
\showISBNx{979-8-4007-0718-6/24/10}
\urldef\tempurl%
\url{https://doi.org/10.1145/3672539.3686749}
\showURL{%
\tempurl}


\bibitem[\protect\citeauthoryear{Islam, Sojib, Kabir, Amit, Ruhul~Amin, and Billah}{Islam et~al\mbox{.}}{2024d}]%
        {islam2024wheeler}
\bibfield{author}{\bibinfo{person}{Md~Touhidul Islam}, \bibinfo{person}{Noushad Sojib}, \bibinfo{person}{Imran Kabir}, \bibinfo{person}{Ashiqur~Rahman Amit}, \bibinfo{person}{Mohammad Ruhul~Amin}, {and} \bibinfo{person}{Syed~Masum Billah}.} \bibinfo{year}{2024}\natexlab{d}.
\newblock \showarticletitle{Wheeler: A Three-Wheeled Input Device for Usable, Efficient, and Versatile Non-Visual Interaction}. In \bibinfo{booktitle}{\emph{The 37th Annual ACM Symposium on User Interface Software and Technology}}. \bibinfo{publisher}{Association for Computing Machinery}, \bibinfo{address}{Pittsburgh, PA, USA}.
\newblock
\showISBNx{979-8-4007-0628-8/24/10}
\urldef\tempurl%
\url{https://doi.org/10.1145/3654777.3676396}
\showURL{%
\tempurl}


\bibitem[\protect\citeauthoryear{Jia, Gallagher, Saxena, and Chen}{Jia et~al\mbox{.}}{2014}]%
        {jia-3d-stability-pami}
\bibfield{author}{\bibinfo{person}{Zhaoyin Jia}, \bibinfo{person}{Andy Gallagher}, \bibinfo{person}{Ashutosh Saxena}, {and} \bibinfo{person}{Tsuhan Chen}.} \bibinfo{year}{2014}\natexlab{}.
\newblock \showarticletitle{3D Reasoning from Blocks to Stability}.
\newblock \bibinfo{journal}{\emph{IEEE Trans PAMI}} (\bibinfo{year}{2014}).
\newblock


\bibitem[\protect\citeauthoryear{Jin and Si}{Jin and Si}{2004}]%
        {jin2004study}
\bibfield{author}{\bibinfo{person}{Rong Jin} {and} \bibinfo{person}{Luo Si}.} \bibinfo{year}{2004}\natexlab{}.
\newblock \showarticletitle{A study of methods for normalizing user ratings in collaborative filtering}. In \bibinfo{booktitle}{\emph{Proceedings of the 27th annual international ACM SIGIR conference on Research and development in information retrieval}}. \bibinfo{pages}{568--569}.
\newblock


\bibitem[\protect\citeauthoryear{Johnson and Johnson}{Johnson and Johnson}{1986}]%
        {johnson1986cooperative}
\bibfield{author}{\bibinfo{person}{Roger~T Johnson} {and} \bibinfo{person}{David~W Johnson}.} \bibinfo{year}{1986}\natexlab{}.
\newblock \showarticletitle{Cooperative learning in the science classroom}.
\newblock \bibinfo{journal}{\emph{Science and children}} \bibinfo{volume}{24}, \bibinfo{number}{2} (\bibinfo{year}{1986}), \bibinfo{pages}{31--32}.
\newblock


\bibitem[\protect\citeauthoryear{Kabir, Reza, and Billah}{Kabir et~al\mbox{.}}{2025}]%
        {kabir2025logicrag}
\bibfield{author}{\bibinfo{person}{Imran Kabir}, \bibinfo{person}{Md~Alimoor Reza}, {and} \bibinfo{person}{Syed Billah}.} \bibinfo{year}{2025}\natexlab{}.
\newblock \showarticletitle{Logic-RAG: Augmenting Large Multimodal Models with Visual-Spatial Knowledge for Road Scene Understanding}. In \bibinfo{booktitle}{\emph{2025 IEEE International Conference on Robotics and Automation (ICRA)}}. IEEE.
\newblock


\bibitem[\protect\citeauthoryear{Kahng, Andrews, Kalro, and Chau}{Kahng et~al\mbox{.}}{2017}]%
        {kahng2017cti}
\bibfield{author}{\bibinfo{person}{Minsuk Kahng}, \bibinfo{person}{Pierre~Y Andrews}, \bibinfo{person}{Aditya Kalro}, {and} \bibinfo{person}{Duen~Horng Chau}.} \bibinfo{year}{2017}\natexlab{}.
\newblock \showarticletitle{A cti v is: Visual exploration of industry-scale deep neural network models}.
\newblock \bibinfo{journal}{\emph{IEEE transactions on visualization and computer graphics}} \bibinfo{volume}{24}, \bibinfo{number}{1} (\bibinfo{year}{2017}), \bibinfo{pages}{88--97}.
\newblock


\bibitem[\protect\citeauthoryear{Kamath, Hessel, and Chang}{Kamath et~al\mbox{.}}{2023}]%
        {kamath2023s}
\bibfield{author}{\bibinfo{person}{Amita Kamath}, \bibinfo{person}{Jack Hessel}, {and} \bibinfo{person}{Kai-Wei Chang}.} \bibinfo{year}{2023}\natexlab{}.
\newblock \showarticletitle{What's" up" with vision-language models? Investigating their struggle with spatial reasoning}.
\newblock \bibinfo{journal}{\emph{arXiv preprint arXiv:2310.19785}} (\bibinfo{year}{2023}).
\newblock


\bibitem[\protect\citeauthoryear{Karaoguz and Jensfelt}{Karaoguz and Jensfelt}{2019}]%
        {karaoguz2019object}
\bibfield{author}{\bibinfo{person}{Hakan Karaoguz} {and} \bibinfo{person}{Patric Jensfelt}.} \bibinfo{year}{2019}\natexlab{}.
\newblock \showarticletitle{Object detection approach for robot grasp detection}. In \bibinfo{booktitle}{\emph{2019 International Conference on Robotics and Automation (ICRA)}}. IEEE, \bibinfo{pages}{4953--4959}.
\newblock


\bibitem[\protect\citeauthoryear{Li, Ouyang, Sheng, Zeng, and Wang}{Li et~al\mbox{.}}{2019}]%
        {li2019gs3d}
\bibfield{author}{\bibinfo{person}{Buyu Li}, \bibinfo{person}{Wanli Ouyang}, \bibinfo{person}{Lu Sheng}, \bibinfo{person}{Xingyu Zeng}, {and} \bibinfo{person}{Xiaogang Wang}.} \bibinfo{year}{2019}\natexlab{}.
\newblock \showarticletitle{Gs3d: An efficient 3d object detection framework for autonomous driving}. In \bibinfo{booktitle}{\emph{Proceedings of the IEEE/CVF Conference on Computer Vision and Pattern Recognition}}. \bibinfo{pages}{1019--1028}.
\newblock


\bibitem[\protect\citeauthoryear{Li, Li, Le, Wang, Savarese, and Hoi}{Li et~al\mbox{.}}{2022a}]%
        {li2022lavis}
\bibfield{author}{\bibinfo{person}{Dongxu Li}, \bibinfo{person}{Junnan Li}, \bibinfo{person}{Hung Le}, \bibinfo{person}{Guangsen Wang}, \bibinfo{person}{Silvio Savarese}, {and} \bibinfo{person}{Steven~CH Hoi}.} \bibinfo{year}{2022}\natexlab{a}.
\newblock \showarticletitle{{Lavis}: A library for language-vision intelligence}.
\newblock \bibinfo{journal}{\emph{arXiv preprint arXiv:2209.09019}} (\bibinfo{year}{2022}).
\newblock


\bibitem[\protect\citeauthoryear{Li, Li, Xiong, and Hoi}{Li et~al\mbox{.}}{2022b}]%
        {li2022blip}
\bibfield{author}{\bibinfo{person}{Junnan Li}, \bibinfo{person}{Dongxu Li}, \bibinfo{person}{Caiming Xiong}, {and} \bibinfo{person}{Steven Hoi}.} \bibinfo{year}{2022}\natexlab{b}.
\newblock \showarticletitle{Blip: Bootstrapping language-image pre-training for unified vision-language understanding and generation}. In \bibinfo{booktitle}{\emph{International Conference on Machine Learning}}. PMLR, \bibinfo{pages}{12888--12900}.
\newblock


\bibitem[\protect\citeauthoryear{Liu, Emerson, and Collier}{Liu et~al\mbox{.}}{2023a}]%
        {liu2023visualspatial}
\bibfield{author}{\bibinfo{person}{Fangyu Liu}, \bibinfo{person}{Guy Emerson}, {and} \bibinfo{person}{Nigel Collier}.} \bibinfo{year}{2023}\natexlab{a}.
\newblock \showarticletitle{Visual spatial reasoning}.
\newblock \bibinfo{journal}{\emph{Transactions of the Association for Computational Linguistics}}  \bibinfo{volume}{11} (\bibinfo{year}{2023}), \bibinfo{pages}{635--651}.
\newblock


\bibitem[\protect\citeauthoryear{Liu, Li, Wu, and Lee}{Liu et~al\mbox{.}}{2023b}]%
        {liu2023visual}
\bibfield{author}{\bibinfo{person}{Haotian Liu}, \bibinfo{person}{Chunyuan Li}, \bibinfo{person}{Qingyang Wu}, {and} \bibinfo{person}{Yong~Jae Lee}.} \bibinfo{year}{2023}\natexlab{b}.
\newblock \showarticletitle{Visual instruction tuning}.
\newblock \bibinfo{journal}{\emph{arXiv preprint arXiv:2304.08485}} (\bibinfo{year}{2023}).
\newblock


\bibitem[\protect\citeauthoryear{Liu, Li, Wu, and Lee}{Liu et~al\mbox{.}}{2024}]%
        {liu2024visual}
\bibfield{author}{\bibinfo{person}{Haotian Liu}, \bibinfo{person}{Chunyuan Li}, \bibinfo{person}{Qingyang Wu}, {and} \bibinfo{person}{Yong~Jae Lee}.} \bibinfo{year}{2024}\natexlab{}.
\newblock \showarticletitle{Visual instruction tuning}.
\newblock \bibinfo{journal}{\emph{Advances in neural information processing systems}}  \bibinfo{volume}{36} (\bibinfo{year}{2024}).
\newblock


\bibitem[\protect\citeauthoryear{Liu, Shi, Cao, Zhu, and Liu}{Liu et~al\mbox{.}}{2017a}]%
        {liu2017analyzing}
\bibfield{author}{\bibinfo{person}{Mengchen Liu}, \bibinfo{person}{Jiaxin Shi}, \bibinfo{person}{Kelei Cao}, \bibinfo{person}{Jun Zhu}, {and} \bibinfo{person}{Shixia Liu}.} \bibinfo{year}{2017}\natexlab{a}.
\newblock \showarticletitle{Analyzing the training processes of deep generative models}.
\newblock \bibinfo{journal}{\emph{IEEE transactions on visualization and computer graphics}} \bibinfo{volume}{24}, \bibinfo{number}{1} (\bibinfo{year}{2017}), \bibinfo{pages}{77--87}.
\newblock


\bibitem[\protect\citeauthoryear{Liu, Wang, Liu, and Zhu}{Liu et~al\mbox{.}}{2017b}]%
        {liu2017towards}
\bibfield{author}{\bibinfo{person}{Shixia Liu}, \bibinfo{person}{Xiting Wang}, \bibinfo{person}{Mengchen Liu}, {and} \bibinfo{person}{Jun Zhu}.} \bibinfo{year}{2017}\natexlab{b}.
\newblock \showarticletitle{Towards better analysis of machine learning models: A visual analytics perspective}.
\newblock \bibinfo{journal}{\emph{Visual Informatics}} \bibinfo{volume}{1}, \bibinfo{number}{1} (\bibinfo{year}{2017}), \bibinfo{pages}{48--56}.
\newblock


\bibitem[\protect\citeauthoryear{Maaz, Rasheed, Khan, and Khan}{Maaz et~al\mbox{.}}{2023}]%
        {maaz2023video}
\bibfield{author}{\bibinfo{person}{Muhammad Maaz}, \bibinfo{person}{Hanoona Rasheed}, \bibinfo{person}{Salman Khan}, {and} \bibinfo{person}{Fahad~Shahbaz Khan}.} \bibinfo{year}{2023}\natexlab{}.
\newblock \showarticletitle{Video-chatgpt: Towards detailed video understanding via large vision and language models}.
\newblock \bibinfo{journal}{\emph{arXiv preprint arXiv:2306.05424}} (\bibinfo{year}{2023}).
\newblock


\bibitem[\protect\citeauthoryear{Microsoft}{Microsoft}{2017}]%
        {surface_dial}
\bibfield{author}{\bibinfo{person}{Microsoft}.} \bibinfo{year}{2017}\natexlab{}.
\newblock \bibinfo{title}{Surface Dial}.
\newblock
\newblock
\urldef\tempurl%
\url{https://www.microsoft.com/en-us/surface/accessories/surface-dial}
\showURL{%
\tempurl}


\bibitem[\protect\citeauthoryear{Newman}{Newman}{2004}]%
        {newman2004coauthorship}
\bibfield{author}{\bibinfo{person}{Mark~EJ Newman}.} \bibinfo{year}{2004}\natexlab{}.
\newblock \showarticletitle{Coauthorship networks and patterns of scientific collaboration}.
\newblock \bibinfo{journal}{\emph{Proceedings of the national academy of sciences}} \bibinfo{volume}{101}, \bibinfo{number}{suppl\_1} (\bibinfo{year}{2004}), \bibinfo{pages}{5200--5205}.
\newblock


\bibitem[\protect\citeauthoryear{Nison}{Nison}{2001}]%
        {nison2001japanese}
\bibfield{author}{\bibinfo{person}{Steve Nison}.} \bibinfo{year}{2001}\natexlab{}.
\newblock \bibinfo{booktitle}{\emph{Japanese candlestick charting techniques: a contemporary guide to the ancient investment techniques of the Far East}}.
\newblock \bibinfo{publisher}{Penguin}.
\newblock


\bibitem[\protect\citeauthoryear{Norman and Draper}{Norman and Draper}{1986}]%
        {norman1986user}
\bibfield{author}{\bibinfo{person}{Donald~A Norman} {and} \bibinfo{person}{Stephen~W Draper}.} \bibinfo{year}{1986}\natexlab{}.
\newblock \bibinfo{booktitle}{\emph{User centered system design; new perspectives on human-computer interaction}}.
\newblock \bibinfo{publisher}{L. Erlbaum Associates Inc.}
\newblock


\bibitem[\protect\citeauthoryear{OpenAI}{OpenAI}{[n.d.]}]%
        {gptv_system_card}
\bibfield{author}{\bibinfo{person}{OpenAI}.} \bibinfo{year}{[n.d.]}\natexlab{}.
\newblock \bibinfo{title}{GPTV Sysmtem Card}.
\newblock
\newblock
\urldef\tempurl%
\url{https://cdn.openai.com/papers/GPTV_System_Card.pdf}
\showURL{%
\tempurl}


\bibitem[\protect\citeauthoryear{OpenAI}{OpenAI}{2023a}]%
        {openai2023gpt4}
\bibfield{author}{\bibinfo{person}{OpenAI}.} \bibinfo{year}{2023}\natexlab{a}.
\newblock \bibinfo{booktitle}{\emph{GPT-4 Technical Report}}.
\newblock
\showeprint[arxiv]{2303.08774v2}
\urldef\tempurl%
\url{https://arxiv.org/abs/2303.08774v2}
\showURL{%
\tempurl}


\bibitem[\protect\citeauthoryear{OpenAI}{OpenAI}{2023b}]%
        {openai2023gpt4vsc}
\bibfield{author}{\bibinfo{person}{OpenAI}.} \bibinfo{year}{2023}\natexlab{b}.
\newblock \bibinfo{booktitle}{\emph{GPT-4V(ision) System Card}}.
\newblock
\urldef\tempurl%
\url{https://cdn.openai.com/papers/GPTV_System_Card.pdf}
\showURL{%
\tempurl}


\bibitem[\protect\citeauthoryear{OpenAI}{OpenAI}{2023c}]%
        {openai2023gpt4vtwa}
\bibfield{author}{\bibinfo{person}{OpenAI}.} \bibinfo{year}{2023}\natexlab{c}.
\newblock \bibinfo{booktitle}{\emph{GPT-4V(ision) technical work and authors}}.
\newblock
\urldef\tempurl%
\url{https://openai.com/contributions/gpt-4v}
\showURL{%
\tempurl}


\bibitem[\protect\citeauthoryear{Page, Brin, Motwani, and Winograd}{Page et~al\mbox{.}}{1999}]%
        {page1999pagerank}
\bibfield{author}{\bibinfo{person}{Lawrence Page}, \bibinfo{person}{Sergey Brin}, \bibinfo{person}{Rajeev Motwani}, {and} \bibinfo{person}{Terry Winograd}.} \bibinfo{year}{1999}\natexlab{}.
\newblock \bibinfo{booktitle}{\emph{The PageRank citation ranking: Bringing order to the web.}}
\newblock \bibinfo{type}{{T}echnical {R}eport}. \bibinfo{institution}{Stanford infolab}.
\newblock


\bibitem[\protect\citeauthoryear{Paul, Chowdhury, Nicolescu, Nicolescu, and Feil-Seifer}{Paul et~al\mbox{.}}{2021}]%
        {paul2021object}
\bibfield{author}{\bibinfo{person}{Shuvo~Kumar Paul}, \bibinfo{person}{Muhammed~Tawfiq Chowdhury}, \bibinfo{person}{Mircea Nicolescu}, \bibinfo{person}{Monica Nicolescu}, {and} \bibinfo{person}{David Feil-Seifer}.} \bibinfo{year}{2021}\natexlab{}.
\newblock \showarticletitle{Object detection and pose estimation from rgb and depth data for real-time, adaptive robotic grasping}.
\newblock In \bibinfo{booktitle}{\emph{Advances in Computer Vision and Computational Biology: Proceedings from IPCV'20, HIMS'20, BIOCOMP'20, and BIOENG'20}}. \bibinfo{publisher}{Springer}, \bibinfo{pages}{121--142}.
\newblock


\bibitem[\protect\citeauthoryear{Powers}{Powers}{2020}]%
        {powers2020evaluation}
\bibfield{author}{\bibinfo{person}{David~MW Powers}.} \bibinfo{year}{2020}\natexlab{}.
\newblock \showarticletitle{Evaluation: from precision, recall and F-measure to ROC, informedness, markedness and correlation}.
\newblock \bibinfo{journal}{\emph{arXiv preprint arXiv:2010.16061}} (\bibinfo{year}{2020}).
\newblock


\bibitem[\protect\citeauthoryear{Qadir, Islam, and Al-Fuqaha}{Qadir et~al\mbox{.}}{2022}]%
        {qadir2022toward}
\bibfield{author}{\bibinfo{person}{Junaid Qadir}, \bibinfo{person}{Mohammad~Qamar Islam}, {and} \bibinfo{person}{Ala Al-Fuqaha}.} \bibinfo{year}{2022}\natexlab{}.
\newblock \showarticletitle{Toward accountable human-centered AI: rationale and promising directions}.
\newblock \bibinfo{journal}{\emph{Journal of Information, Communication and Ethics in Society}} \bibinfo{volume}{20}, \bibinfo{number}{2} (\bibinfo{year}{2022}), \bibinfo{pages}{329--342}.
\newblock


\bibitem[\protect\citeauthoryear{Qi, Cun, Zhang, Lei, Wang, Shan, and Chen}{Qi et~al\mbox{.}}{2023}]%
        {qi2023fatezero}
\bibfield{author}{\bibinfo{person}{Chenyang Qi}, \bibinfo{person}{Xiaodong Cun}, \bibinfo{person}{Yong Zhang}, \bibinfo{person}{Chenyang Lei}, \bibinfo{person}{Xintao Wang}, \bibinfo{person}{Ying Shan}, {and} \bibinfo{person}{Qifeng Chen}.} \bibinfo{year}{2023}\natexlab{}.
\newblock \showarticletitle{Fatezero: Fusing attentions for zero-shot text-based video editing}.
\newblock \bibinfo{journal}{\emph{arXiv preprint arXiv:2303.09535}} (\bibinfo{year}{2023}).
\newblock


\bibitem[\protect\citeauthoryear{Ren, Amershi, Lee, Suh, and Williams}{Ren et~al\mbox{.}}{2016}]%
        {ren2016squares}
\bibfield{author}{\bibinfo{person}{Donghao Ren}, \bibinfo{person}{Saleema Amershi}, \bibinfo{person}{Bongshin Lee}, \bibinfo{person}{Jina Suh}, {and} \bibinfo{person}{Jason~D Williams}.} \bibinfo{year}{2016}\natexlab{}.
\newblock \showarticletitle{Squares: Supporting interactive performance analysis for multiclass classifiers}.
\newblock \bibinfo{journal}{\emph{IEEE transactions on visualization and computer graphics}} \bibinfo{volume}{23}, \bibinfo{number}{1} (\bibinfo{year}{2016}), \bibinfo{pages}{61--70}.
\newblock


\bibitem[\protect\citeauthoryear{Riedl}{Riedl}{2019}]%
        {riedl2019human}
\bibfield{author}{\bibinfo{person}{Mark~O Riedl}.} \bibinfo{year}{2019}\natexlab{}.
\newblock \showarticletitle{Human-centered artificial intelligence and machine learning}.
\newblock \bibinfo{journal}{\emph{Human behavior and emerging technologies}} \bibinfo{volume}{1}, \bibinfo{number}{1} (\bibinfo{year}{2019}), \bibinfo{pages}{33--36}.
\newblock


\bibitem[\protect\citeauthoryear{Sanchez and Wiley}{Sanchez and Wiley}{2009}]%
        {sanchez2009scroll}
\bibfield{author}{\bibinfo{person}{Christopher~A Sanchez} {and} \bibinfo{person}{Jennifer Wiley}.} \bibinfo{year}{2009}\natexlab{}.
\newblock \showarticletitle{To scroll or not to scroll: Scrolling, working memory capacity, and comprehending complex texts}.
\newblock \bibinfo{journal}{\emph{Human Factors}} \bibinfo{volume}{51}, \bibinfo{number}{5} (\bibinfo{year}{2009}), \bibinfo{pages}{730--738}.
\newblock


\bibitem[\protect\citeauthoryear{Simons and Levin}{Simons and Levin}{1997}]%
        {simons1997change}
\bibfield{author}{\bibinfo{person}{Daniel~J Simons} {and} \bibinfo{person}{Daniel~T Levin}.} \bibinfo{year}{1997}\natexlab{}.
\newblock \showarticletitle{Change blindness}.
\newblock \bibinfo{journal}{\emph{Trends in cognitive sciences}} \bibinfo{volume}{1}, \bibinfo{number}{7} (\bibinfo{year}{1997}), \bibinfo{pages}{261--267}.
\newblock


\bibitem[\protect\citeauthoryear{Thrush, Jiang, Bartolo, Singh, Williams, Kiela, and Ross}{Thrush et~al\mbox{.}}{2022}]%
        {thrush2022winoground}
\bibfield{author}{\bibinfo{person}{Tristan Thrush}, \bibinfo{person}{Ryan Jiang}, \bibinfo{person}{Max Bartolo}, \bibinfo{person}{Amanpreet Singh}, \bibinfo{person}{Adina Williams}, \bibinfo{person}{Douwe Kiela}, {and} \bibinfo{person}{Candace Ross}.} \bibinfo{year}{2022}\natexlab{}.
\newblock \showarticletitle{Winoground: Probing vision and language models for visio-linguistic compositionality}. In \bibinfo{booktitle}{\emph{Proceedings of the IEEE/CVF Conference on Computer Vision and Pattern Recognition}}. \bibinfo{pages}{5238--5248}.
\newblock


\bibitem[\protect\citeauthoryear{Todorovic}{Todorovic}{2008}]%
        {todorovic2008gestalt}
\bibfield{author}{\bibinfo{person}{Dejan Todorovic}.} \bibinfo{year}{2008}\natexlab{}.
\newblock \showarticletitle{Gestalt principles}.
\newblock \bibinfo{journal}{\emph{Scholarpedia}} \bibinfo{volume}{3}, \bibinfo{number}{12} (\bibinfo{year}{2008}), \bibinfo{pages}{5345}.
\newblock


\bibitem[\protect\citeauthoryear{Treisman}{Treisman}{1985}]%
        {treisman1985preattentive}
\bibfield{author}{\bibinfo{person}{Anne Treisman}.} \bibinfo{year}{1985}\natexlab{}.
\newblock \showarticletitle{Preattentive processing in vision}.
\newblock \bibinfo{journal}{\emph{Computer vision, graphics, and image processing}} \bibinfo{volume}{31}, \bibinfo{number}{2} (\bibinfo{year}{1985}), \bibinfo{pages}{156--177}.
\newblock


\bibitem[\protect\citeauthoryear{Treisman and Gelade}{Treisman and Gelade}{1980}]%
        {treisman1980feature}
\bibfield{author}{\bibinfo{person}{Anne~M Treisman} {and} \bibinfo{person}{Garry Gelade}.} \bibinfo{year}{1980}\natexlab{}.
\newblock \showarticletitle{A feature-integration theory of attention}.
\newblock \bibinfo{journal}{\emph{Cognitive psychology}} \bibinfo{volume}{12}, \bibinfo{number}{1} (\bibinfo{year}{1980}), \bibinfo{pages}{97--136}.
\newblock


\bibitem[\protect\citeauthoryear{Wang, Bochkovskiy, and Liao}{Wang et~al\mbox{.}}{2023a}]%
        {wang2023yolov7}
\bibfield{author}{\bibinfo{person}{Chien-Yao Wang}, \bibinfo{person}{Alexey Bochkovskiy}, {and} \bibinfo{person}{Hong-Yuan~Mark Liao}.} \bibinfo{year}{2023}\natexlab{a}.
\newblock \showarticletitle{YOLOv7: Trainable bag-of-freebies sets new state-of-the-art for real-time object detectors}. In \bibinfo{booktitle}{\emph{Proceedings of the IEEE/CVF Conference on Computer Vision and Pattern Recognition}}. \bibinfo{pages}{7464--7475}.
\newblock


\bibitem[\protect\citeauthoryear{Wang, Zhang, Wang, Cao, Zhou, Zhang, Liu, Fan, and Tian}{Wang et~al\mbox{.}}{2023b}]%
        {wang2023human}
\bibfield{author}{\bibinfo{person}{Liuping Wang}, \bibinfo{person}{Zhan Zhang}, \bibinfo{person}{Dakuo Wang}, \bibinfo{person}{Weidan Cao}, \bibinfo{person}{Xiaomu Zhou}, \bibinfo{person}{Ping Zhang}, \bibinfo{person}{Jianxing Liu}, \bibinfo{person}{Xiangmin Fan}, {and} \bibinfo{person}{Feng Tian}.} \bibinfo{year}{2023}\natexlab{b}.
\newblock \showarticletitle{Human-centered design and evaluation of AI-empowered clinical decision support systems: a systematic review}.
\newblock \bibinfo{journal}{\emph{Frontiers in Computer Science}}  \bibinfo{volume}{5} (\bibinfo{year}{2023}), \bibinfo{pages}{1187299}.
\newblock


\bibitem[\protect\citeauthoryear{Ward, Grinstein, and Keim}{Ward et~al\mbox{.}}{2010}]%
        {ward2010interactive}
\bibfield{author}{\bibinfo{person}{Matthew~O Ward}, \bibinfo{person}{Georges Grinstein}, {and} \bibinfo{person}{Daniel Keim}.} \bibinfo{year}{2010}\natexlab{}.
\newblock \bibinfo{booktitle}{\emph{Interactive Data Disualization: Foundations, Techniques, and Applications}}.
\newblock \bibinfo{publisher}{CRC Press}.
\newblock


\bibitem[\protect\citeauthoryear{West, Lu, Dziri, Brahman, Li, Hwang, Jiang, Fisher, Ravichander, Chandu, et~al\mbox{.}}{West et~al\mbox{.}}{2023}]%
        {west2023generative}
\bibfield{author}{\bibinfo{person}{Peter West}, \bibinfo{person}{Ximing Lu}, \bibinfo{person}{Nouha Dziri}, \bibinfo{person}{Faeze Brahman}, \bibinfo{person}{Linjie Li}, \bibinfo{person}{Jena~D Hwang}, \bibinfo{person}{Liwei Jiang}, \bibinfo{person}{Jillian Fisher}, \bibinfo{person}{Abhilasha Ravichander}, \bibinfo{person}{Khyathi Chandu}, {et~al\mbox{.}}} \bibinfo{year}{2023}\natexlab{}.
\newblock \showarticletitle{THE GENERATIVE AI PARADOX:“What It Can Create, It May Not Understand”}. In \bibinfo{booktitle}{\emph{The Twelfth International Conference on Learning Representations}}.
\newblock


\bibitem[\protect\citeauthoryear{Xie, Yu, Zhang, Billah, Lee, and Carroll}{Xie et~al\mbox{.}}{2025}]%
        {xie2025beyond}
\bibfield{author}{\bibinfo{person}{Jingyi Xie}, \bibinfo{person}{Rui Yu}, \bibinfo{person}{He Zhang}, \bibinfo{person}{Syed~Masum Billah}, \bibinfo{person}{Sooyeon Lee}, {and} \bibinfo{person}{John~M Carroll}.} \bibinfo{year}{2025}\natexlab{}.
\newblock \showarticletitle{Beyond Visual Perception: Insights from Smartphone Interaction of Visually Impaired Users with Large Multimodal Models}. In \bibinfo{booktitle}{\emph{Proceedings of the 2025 CHI Conference on Human Factors in Computing Systems}}. \bibinfo{pages}{1--17}.
\newblock


\bibitem[\protect\citeauthoryear{Xie, Yu, Zhang, Lee, Billah, and Carroll}{Xie et~al\mbox{.}}{2024}]%
        {xie2024emerging}
\bibfield{author}{\bibinfo{person}{Jingyi Xie}, \bibinfo{person}{Rui Yu}, \bibinfo{person}{He Zhang}, \bibinfo{person}{Sooyeon Lee}, \bibinfo{person}{Syed~Masum Billah}, {and} \bibinfo{person}{John~M Carroll}.} \bibinfo{year}{2024}\natexlab{}.
\newblock \showarticletitle{Emerging practices for large multimodal model (lmm) assistance for people with visual impairments: Implications for design}.
\newblock \bibinfo{journal}{\emph{arXiv preprint arXiv:2407.08882}} (\bibinfo{year}{2024}).
\newblock


\bibitem[\protect\citeauthoryear{Yang, Jia, Li, and Yan}{Yang et~al\mbox{.}}{2023}]%
        {yang2023survey}
\bibfield{author}{\bibinfo{person}{Zhenjie Yang}, \bibinfo{person}{Xiaosong Jia}, \bibinfo{person}{Hongyang Li}, {and} \bibinfo{person}{Junchi Yan}.} \bibinfo{year}{2023}\natexlab{}.
\newblock \showarticletitle{A survey of large language models for autonomous driving}.
\newblock \bibinfo{journal}{\emph{arXiv preprint arXiv:2311.01043}} (\bibinfo{year}{2023}).
\newblock


\bibitem[\protect\citeauthoryear{Yuksekgonul, Bianchi, Kalluri, Jurafsky, and Zou}{Yuksekgonul et~al\mbox{.}}{2022}]%
        {yuksekgonul2022and}
\bibfield{author}{\bibinfo{person}{Mert Yuksekgonul}, \bibinfo{person}{Federico Bianchi}, \bibinfo{person}{Pratyusha Kalluri}, \bibinfo{person}{Dan Jurafsky}, {and} \bibinfo{person}{James Zou}.} \bibinfo{year}{2022}\natexlab{}.
\newblock \showarticletitle{When and why vision-language models behave like bag-of-words models, and what to do about it}.
\newblock \bibinfo{journal}{\emph{arXiv preprint arXiv:2210.01936}}  \bibinfo{volume}{5} (\bibinfo{year}{2022}).
\newblock


\bibitem[\protect\citeauthoryear{Zhang, Falletta, Xie, Yu, Lee, Billah, and Carroll}{Zhang et~al\mbox{.}}{2025}]%
        {zhang2025enhancing}
\bibfield{author}{\bibinfo{person}{He Zhang}, \bibinfo{person}{Nicholas~J Falletta}, \bibinfo{person}{Jingyi Xie}, \bibinfo{person}{Rui Yu}, \bibinfo{person}{Sooyeon Lee}, \bibinfo{person}{Syed~Masum Billah}, {and} \bibinfo{person}{John~M Carroll}.} \bibinfo{year}{2025}\natexlab{}.
\newblock \showarticletitle{Enhancing the Travel Experience for People with Visual Impairments through Multimodal Interaction: NaviGPT, A Real-Time AI-Driven Mobile Navigation System}. In \bibinfo{booktitle}{\emph{Companion Proceedings of the 2025 ACM International Conference on Supporting Group Work}}. \bibinfo{pages}{29--35}.
\newblock


\bibitem[\protect\citeauthoryear{Zhang, Wang, Molino, Li, and Ebert}{Zhang et~al\mbox{.}}{2018b}]%
        {zhang2018manifold}
\bibfield{author}{\bibinfo{person}{Jiawei Zhang}, \bibinfo{person}{Yang Wang}, \bibinfo{person}{Piero Molino}, \bibinfo{person}{Lezhi Li}, {and} \bibinfo{person}{David~S Ebert}.} \bibinfo{year}{2018}\natexlab{b}.
\newblock \showarticletitle{Manifold: A model-agnostic framework for interpretation and diagnosis of machine learning models}.
\newblock \bibinfo{journal}{\emph{IEEE transactions on visualization and computer graphics}} \bibinfo{volume}{25}, \bibinfo{number}{1} (\bibinfo{year}{2018}), \bibinfo{pages}{364--373}.
\newblock


\bibitem[\protect\citeauthoryear{Zhang, Isola, Efros, Shechtman, and Wang}{Zhang et~al\mbox{.}}{2018a}]%
        {zhang2018unreasonable}
\bibfield{author}{\bibinfo{person}{Richard Zhang}, \bibinfo{person}{Phillip Isola}, \bibinfo{person}{Alexei~A Efros}, \bibinfo{person}{Eli Shechtman}, {and} \bibinfo{person}{Oliver Wang}.} \bibinfo{year}{2018}\natexlab{a}.
\newblock \showarticletitle{The unreasonable effectiveness of deep features as a perceptual metric}. In \bibinfo{booktitle}{\emph{Proceedings of the IEEE conference on computer vision and pattern recognition}}. \bibinfo{pages}{586--595}.
\newblock


\bibitem[\protect\citeauthoryear{Zhou, Liu, Zagar, Yurtsever, and Knoll}{Zhou et~al\mbox{.}}{2023}]%
        {zhou2023vision}
\bibfield{author}{\bibinfo{person}{Xingcheng Zhou}, \bibinfo{person}{Mingyu Liu}, \bibinfo{person}{Bare~Luka Zagar}, \bibinfo{person}{Ekim Yurtsever}, {and} \bibinfo{person}{Alois~C Knoll}.} \bibinfo{year}{2023}\natexlab{}.
\newblock \showarticletitle{Vision language models in autonomous driving and intelligent transportation systems}.
\newblock \bibinfo{journal}{\emph{arXiv preprint arXiv:2310.14414}} (\bibinfo{year}{2023}).
\newblock


\bibitem[\protect\citeauthoryear{Zimmerman, Forlizzi, and Evenson}{Zimmerman et~al\mbox{.}}{2007}]%
        {research_through_design}
\bibfield{author}{\bibinfo{person}{John Zimmerman}, \bibinfo{person}{Jodi Forlizzi}, {and} \bibinfo{person}{Shelley Evenson}.} \bibinfo{year}{2007}\natexlab{}.
\newblock \bibinfo{booktitle}{\emph{Research through Design as a Method for Interaction Design Research in HCI}}.
\newblock \bibinfo{publisher}{Association for Computing Machinery}, \bibinfo{address}{New York, NY, USA}, \bibinfo{pages}{493–502}.
\newblock
\showISBNx{9781595935939}
\urldef\tempurl%
\url{https://doi.org/10.1145/1240624.1240704}
\showURL{%
\tempurl}


\end{thebibliography}

% \pagebreak
\appendix
% \clearpage
\section{Appendix: Summary of Dataset Videos}
\label{app:dataset}

\subsection{Video Analysis}
\label{app:video-analysis}
To analyze the 21 collected videos, we divided each video into smaller clips, ranging from 5 to 95 seconds, resulting in 31 video segments. 
Each segment focuses on the presence of objects relevant to navigation on roads and sidewalks. Table~\ref{table:dataset} summarizes the number of segments created from each video.

Using the \textit{Katna} keyframe extraction tool\footnote{\url{https://katna.readthedocs.io/en/latest/}}, we further divided these video segments into keyframes. 
Keyframes serve as representative frames summarizing the video content, accounting for scene transitions, lighting changes, and activities. 
The number of keyframes extracted per segment ranged from three to ninety-three. 
We then manually annotated a subset of these keyframes to indicate the presence or absence of objects from our finalized list.

\begin{table*}[!t]
    \begin{center}
    \small{
        \begin{tabular}{l C{5cm}  C{1.0cm} C{0.50cm} C{1.0cm} C{0.75cm} C{1.25cm} C{3.75cm}}
            \toprule
             \textbf{ID} & 
             \textbf{Title/Context}  &
             {\textbf{Duration}} &
             {\textbf{\# Segments}} &
             {\textbf{\# Annotated Seg.}} &
             \textbf{Year} &
             \textbf{Location} &
             \textbf{URL} \\
             \toprule
            %  \midrule
            % 
             V1 & Blind Man Walking & 2:24  & 5 & 2 & 2011 & London & \url{https://youtu.be/RmsoHyMRtbg}  \\ \hline
             \rowcolor{gray!10} 
             V2 & following a blind person for a day | JAYKEEOUT & 7:02 & 1& 1 &  2021 & Seoul & \url{https://youtu.be/dPisedvLKQQ} \\ \hline
             V3 & Orientation \& Mobility for the Blind-1* & 0:00-10:00  & 8 & 2 &  2012 & --- & \url{https://youtu.be/Gkf5tEbP-oo}\\ 
             \hline
             \rowcolor{gray!10} 
             V4 & Orientation \& Mobility for the Blind-2* & 10:01-19:10 & 4 & 3 & 2012 & --- & \url{https://youtu.be/Gkf5tEbP-oo?t=602} \\ 
             \hline
             V5 & My First Blind Cane Adventure to Get Coffee | Did I Succeed or Give Up* & 10:00 & 3 & 1 & 2019 & Caribbean Cruise Ship & \url{https://youtu.be/SZM-Le6MEE0} \\ 
             \hline              
             \rowcolor{gray!10} 
             V6 & Using A White Cane | Legally Blind* & 10:00 & 2 & 1 & 2018 & --- & \url{https://youtu.be/TxUxbXyh7Y4} \\ 
             \hline
             V7 & How a Blind Person Uses a Cane & 4:18 & 4 & 1 & 2013 & --- & \url{https://youtu.be/xi0JMS1rulo} \\ 
             \hline             
             \rowcolor{gray!10}              
             V8 & Orientation mobility & 9:36 & 2 & 1 & 2022 & --- & \url{https://youtu.be/6u53Q7IvVIY} \\ 
             \hline                           
             V9 & TAKING THE METRO AND WALKING THROUGH MADRID ALONE AND BLIND-1* & 9:19 & 4 & 1 & 2020 & Madrid & \url{https://youtu.be/Vx3-ltp9p-Y} \\ 
             \hline                                        
             \rowcolor{gray!10}  
             V10 & TAKING THE METRO AND WALKING THROUGH MADRID ALONE AND BLIND-2* & 10:00 - 19:00 & 1 & 1 & 2020 & Madrid & \url{https://youtu.be/Vx3-ltp9p-Y?t=600} \\ 
             \hline                           
             V11 & Mobility and Orientation Training for Young People with Vision Impairment & 5:48 & 3  & 1 & 2019 & Edinburgh & \url{https://youtu.be/u-3GlbJ5RMc} \\ 
             \hline                           
             \rowcolor{gray!10}  
             V12 & Mobility and Orientation & 8:49 & 4 & 1 & 2018 & New York City & \url{https://vimeo.com/296488214} \\ 
             \hline                           
             V13 & Traveling with the white cane & 2:14 & 3 & 1 & 2009 & Maryland & \url{https://vimeo.com/2851243} \\ 
             \hline              
             \rowcolor{gray!10} 
             V14 & Blindness Awareness Month - Orientation and Mobility with ELC and 1st Grade Students & 5:52 & 5 & 2 & 2022 & --- & \url{https://vimeo.com/758153786} \\ 
             \hline              
             V15 & The White Cane documentary & 5:40 & 3 & 1 & 2021 & --- & \url{https://vimeo.com/497359578} \\ 
             \hline              
             \rowcolor{gray!10} 
             V16 & Craig Eckhardt takes the subway on Vimeo	& 4:43 & 4 & 1 & 2010 & New York & \url{https://vimeo.com/17293270} \\ 
             \hline              
             V17 & Guide Techniques for people who are blind or visually impaired* & 10:00 & 3  & 2 & 2015 & --- & \url{https://youtu.be/iJfxkBOekvs} \\ 
             \hline              
             \rowcolor{gray!10} 
             V18 & Russia: Blind Commuter Faces Obstacles Every Day & 3:20 & 6 & 2 & 2013 & Moscow & \url{https://youtu.be/20W2ckx-BcE} \\ 
             \hline              
             V19 & The ``Challenges'' you may not know about ``Blind'' People | A Day in Bright Darkness & 8:00 & 6  & 2 & 2016 & Malaysia & \url{https://youtu.be/xdyj1Is5IFs} \\ 
             \hline              
             \rowcolor{gray!10} 
             V20 & Blind Challenges in a Sighted World & 3:54 & 5 & 2 & 2017 & --- & \url{https://youtu.be/3pRWq8ritc8} \\ 
             \hline              
             V21 & What to expect from Orientation \& Mobility Training (O\&M) at VisionCorps & 2:21 & 7 & 2 & 2012 & Pennsylvania & \url{https://youtu.be/wU7b8rwr2dM} \\ 
            \bottomrule
    \end{tabular}
    }
    \caption{List of our collected videos~\cite{islam2024identifying, islam2024dataset}. We cropped the YouTube videos using \url{https://streamable.com}, which has a crop limit of 10 mins.} 
    \label{table:dataset}
    \end{center}
\end{table*}

\subsection{Ground Truth Labeling}
\label{app:gt-labeling}
All authors of this paper annotated the 31 video segments, visually inspecting the keyframes to label the presence of objects. 
Each author annotated a subset of segments by comparing changes between consecutive keyframes. 
The presence (1) or absence (0) of all 90 objects was recorded for each frame.

To mitigate the risk of "change blindness," a phenomenon where changes are missed due to interruptions in visual continuity~\cite{simons1997change}, keyframe pairs were displayed side-by-side, allowing authors to glance between them for comparison. 
Annotating the first keyframe of a segment typically took 5–7 minutes. 
For subsequent frames, only new object appearances or disappearances were noted, reducing annotation time to under 60 seconds in most cases. 
However, changes in frames with significant background or camera viewports required more time.

With an average of 15 keyframes per video segment, annotating each segment took approximately 20 minutes. 
At least two authors independently annotated each segment, and discrepancies were resolved collaboratively.

More details on the video collection, crucial object identification, ground truth labeling, and evaluations on state-of-the-art models are available in our prior research~\cite{islam2024identifying, islam2024dataset}.

\end{document}